%% file: main.tex
\documentclass{article} %
\usepackage{iclr2023_conference,times}

\usepackage[utf8]{inputenc} %
\usepackage[T1]{fontenc}    %
\usepackage{hyperref}       %
\usepackage{url}            %
\usepackage{booktabs}       %
\usepackage{amsfonts}       %
\usepackage{nicefrac}       %
\usepackage{microtype}      %
\usepackage{xcolor}         %
\usepackage{paralist}
\usepackage{url}
\usepackage{amsmath,amsfonts,bm}

\usepackage{algorithm}
\usepackage{algorithmic}
\usepackage{cleveref}
\usepackage{graphicx}
\usepackage{subcaption}
\usepackage{wrapfig}
\usepackage{xspace}
\usepackage{multirow}
\input{definitions}

\newcommand{\rv}[1]{{#1}}

\title{\ptitle}

\author{%
Andrew Szot$^{1,2}$,
Amy Zhang$^{1}$,
Dhruv Batra$^{1,2}$,
Zsolt Kira$^{2}$,
Franziska Meier$^{1}$\\
$^{1}$Meta AI, $^{2}$Georgia Tech
}

\iclrfinalcopy %
\begin{document}

\maketitle

\input{sections/abstract}

\input{sections/introduction}

\input{sections/related_work}

\input{sections/method}

\input{sections/reward_analysis}
\input{sections/new_experiments}
\input{sections/discussion}

\input{sections/acknowledgements}

\bibliography{main}
\bibliographystyle{iclr2023_conference}

\appendix

\input{supp/pm_experiments}

\input{supp/experiments}

\input{supp/pm_details}

\input{supp/reach_details}
\input{supp/trifinger-details}

\end{document}

%% file: definitions.tex
\newcommand*{\higherobj}{ \mathcal{L}_{\text{BC-IRL}}}
\newcommand*{\maxentobj}{ \mathcal{L}_{\text{MaxEnt-IRL}}}
\newcommand*{\bcirl}{ \mathcal{L}_{\text{BC-IRL}}}
\newcommand*{\popt}{ \text{POLICY\_OPT}}
\newcommand*{\ptitle}{BC-IRL: Learning Generalizable Reward Functions from Demonstrations}
\DeclareMathOperator{\st}{\text{s.t. }}

\DeclareMathOperator*{\argmax}{\arg\!\max}
\DeclareMathOperator*{\E}{\mathbb{E}}

\newcommand{\genmethod}{BC-IRL\xspace}
\newcommand{\ppomethod}{BC-IRL-PPO\xspace}

\newcommand{\extraj}{\tau^e}
\newcommand{\exdata}{\mathcal{D}^e}

\newcommand{\adapt}{``Policy+Reward"\xspace}
\newcommand{\scratch}{``Reward"\xspace}

\newcommand{\Adapt}{Policy+Reward\xspace}
\newcommand{\Scratch}{Reward\xspace}
\newcommand{\Zero}{Policy\xspace}

\newcommand{\usetrain}{Train $(g_\text{max}=0.2)$\xspace}
\newcommand{\useeasy}{Easy $(g_\text{max}=0.25)$\xspace}
\newcommand{\usemedium}{Medium $(g_\text{max}=0.4)$\xspace}
\newcommand{\usehard}{Hard $(g_\text{max}=0.55)$\xspace}

\newcommand{\seasy}{$(g_\text{max}=0.25)$\xspace}
\newcommand{\smed}{$(g_\text{max}=0.4)$\xspace}
\newcommand{\shard}{$(g_\text{max}=0.55)$\xspace}

\newcommand{\pmtrain}{Train\xspace}
\newcommand{\pmevaltrain}{Eval (Train)\xspace}
\newcommand{\pmevaltest}{Eval (Test)\xspace}

%% file: sections/abstract.tex
\begin{abstract}
How well do reward functions learned with inverse reinforcement learning (IRL) generalize? We illustrate that state-of-the-art IRL algorithms, which maximize a maximum-entropy objective, learn rewards that overfit to the demonstrations. Such rewards struggle to provide meaningful rewards for states not covered by the demonstrations, a major detriment when using the reward to learn policies in new situations. We introduce \genmethod, a new inverse reinforcement learning method that learns reward functions that generalize better when compared to maximum-entropy IRL approaches. In contrast to the MaxEnt framework, which learns to maximize rewards around demonstrations, \genmethod updates reward parameters such that the policy trained with the new reward matches the expert demonstrations better. We show that \genmethod learns rewards that generalize better on an illustrative simple task and two continuous robotic control tasks, achieving over twice the success rate of baselines in challenging generalization settings.
\end{abstract}

%% file: sections/introduction.tex
\begin{figure}[H]
 	\centering
 	\begin{subfigure}{0.21\textwidth}
    \includegraphics[width=\textwidth]{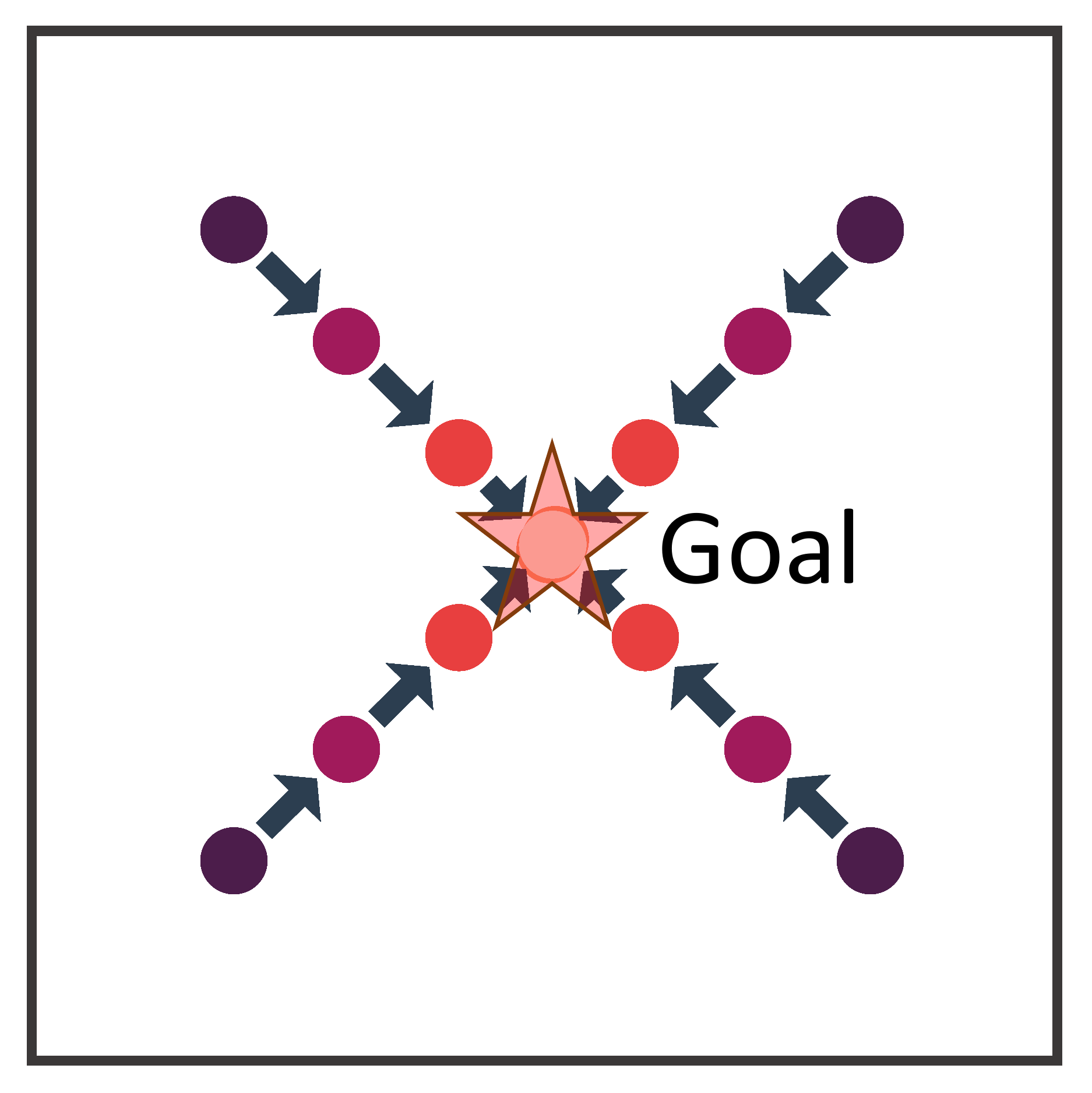}
    \caption{Demonstrations}
    \label{fig:teaser:demo}
  \end{subfigure}\qquad
   \begin{subfigure}{0.24\textwidth}
    \includegraphics[width=\textwidth]{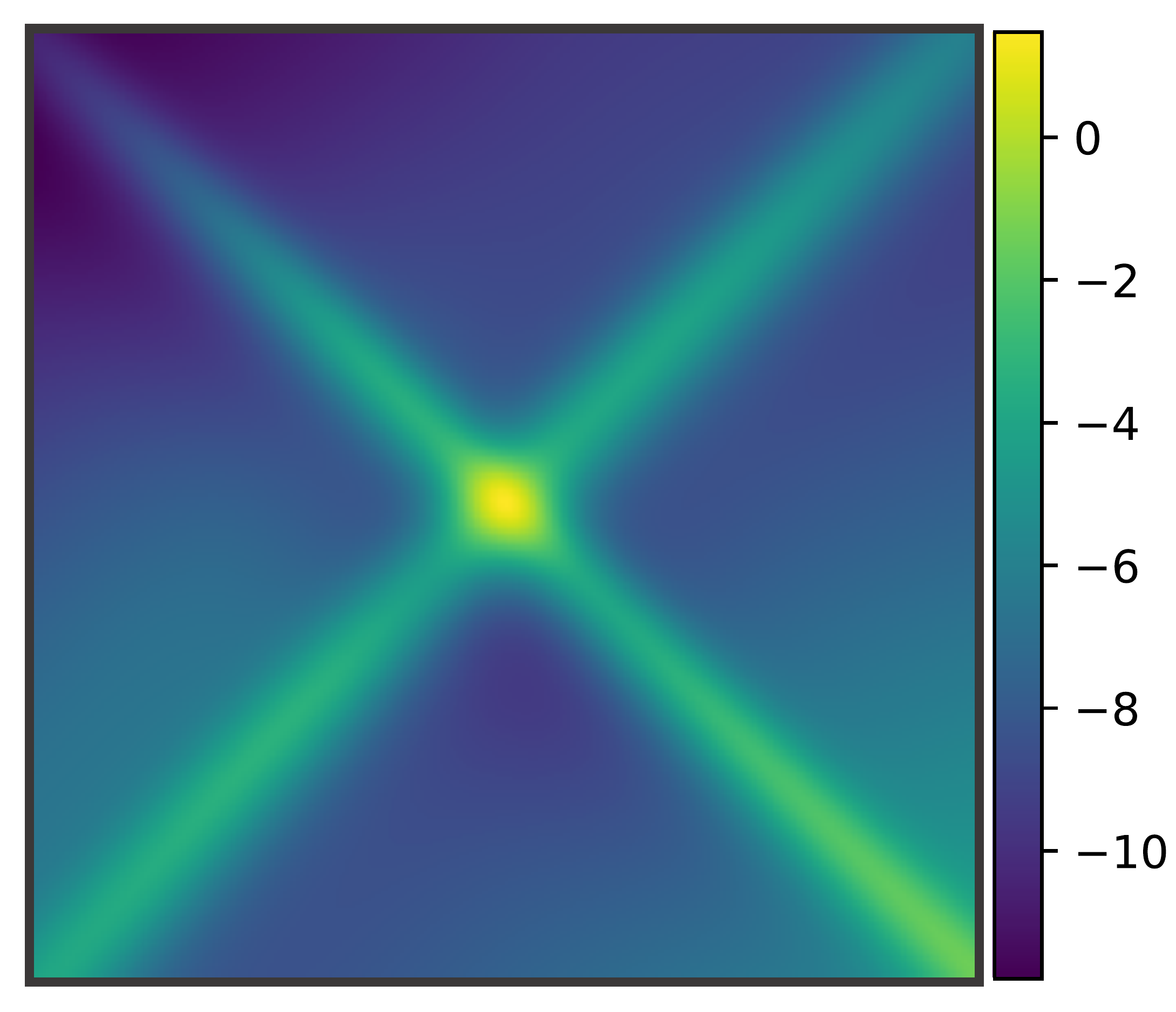}
    \caption{Max-Ent IRL Reward}
    \label{fig:teaser:me}
  \end{subfigure}\qquad
  \begin{subfigure}{0.24\textwidth}
    \includegraphics[width=\textwidth]{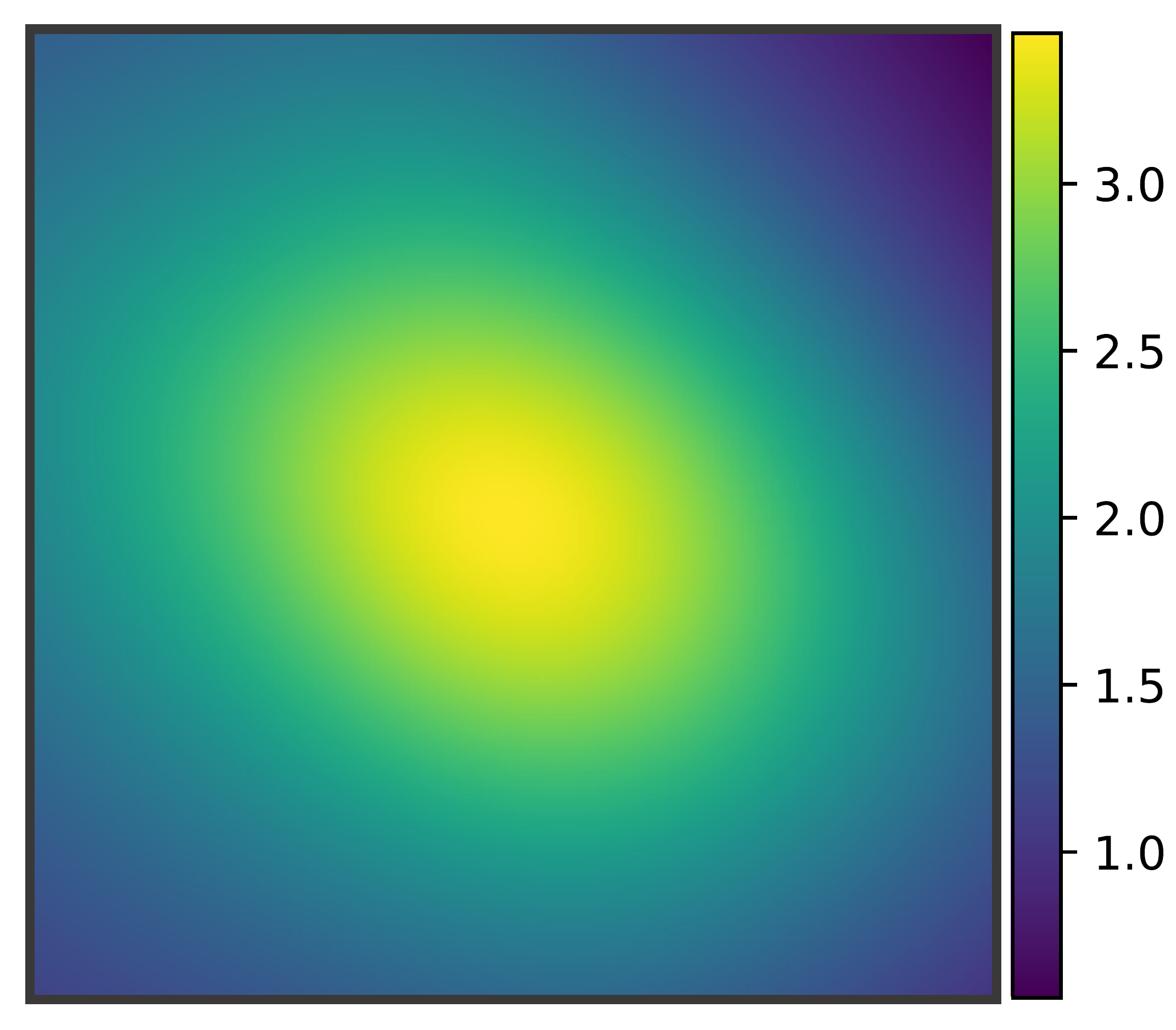}
    \caption{\genmethod Reward}
    \label{fig:teaser:mirl} 
  \end{subfigure}
 	\caption{\small 
    A visualization of learned rewards on a task where a 2D agent must navigate to the goal at the center.  \Cref{fig:teaser:demo}: Four trajectories are provided as demonstrations and the demonstrated states are visualized as points. 
    Rewards learned via Maximum Entropy are in \Cref{fig:teaser:me} and \genmethod in \Cref{fig:teaser:mirl}. Lighter colors represent larger predicted rewards.
    The MaxEnt objective overfits to the demonstrations, giving high rewards only close to the expert states, preventing the reward from providing meaningful learning signals in new states.
  }
 	\label{fig:teaser}
\end{figure}
\vspace{-20pt}
\section{Introduction}\label{sec:intro}
Reinforcement learning has demonstrated success on a broad range of tasks from navigation \cite{wijmans2019dd}, locomotion \cite{kumar2021rma, iscen2018policies}, and  manipulation \cite{qt-opt}. 
However, this success depends on specifying an accurate and informative reward signal to guide the agent towards solving the task. 
For instance, imagine designing a reward function for a robot window cleaning task. 
The reward should tell the robot how to grasp the cleaning rag, how to use the rag to clean the window, and to wipe hard enough to remove dirt, but not hard enough to break the window.
Manually shaping such reward functions is difficult, non-intuitive, and time-consuming. 
Furthermore, the need for an expert to design a reward function for every new skill limits the ability of agents to autonomously acquire new skills. 

Inverse reinforcement learning (IRL) \citep{abbeel2004apprenticeship, ziebart2008maximum, osa2018algorithmic} is one way of addressing the challenge of acquiring rewards by learning reward functions from demonstrations and then using the learned rewards to learn policies via reinforcement learning. 
When compared to direct imitation learning, which learns policies from demonstrations directly, potential benefits of IRL are at least two-fold: first, IRL does not suffer from the compounding error problem that is often observed with policies directly learned from demonstrations~\citep{ross2011reduction, barde2020adversarial}; and second, a reward function could be a more abstract and parsimonious description of the observed task that generalizes better to unseen task settings \citep{ng2000algorithms, osa2018algorithmic}. 
This second potential benefit is appealing as it allows the agent to learn a reward function to train policies not only for the demonstrated task setting (e.g. specific start-goal configurations in a reaching task) but also for unseen settings (e.g. unseen start-goal configurations), autonomously without additional expert supervision. 

However, thus far the generalization properties of reward functions learned via IRL are poorly understood. 
Here, we study the generalization of learned reward functions and find that prior IRL methods fail to learn generalizable rewards and instead overfit to the demonstrations.
\Cref{fig:teaser} demonstrates this on a task where a point mass agent must navigate in a 2D space to a goal location at the center.
An important reward characteristic for this task is that an agent, located anywhere in the state-space, should receive increasing rewards as it gets closer to the goal. 
Most recent prior work \cite{fu2017learning,ni2020f,finn2016guidedirl}  developed IRL algorithms that optimize the maximum entropy objective  \citep{ziebart2008maximum} (\Cref{fig:teaser:me}), which fails to capture goal distance in the reward. Instead, the MaxEnt objective leads to rewards that separate non-expert from expert behavior by maximizing reward values along the expert demonstration. While useful for imitating the experts, the MaxEnt objective prevents the IRL algorithms from learning to assign meaningful rewards to other parts of the state space, thus limiting generalization of the reward function. 

As a remedy to the reward generalization challenge in the maximum entropy IRL framework, we propose a new IRL framework called \textbf{Behavioral Cloning Inverse Reinforcement Learning (\genmethod)}. In contrast to the MaxEnt framework, which learns to maximize rewards around demonstrations, the \genmethod framework updates reward parameters such that the policy trained with the new reward matches the expert demonstrations better. This is akin to the model-agnostic meta-learning \citep{finn2017model} and loss learning  \citep{bechtle2019meta} frameworks where model or loss function parameters are learned such that the downstream task performs well when utilizing the meta-learned parameters. 
By using gradient-based bi-level optimization \cite{higher}, \genmethod can optimize the behavior cloning loss to learn the reward, rather than a separation objective like the maximum entropy objective. 
Importantly, to learn the reward, \genmethod differentiates through the reinforcement learning policy optimization, which incorporates exploration and requires the reward to provide a meaningful reward throughout the state space to guide the policy to better match the expert.
We find \genmethod learns more generalizable rewards (\Cref{fig:teaser:mirl}), and achieves over twice the success rate of baseline IRL methods in challenging generalization settings.

Our contributions are as follows:
1) The general \genmethod framework for learning more generalizable rewards from demonstrations, and a specific \ppomethod variant that uses PPO as the RL algorithm. 
2) A quantitative and qualitative analysis of reward functions learned with \genmethod and Maximum-Entropy IRL variants on a simple task for easy analysis. 
3) An evaluation of our novel \genmethod algorithm on two continuous control tasks against state-of-the-art IRL and IL methods. Our method learns rewards that transfer better to novel task settings.

%% file: sections/related_work.tex
\vspace{-10pt}
\section{Background and Related Work}
\label{sec:related_work}

We begin by reviewing Inverse Reinforcement Learning through the lense of bi-level optimization. We assume access to a rewardless Markov decision process (MDP) defined through the tuple $ \mathcal{M} = ( \mathcal{S}, \mathcal{A}, \mathcal{P}, \rho_0, \gamma, H)$ for state-space $ \mathcal{S}$, action space $ \mathcal{A}$, transition distribution $ \mathcal{P}(s' | s,a)$, initial state distribution $\rho_0$, discounting factor $\gamma$, and episode horizon $H$. 
We also have access to a set of expert demonstration trajectories $ \exdata = \left\{ \extraj_i \right\}_{i=1}^{N}$ where each trajectory is a sequence of state, action tuples. %

IRL learns a parameterized reward function $R_\psi(\tau_i)$ which assigns a trajectory a scalar reward. 
Given the reward, a policy $\pi_\theta(a | s)$ is learned which maps from states to a distribution over actions.
The goal of IRL is to produce a reward $R_\psi$, such that a policy trained to maximize the sum of (discounted) rewards under this reward function matches the behavior of the expert. 
This is captured through the following bi-level optimization problem: 
\begin{subequations}
  \begin{align}
    \label{eq:irl:outer_gen}
    \min_{\psi} &\mathcal{L}_\text{IRL}(R_\psi; \pi_\theta) &\text{\bf{(outer obj.)} }\\
    \label{eq:irl:inner_gen}\st &\theta \in \argmax_{\theta} g(R_\psi, \theta) &\text{\bf{(inner obj.)} }  
  \end{align}
\end{subequations}
where $\mathcal{L}_\text{IRL}(R_\psi; \pi_\theta)$ denotes the IRL loss and measures the performance of the learned reward $R_\psi$ and policy $\pi_\theta$; $g(R_\psi, \theta)$ is the reinforcement learning objective used to optimize policy parameters $\theta$. Algorithms for this bi-level optimization consist of an outer loop (\eqref{eq:irl:outer_gen}) that optimizes the reward and an inner loop (\eqref{eq:irl:inner_gen}) that optimizes the policy given the current reward. 

\textbf{Maximum Entropy IRL:}
Early work on IRL learns rewards by separating non-expert from expert trajectories \citep{ng2000algorithms,abbeel2004apprenticeship, abbeel2010autonomous}. 
A primary challenge of these early IRL algorithms was the ambiguous nature of learning reward functions from demonstrations: many possible policies exist for a given demonstration, and thus many possible rewards exist.
The Maximum Entropy (MaxEnt) IRL framework \citep{ziebart2008maximum} seeks to address this ambiguity, by learning a reward (and policy) that is as non-committal (uncertain) as possible, while still explaining the demonstrations. More concretely, given reward parameters $\psi$, MaxEnt IRL optimizes the log probability of the expert trajectories $\extraj$ from demonstration dataset $ \exdata$ through the following loss,
\begin{align*}
  \maxentobj(R_\psi) &= -\mathbb{E}_{\extraj \sim \exdata} \left[\log p(\extraj | \psi)\right] 
  = -\mathbb{E}_{\extraj \sim \exdata}  \left[\log \frac{\exp \left( R_{\psi} (\extraj) \right)}{Z(\psi)} \right]\\
  &= -\mathbb{E}_{\extraj \sim \exdata}\left[R_\psi(\extraj)\right] + \log Z(\psi).
\end{align*} 
A key challenge of MaxEnt IRL is estimating the partition function $Z(\psi) = \int \exp R_\psi d\tau$.
\cite{ziebart2008maximum} approximate $Z$ in small discrete state spaces with dynamic programming. 

\textbf{MaxEnt from the Bi-Level perspective:} However, computing the partition functions becomes intractable for high-dimensional and continuous state spaces. 
Thus algorithms approximate $Z$ using samples from a policy optimized via the current reward. 
This results in the partition function estimate being a function of the current policy $\log \hat{Z}(\psi ; \pi_\theta)$.
As a result, MaxEnt approaches end up following the bi-level optimization template by iterating between: 1) updating reward function parameters given current policy samples via the outer objective (\eqref{eq:irl:outer_gen}); and 2) optimizing the policy parameters with the current reward parameters via an inner policy optimization objective and algorithm \eqref{eq:irl:inner_gen}. 
For instance, model-based IRL methods such as \cite{wulfmeier2017large, levine2012continuous, englert2017inverseRL} use model-based RL (or optimal control) methods to optimize a policy (or trajectory), while model-free IRL methods such as \cite{kalakrishnan_2013_irl, boularias2011relativeirl,finn2016guided,finn2016connection} learn policies via model-free RL in the inner loop. 
All of these methods use policy rollouts to approximate either the partition function of the maximum-entropy IRL objective or its gradient with respect to reward parameters in various ways (outer loop). 
For instance \cite{finn2016guided} learn a stochastic policy $q(\tau)$, and sample from that to estimate $Z(\psi) \approx \frac{1}{M} \sum_{\tau_i \sim q(\tau)}\frac{\exp R_{\psi}(\tau_i)}{q(\tau_i)}$ with $M$ samples from $q(\tau)$. 
\cite{fu2017learning} with adversarial IRL (AIRL) follow this idea and view the problem as an adversarial training process between policy $\pi_\theta(a | s)$ and discriminator $D(s) = \frac{\exp R_\psi(s)}{\exp R_{\psi}(s) + \pi_{\theta}(a|s)}$.
\cite{ni2020f} analytically compute the gradient of the $f$-divergence between the expert state density and the MaxEnt state distribution, circumventing the need to directly compute the partition function. 
 
\textbf{Meta-Learning and IRL:}
Like some prior work \citep{xu2019learning, yu2019meta, wang2021meta, gleave2018multi,seyed2019smile}, \genmethod combines meta-learning and inverse reinforcement learning. 
However, these works focus on fast adaptation of reward functions to new tasks for MaxEnt IRL through meta-learning.
These works require demonstrations of the new task to adapt the reward function.
\genmethod algorithm is a fundamentally new way to learn reward functions and does not require demonstrations for new test settings.
Most related to our work is \cite{das2020model}, which also uses gradient-based bi-level optimization to match the expert.
However, this approach requires a pre-trained dynamics model.
Our work generalizes this idea since \genmethod can optimize general policies, allowing any objective that is a function of the policy and any differentiable RL algorithm. 
We show our method, without an accurate dynamics model, outperforms \cite{das2020model} and scales to more complex tasks where \cite{das2020model} fails to learn.

\textbf{Generalization in IRL:} 
Some prior works have explored how learned rewards can generalize to training policies in new situations. 
For instance, \cite{fu2017learning} explored how rewards can generalize to training policies under changing dynamics. However, most prior work focuses on improving policy generalization to unseen task settings by addressing challenges introduced by the adversarial training objective of GAIL
\citep{xu2019positive,zolna2020combating,zolna2019task,lee2021generalizable, barde2020adversarial,jaegle2021imitation,dadashi2020primal}. Finally, in contrast to most related work on generalization, our work focuses on analyzing and improving reward function transfer to new task settings.

%% file: sections/method.tex
\section{Learning Rewards via Behavioral Cloning Inverse Reinforcement Learning (\genmethod)}
\label{sec:main_method} 

We now present our algorithm for learning reward functions via behavioral cloning inverse reinforcement learning.
We start by contrasting the maximum entropy and imitation loss objectives for inverse reinforcement learning in \Cref{sec:method:irl-objectives}. 
We then introduce a general formulation for \genmethod in \Cref{sec:method:meta-irl}, and present an algorithmic instantiation that optimizes a BC objective to update the \emph{reward} parameters via gradient-based bi-level optimization with a model-free RL algorithm in the inner loop in \Cref{sec:method:ppo_mirl}. 

\subsection{Outer Objectives: Max-Ent vs Behavior Cloning }\label{sec:method:irl-objectives}
In this work, we study an alternative IRL objective from the maximum entropy objective. 
While this maximum entropy IRL objective has led to impressive results, it is unclear how well this objective is suited for learning reward functions that generalize to new task settings, such as new start and goal distributions.
Intuitively, assigning a high reward to demonstrated states (without task-specific hand-designed feature engineering) makes sense when you want to learn a reward function that can recover exactly the expert behavior, but it leads to reward landscapes that do not necessarily capture the essence of the task (e.g. to reach a goal, see \Cref{fig:teaser:me}). 

Instead of specifying an IRL objective that is directly a function of reward parameters (like maximum entropy), we aim to measure the reward function's performance through the policy that results from optimizing the reward.  
With such an objective, we can  optimize reward parameters for what we care about: for the resulting policy to match the behavior of the expert.
The behavioral cloning (BC) loss measures how well the policy and expert actions match, defined for continuous actions as $\mathbb{E}_{(s_t, a_t) \sim \extraj} \left( \pi_\theta(s_t) - a_t \right)^2 $ where $\extraj$ is an expert demonstration trajectory. 
Policy parameters $\theta$ are a result of using the current reward parameters $\psi$, which we can make explicit by making $\theta$ a function of $\psi$ in the objective: $\bcirl = \mathbb{E}_{(s_t, a_t) \sim \extraj}  (\pi_{\theta(\psi)}(s_t)  - a_t)^2 $. 
The IRL objective is now formulated in terms of the policy rollout ``matching" the expert demonstration through the BC loss. 

We use the chain-rule to decompose the gradient of $\bcirl$ with respect to reward parameters $\psi$. We also expand how the policy parameters $\theta(\psi)$ are updated via a REINFORCE update with learning rate $ \alpha$ to optimize the current reward $R_\psi$ (but any differentiable policy update applies).
\begin{align}
  \label{eq:bi_level_opt} 
   \frac{\partial}{\partial \psi} \bcirl
   &= \frac{\partial}{\partial \psi} \left[ 
     \E_{(s_t, a_t) \sim \extraj}
     \left[\left( \pi_{\theta(\psi)}(s_t) - a_{t} \right)^{2} \right]
   \right] 
   = \E_{(s_t, a_t) \sim \extraj} \left[ 2 \left( \pi_{\theta(\psi)}(s_t) - a_{t} \right)\right] \frac{\partial}{\partial \psi}   \pi_{\theta(\psi)} \nonumber \\
   &\text{where   } \theta(\psi) = \theta_{\text{old}} + \alpha \E_{(s_t, a_t) \sim \pi_{\theta_{\text{old}}}} \left[ 
     \left( \sum_{k=t+1}^{T} \gamma^{k-t-1} R_\psi(s_k) \right) \nabla \ln \pi_{\theta_{\text{old}}} (a_t | s_t)
   \right] 
\end{align} 
Computing the gradient for the reward update in \Cref{eq:bi_level_opt} includes samples from $ \pi$ collected in the reinforcement learning (RL) inner loop.
This means the reward is trained on diverse states beyond the expert demonstrations through data collected via exploration in RL.
As the agent explores during training, \genmethod must provide a meaningful reward signal throughout the state-space to guide the policy to better match the expert.
Note that this is a fundamentally different reward update rule as compared to current state-of-the-art methods that maximize a maximum entropy objective.
We show in our experiments that this results in twice as high success rates compared to state-of-the-art MaxEnt IRL baselines in challenging generalization settings, demonstrating that \genmethod learns more generalizable rewards that provide meaningful rewards beyond the expert demonstrations.

The BC loss updates only the reward, as opposed to updating the policy as typical BC for imitation learning does \cite{bain1995framework}.
\genmethod is a IRL method that produces a reward, unlike regular BC that learns only a policy.
Since \genmethod uses RL, not BC, to update the policy, it avoids the pitfalls of BC for policy optimization such as compounding errors.
Our experiments show that policies trained with rewards from \genmethod generalize over twice as well to new settings as those trained with BC.
In the following section, we show how to optimize this objective via bi-level optimization.

\vspace{-5pt}
\subsection{BC-IRL}\label{sec:method:meta-irl} 
\vspace{-5pt}

We formulate the IRL problem as a gradient-based bi-level optimization problem, where the outer objective is optimized by differentiating through the optimization of the inner objective. 
We first describe how the policy is updated with a fixed reward, then how the reward is updated for the policy to better match the expert.

\begin{wrapfigure}{R}{0.55\textwidth}
\vspace{-20pt}
\begin{minipage}{0.55\textwidth}
  \begin{algorithm}[H]
  \begin{algorithmic}[1]
  \footnotesize{
    \STATE{Initial reward $R_\psi$, policy $\pi_\theta$}
    \STATE{Policy updater $\popt(R, \pi)$}
    \STATE{Expert demonstrations $ \exdata$}
    \FOR{each epoch}
      \STATE{Policy Update:}
      \STATE{$ \theta' \gets \popt(R_\psi, \pi_\theta) $}
      \STATE{Sample demo batch $\extraj \sim \exdata$}
      \STATE{Compute IRL loss}
      \STATE{$ \bcirl = \mathbb{E}_{(s_t, a_t) \sim \extraj} \left( \pi_{\theta'}(s_t) - a_t \right)^2 $}
      \STATE{Compute gradient of IRL loss wrt reward}
      \STATE{$\nabla_\psi \bcirl = \frac{\partial \bcirl}{\partial \theta'} \frac{\partial \popt(R_\psi, \pi_\theta)}{\partial \psi}$}
      \STATE{$\psi \gets \psi - \nabla_{\psi} \bcirl$}
    \ENDFOR
  }
  \end{algorithmic}
  \caption{\genmethod (general framework)}
  \label{algo:method:mirl}
  \end{algorithm}
\end{minipage}
  \vspace{-10pt}
\end{wrapfigure}

\textbf{Inner loop (policy optimization):} The inner loop optimizes policy parameters $\theta$ given current reward function $R_\psi$. 
The inner loop takes $K$ gradient steps to optimize the policy given the current reward.
Since the reward update will differentiate through this policy update, we require the policy update to be differentiable with respect to the reward function parameters.
Thus, any reinforcement learning algorithm which is differentiable with respect to the reward function parameters can be plugged in here, which is the case for many policy gradient and model-based methods. 
However, this does not include value-based methods such as DDPG \cite{lillicrap2015continuous} or SAC \cite{haarnoja2018soft} that directly optimize value estimates since the reward function is not directly used in the policy update. 

\textbf{Outer loop (reward optimization)}: The outer loop optimization updates the reward parameters $\psi$ via gradient descent. 
More concretely: after the inner loop, we compute the gradient of the outer loop objective $\nabla_\psi \higherobj$ wrt to reward parameters $\psi$ by propagating through the inner loop. 
Intuitively, the new policy is a function of reward parameters since the old policy was updated to better maximize the reward.
The gradient update on $\psi$ tries to adjust reward function parameters such that the policy trained with this reward produces trajectories that match the demonstrations more closely. 
We use \citet{higher} for this higher-order optimization. 

\genmethod is summarized in \Cref{algo:method:mirl}.
Line 5 describes the inner loop update, where we update the policy $\pi_\theta$ to  maximize the current reward $R_\psi$. 
Lines 6-7 compute the BC loss between the updated policy $\pi_{\theta'}$ and expert actions sampled from expert dataset $\exdata$.
The BC loss is then used in the outer loop to perform a gradient step on reward parameters in lines 8-9, where the gradient computation requires differentiating through the policy update in line 5.

\vspace{-15pt}
\subsection{\ppomethod}
\vspace{-5pt}
\label{sec:method:ppo_mirl}

We now instantiate a specific version of the \genmethod framework that uses proximal policy optimization (PPO) \cite{schulman2017proximal} to optimize the policy in the inner loop. 
This specific version, called \ppomethod, is summarized in \Cref{algo:method:ppo_mirl}.

\begin{wrapfigure}{R}{0.55\textwidth}
\vspace{-20pt}
\begin{minipage}{0.53\textwidth}
  \begin{algorithm}[H]
  \begin{algorithmic}[1]
  \footnotesize{
    \STATE{Initial reward $R_\psi$, policy $\pi_\theta$, value function $V_\nu$}
    \STATE{Expert demonstrations $ \exdata$}
    \FOR{each epoch}
      \FOR{$k=1 \to K$} 
        \STATE{Run policy $\pi_{\theta}$ in environment for $T$ timesteps}
        \STATE{Compute rewards $\hat{r}_t^{\psi}$ for rollout with current $R_\psi$}
        \STATE{Compute advantages $\hat{A}^{\psi}$ using $\hat{r}^{\psi}$ and $V_\nu$}
        \STATE{Compute $ \mathcal{L}_{\text{PPO}}$ using $\hat{A}^{\psi}$}
        \STATE{Update $\pi_\theta$ with $\nabla_\theta \mathcal{L}_{\text{PPO}}$}
      \ENDFOR
      \STATE{Sample demo batch $\extraj \sim \exdata$}
      \STATE{Compute $ \bcirl = \mathbb{E}_{(s_t, a_t) \sim \extraj} \left( \pi_\theta(s_t) - a_t \right)^2 $}
      \STATE{Update reward $R_\psi$ with $ \nabla_{\psi} \bcirl$}
    \ENDFOR
  }
  \end{algorithmic}
  \caption{\ppomethod}
  \label{algo:method:ppo_mirl}
  \end{algorithm}
  \end{minipage}
\vspace{-10pt}
\end{wrapfigure}

\ppomethod learns a state-only parameterized reward function $R_{\psi}(s)$, which assigns a state $s \in \mathcal{S}$ a scalar reward. 
The state-only reward has been shown to lead to rewards that generalize better \cite{fu2017learning}. 
\ppomethod begins by collecting a batch of rollouts in the environment from the current policy (line 5 of \Cref{algo:method:ppo_mirl}). 
For each state $s$ in this batch we evaluate the learned reward function $R_\psi(s)$ (line 6). 
From this sequence of rewards, we compute the advantage estimates $\hat{A}_t$ for each state (line 7).  
As is typical in PPO, we also utilize a learned value function $V_\nu(s_t)$ to predict the value of the starting and ending state for partial episodes in the rollouts. 
This learned value function $V_\nu$ is trained to predict the sum of future discounted rewards for the current reward function $R_\psi$ and policy $\pi_\theta$ (part of $ \mathcal{L}_{\text{PPO}}$ in line 8). 
Using the advantages, we then compute the PPO update (line 9 of \Cref{algo:method:ppo_mirl}) using the standard PPO loss in equation 8 of \citet{schulman2017proximal}.
Note the advantages are a function of the reward function parameters used to compute the rewards, so PPO is differentiable with respect to the reward function. 
Next, in the outer loop update, we update the reward parameters, by sampling a batch of demonstration transitions (line 11), computing the behavior cloning IRL objective $\mathcal{L}_\text{BC-IRL}$ (line 12), and updating the reward parameters $\psi$ via gradient descent on $\mathcal{L}_\text{BC-IRL}$ (line 13).
Finally, in this work, we perform one policy optimization step ($K=1$)  per reward function update. 
Furthermore, rather than re-train a policy from scratch for every reward function iteration, we initialize each inner loop from the previous $\pi_{\theta}$. 
This initialization is important in more complex domains where $K$ would otherwise have to be large to acquire a good policy from scratch.

%% file: sections/reward_analysis.tex
\vspace{-5pt}
\section{Illustration \& Qualitative Analysis of Learned Rewards}
\vspace{-5pt}
\label{sec:reward_analysis}

\input{sections/figures/pm}

We first analyze the rewards learned by different IRL methods in a 2D point mass navigation task.
The purpose of this analysis is to test our hypothesis that our method learns more generalizable rewards compared to maximum entropy baselines in simple low-dimensional settings amenable to intuitive visualizations.
Specifically, we compare \ppomethod to the following baselines. 

\textbf{Exact MaxEntIRL (MaxEnt)} \cite{ziebart2008maximum}: The exact MaxEntIRL method where the partition function is exactly computed by discretizing the state space.

\textbf{Guided Cost Learning (GCL)} \cite{finn2016guided}: Uses the maximum-entropy objective to update the reward. The partition function is approximated via adaptive sampling. 

\textbf{Adversarial IRL (AIRL)} \cite{fu2017learning}: An IRL method that uses a learned discriminator to distinguish expert and agent states. As described in \cite{fu2017learning} we also use a shaping network $h$ during reward training, but only visualize and transfer the reward approximator $g$. 

\textbf{f-IRL} \cite{ni2021f}: Another MaxEntIRL based method, f-IRL computes the analytic gradient of the f-divergence between the agent and expert state distributions. We use the JS divergence version.

Our method does not require demonstrations at test time, instead we transfer our learned rewards zero-shot. Thus we forego comparisons to other meta-learning methods, such as \cite{xu2019learning}, which require test time demonstrations.
While a direct comparison with \cite{das2020model} is not possible because their method assumes access to a pre-trained dynamics model, we conduct a separate study comparing their method with an oracle dynamics model against \genmethod in  \Cref{14355}. 
All baselines use PPO \cite{schulman2017proximal} for policy optimization as commonly done in prior work \cite{orsini2021matters}. %
All methods learn a state-dependent reward $r_\psi(s)$, and a policy $\pi(s)$, both parametrized as neural networks.
Further details are described in \Cref{sec:pm_hyperparams}.

\looseness=-1
The 2D point navigation tasks consist of a point agent policy that outputs a desired change in $(x,y)$ position (velocity) $(\Delta x, \Delta y)$ at every time step.
The task has a trajectory length of $T=5$ time steps with 4 demonstrations.
\Cref{fig:qual_results:pm_demo} visualizes the expert demonstrations where darker points are earlier time steps.
The agent starting state distribution is centered around the starting state of each demonstration.

\Cref{fig:main_qual_results}b,c visualize the rewards learned by \genmethod and the AIRL baseline.
Lighter regions indicate higher rewards.
In \Cref{fig:qual_results:pm_mirl}, \genmethod learns a reward that looks like a quadratic bowl centered at the origin, which models the distance to the goal across the entire state space. 
AIRL, the maximum entropy baseline, visualized in \Cref{fig:qual_results:pm_airl}, learns a reward function where high rewards are placed on the demonstrations and low rewards elsewhere. 
Other baselines are visualized in Appendix~\Cref{fig:all_qual_pm}.

To analyze the generalization capabilities of the learned rewards we use them to train policies on a new starting state distribution (visualized in Appendix~\Cref{fig:pm_nav_start_state}).
Concretely, a newly initialized policy is trained from scratch to maximize the learned reward from the testing start state distribution. 
\rv{The policy is trained with 5 million environment steps, which is the same number of steps as for learning the reward.}
The testing starting state distribution has no overlap with the training start state distribution. 
Policy optimization at test time is also done with PPO.
The \Cref{fig:main_qual_results}d,e display trajectories from the trained policies where darker points again correspond to earlier time steps. 

This qualitative evaluation shows that \genmethod learns a meaningful reward for states not covered by the demonstrations. 
Thus at test time agent trajectories are guided towards the goal with the terminal states (lightest points) close to the goal.
The X-shaped rewards learned by the baselines do not provide meaningful rewards in the testing setting as they assign uniformly low rewards to states not covered by the demonstration. 
\rv{This provides poor reward shaping which prevents the agent from reaching the goal within the 5M training interactions with the environment.}
This results in agent trajectories that do not end close to the goal \rv{by the end of training}.

\begin{table}[t] 
  \centering
 \resizebox{0.75\textwidth}{!}{
  \input{sections/tables/pm}
 }
  \caption{\small
    Distance to the goal for the point mass navigation task where numbers are mean and standard error for 3 seeds and 100 evaluation episodes per seed.  
    \pmtrain is policy trained during reward learning. 
    MaxEnt does not learn a policy during reward learning thus its performance is ``NA".
    \pmevaltrain uses the learned reward to train a policy from scratch on the same distribution used to train the reward. 
    \pmevaltest  measures the ability of the learned reward to generalize to a new starting state distribution. 
  }
  \label{table:toy_quant_results}
  \vspace{-10pt}
\end{table}

Next, we report quantitative results in \Cref{table:toy_quant_results}.
We evaluate the performance of the policy trained at test time by reporting the distance from the policy's final trajectory state $s_T$ to the goal $g$: $ \lVert s_{T} - g \rVert_2^2$.
We report the final train performance of the algorithm (``Train"), along with the performance of the policy trained from scratch with the learned reward in the train distribution ``Eval (Train)" and testing distribution ``Eval (Test)".
These results confirm that \genmethod learns more generalizable rewards than baselines.
Specifically, \genmethod achieves a lower distance on the testing starting state distribution at 0.04, compared to 0.53, 1.6, and 0.36 for AIRL, GCL, and MaxEnt respectively.
Surprisingly, \genmethod even performs better than exact MaxEnt, which uses privileged information about the state space to estimate the partition function. 
This fits with our hypothesis that our method learns more generalizable rewards than MaxEnt, even when the MaxEnt objective is exactly computed. 
We repeat this analysis for a version of the task with an obstacle blocking the path to the goal in \Cref{sec:pmo_nav} and reach the same findings even when \genmethod must learn an asymmetric reward function.
We also compare learned rewards to manually defined rewards in \Cref{sec:manual_rewards}.

\rv{
  Despite baselines learning rewards that do not generalize beyond the demonstrations, with enough environment interactions, policies trained under these rewards will eventually reach the high-rewards along the expert demonstrations.
  Since all demonstrations reach the goal in the point mass task, the X-shaped reward baselines learn have high-reward at the center.
  Despite the X-shaped providing little reward shaping off the X, with enough environment interactions, the agent eventually discovers the high-reward point at the goal.
  After training AIRL for 15M steps, 3x the number of steps for reward learning and the experiments in \Cref{table:toy_quant_results} and \Cref{fig:main_qual_results}, the policy eventually reaches $ 0.08 \pm 0.01$ distance to the goal.
  In the same setting, \genmethod achieves $ 0.04 \pm 0.01$ distance to the goal in under 5M steps. 
  The additional performance gap is due to BC-IRL learning a reward with a maximum reward value closer to the center ($ 0.02$ to the center) compared to AIRL ($ 0.04$ to the center).
}

%% file: sections/figures/pm.tex
\begin{figure*}[t]
  \centering
    \begin{subfigure}[t]{0.19\textwidth}
    \centering
    \includegraphics[width=.84\textwidth]{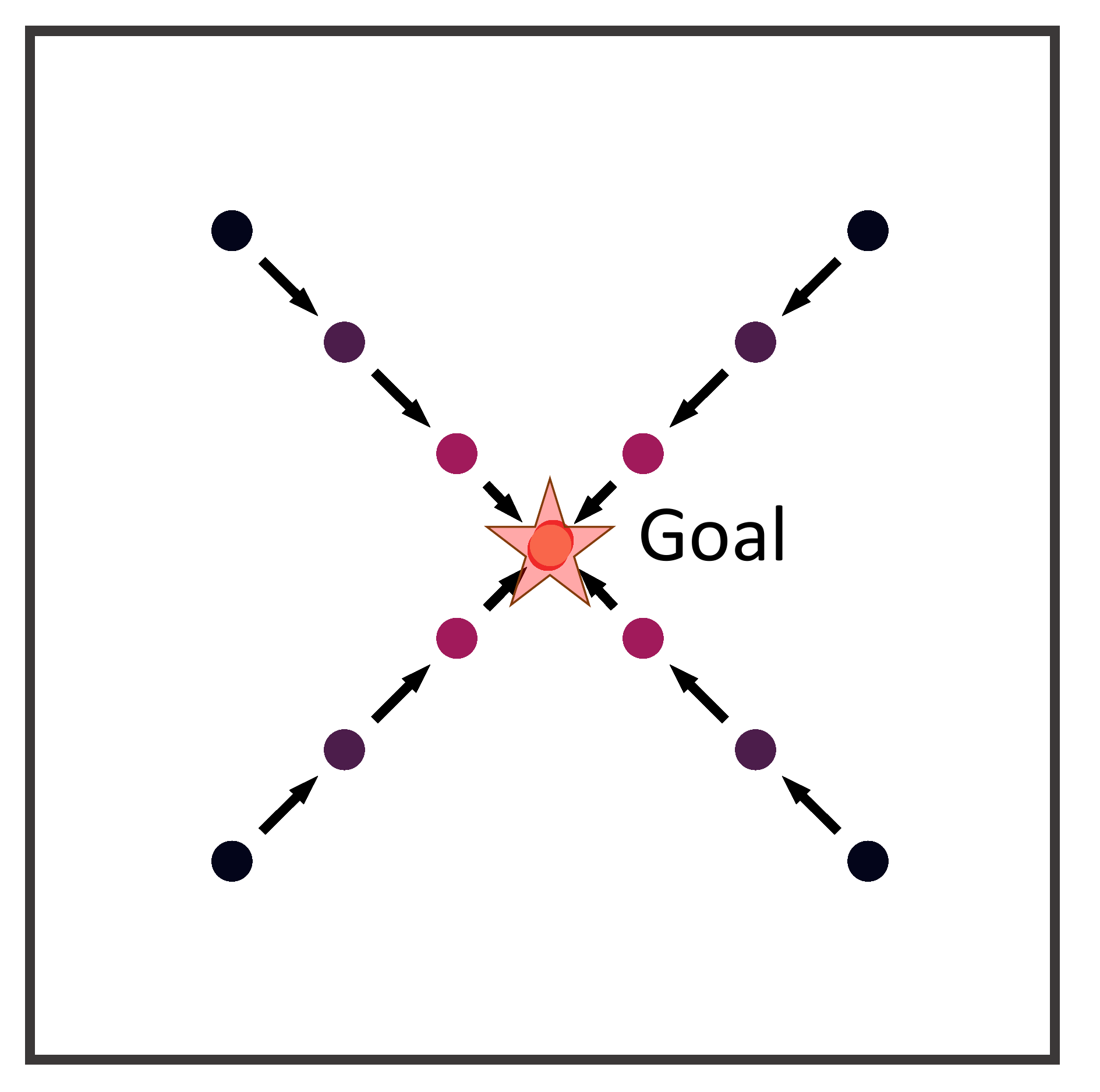}
    \caption{\small
     Task + Demos \\
    }
    \label{fig:qual_results:pm_demo} 
  \end{subfigure}
  \begin{subfigure}[t]{0.19\textwidth}
    \includegraphics[width=\textwidth]{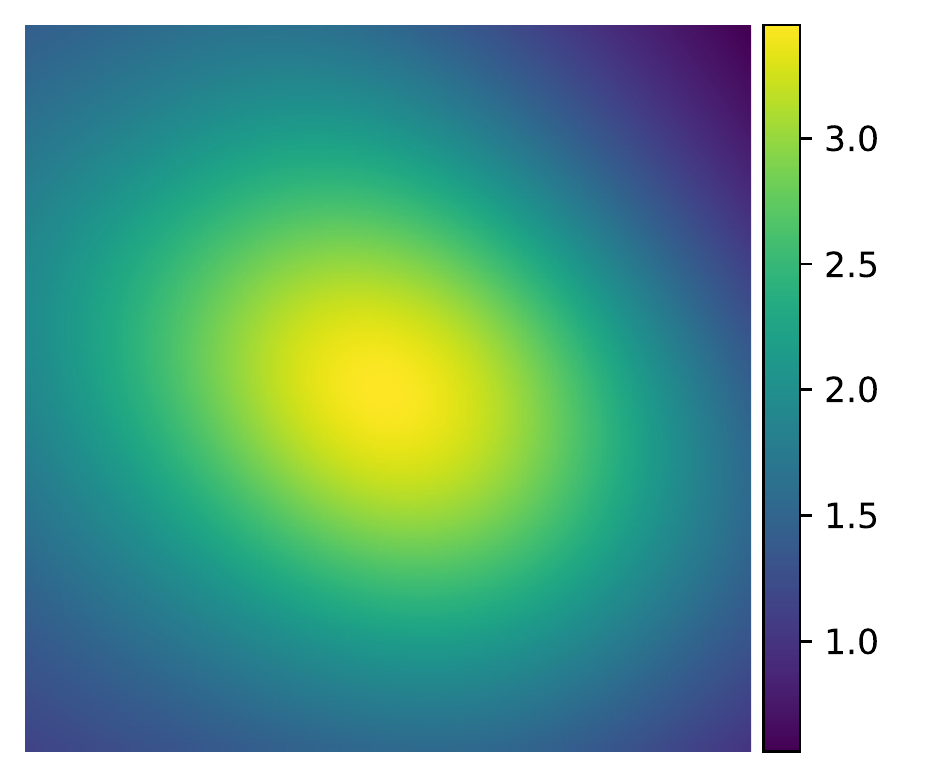}
    \caption{\small
      Ours Reward \\
    }
    \label{fig:qual_results:pm_mirl} 
  \end{subfigure}
  \begin{subfigure}[t]{0.19\textwidth}
    \includegraphics[width=\textwidth]{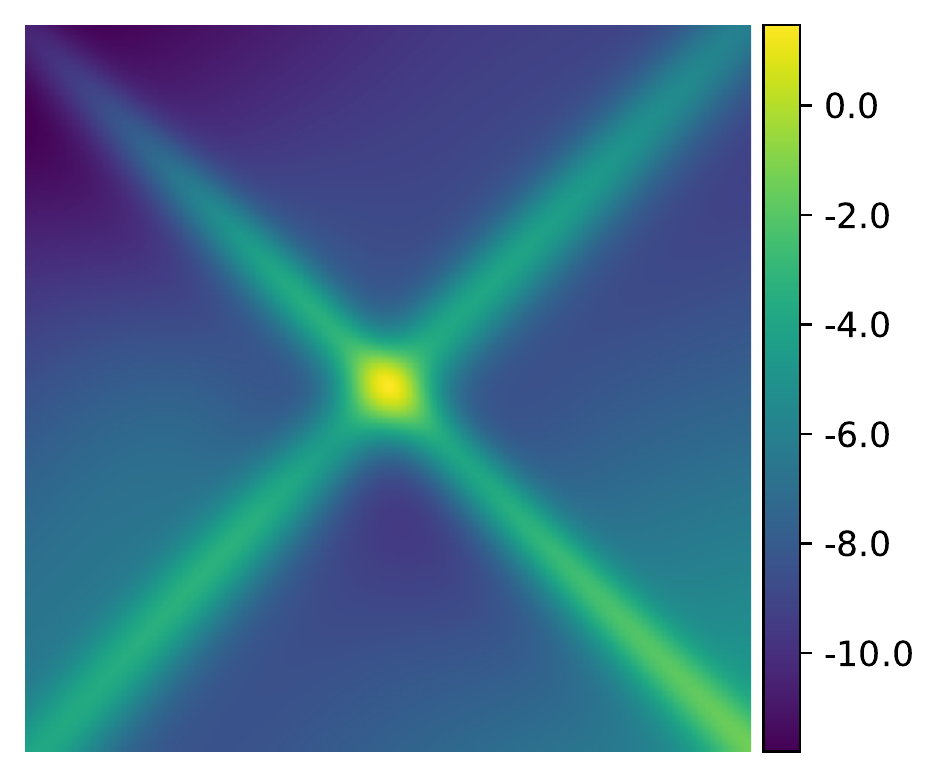}
    \caption{
      \small
      AIRL Reward  \\
    }
    \label{fig:qual_results:pm_airl} 
  \end{subfigure}
    \begin{subfigure}[t]{0.19\textwidth}
    \includegraphics[width=\textwidth]{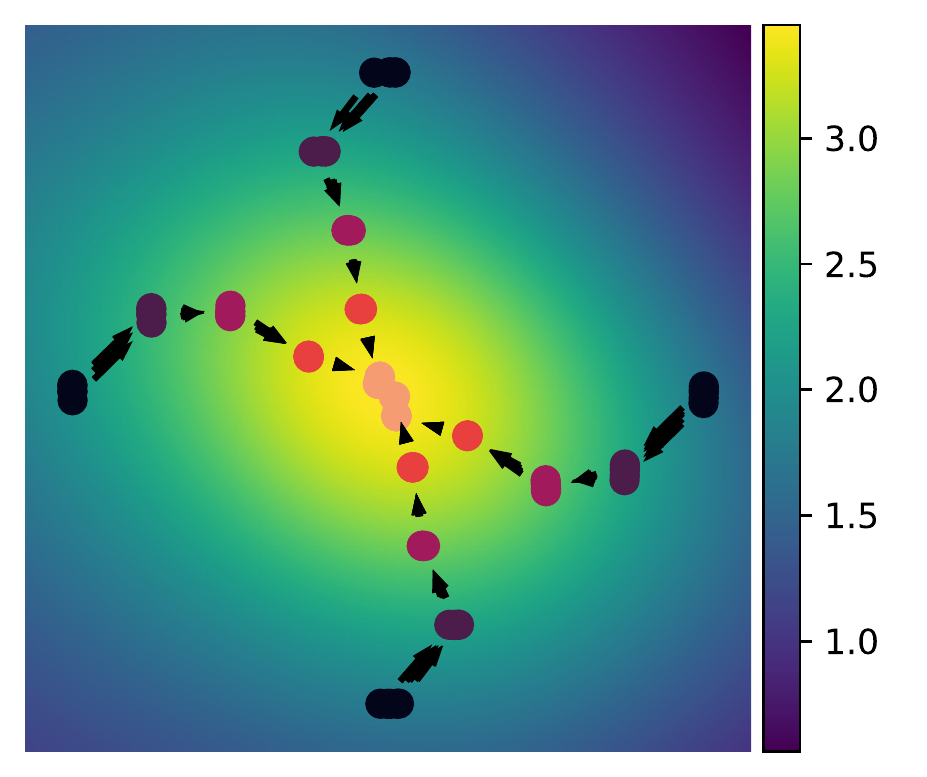}
    \caption{
      \small
      Ours Test  \\
    }
    \label{fig:qual_results:eval_pm_mirl}
  \end{subfigure}
  \begin{subfigure}[t]{0.19\textwidth}
    \includegraphics[width=\textwidth]{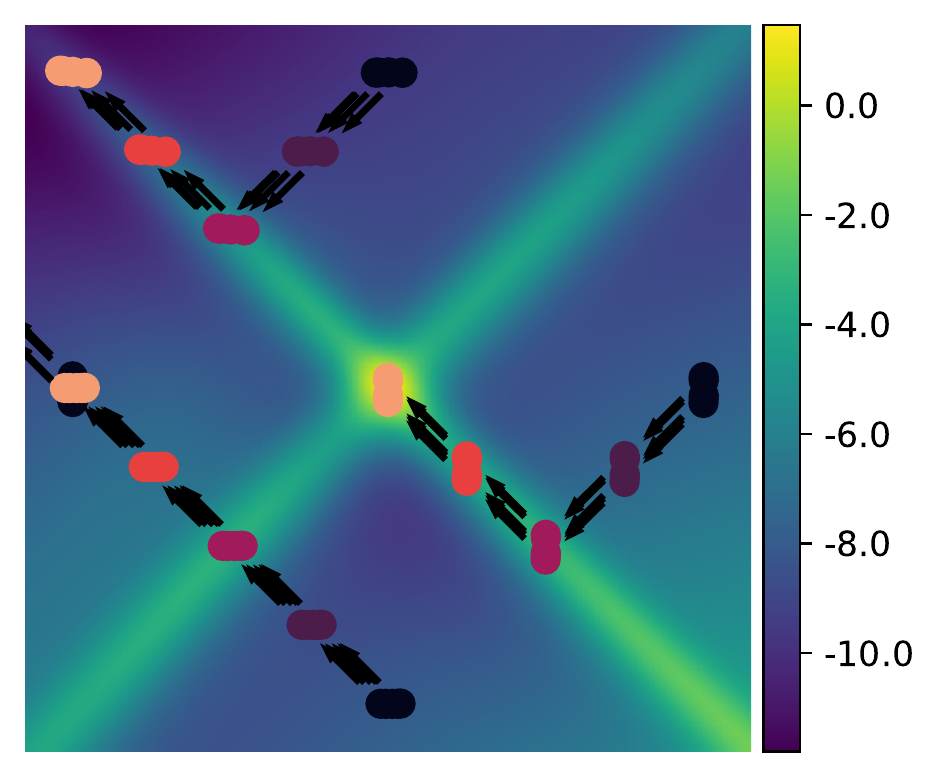}
    \caption{
      \small
      AIRL Test  \\
    }
    \label{fig:qual_results:eval_pm_airl}
  \end{subfigure}
  \vspace{-5pt}
  \caption{\small
    Results on the point mass navigation task. We show the learned reward functions of our method (b) vs. AIRL (c) and the policies learned from scratch using those reward functions (d, e). 
  }
  \vspace{-10pt}
  \label{fig:main_qual_results} 
\end{figure*}

%% file: sections/tables/pm.tex
\begin{tabular}{cccccc}
\toprule
\footnotesize
 & \textbf{\genmethod} & \textbf{AIRL} & \textbf{GCL} & \textbf{MaxEnt} & \textbf{f-IRL} \\
\midrule
  \textbf{Train} &  0.03 {\scriptsize $\pm$ 0.01 }  &  0.08 {\scriptsize $\pm$ 0.00 }  &  0.00 {\scriptsize $\pm$ 0.00 }  & NA & 0.07 {\scriptsize $\pm$ 0.01}\\
                                \textbf{Eval (Train)} &  0.03 {\scriptsize $\pm$ 0.00 }  &  0.08 {\scriptsize $\pm$ 0.02 }  &  2.07 {\scriptsize $\pm$ 0.09 }  &  0.30 {\scriptsize $\pm$ 0.40 } & 0.08 {\scriptsize $\pm$ 0.03} \\
                                \textbf{Eval (Test)} & \textbf{ 0.04 {\scriptsize $\pm$ 0.01 } } &  0.53 {\scriptsize $\pm$ 0.78 }  &  1.60 {\scriptsize $\pm$ 0.08 }  &  0.36 {\scriptsize $\pm$ 0.62 } & 1.04 {\scriptsize $\pm$ 0.15} \\
\bottomrule
\end{tabular}

%% file: sections/new_experiments.tex
\vspace{-10pt}
\section{Experiments}
\vspace{-5pt}
\label{sec:main_experiments} 

\begin{figure}[b] 
  \centering
   \begin{subfigure}[t]{0.23\textwidth}
    \includegraphics[width=\textwidth]{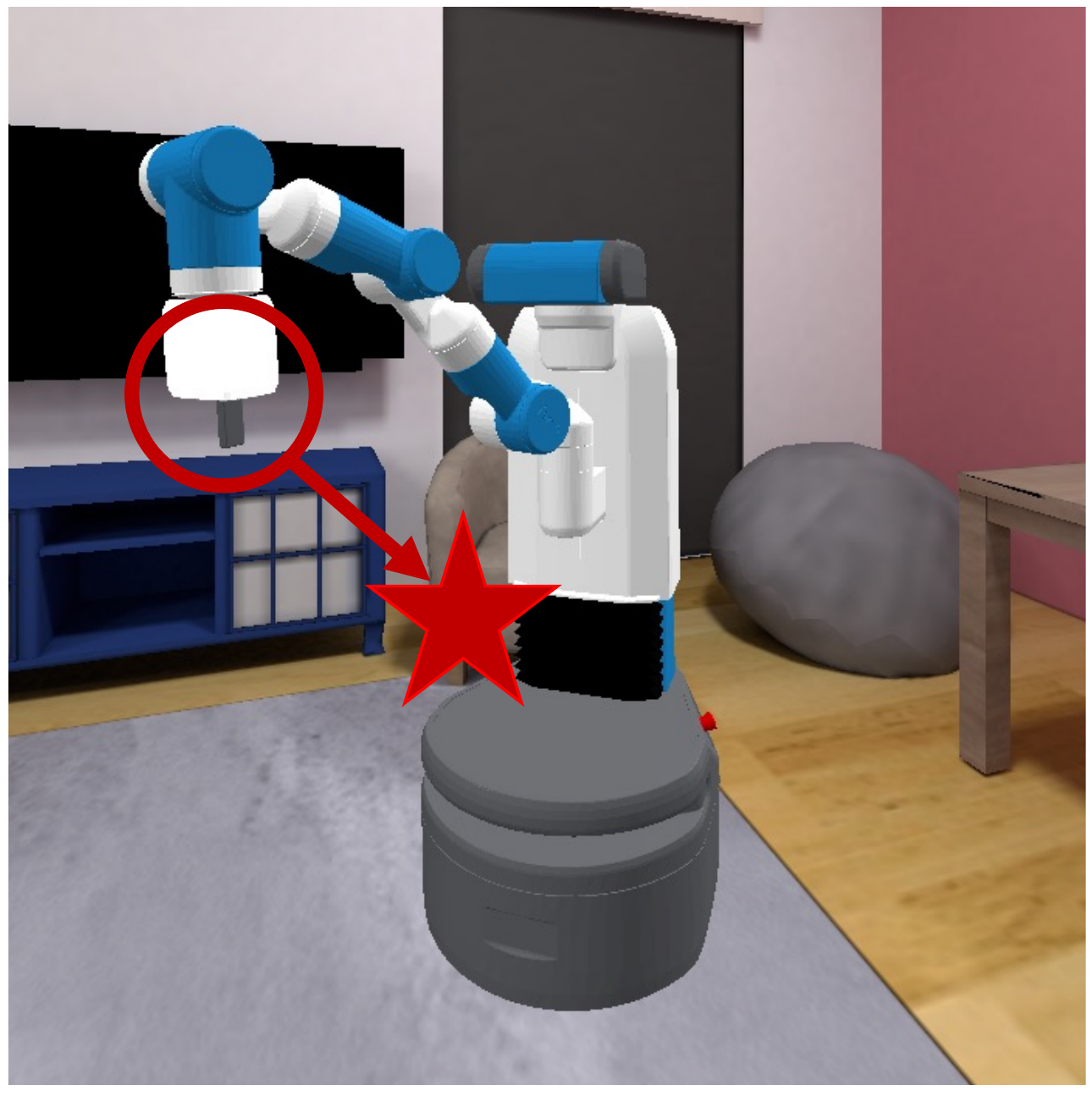}
    \caption{\centering Habitat: Reach Task}
    \label{fig:hab-eval:pick-task} 
  \end{subfigure} \quad 
   \begin{subfigure}[t]{0.23\textwidth}
    \includegraphics[width=\textwidth]{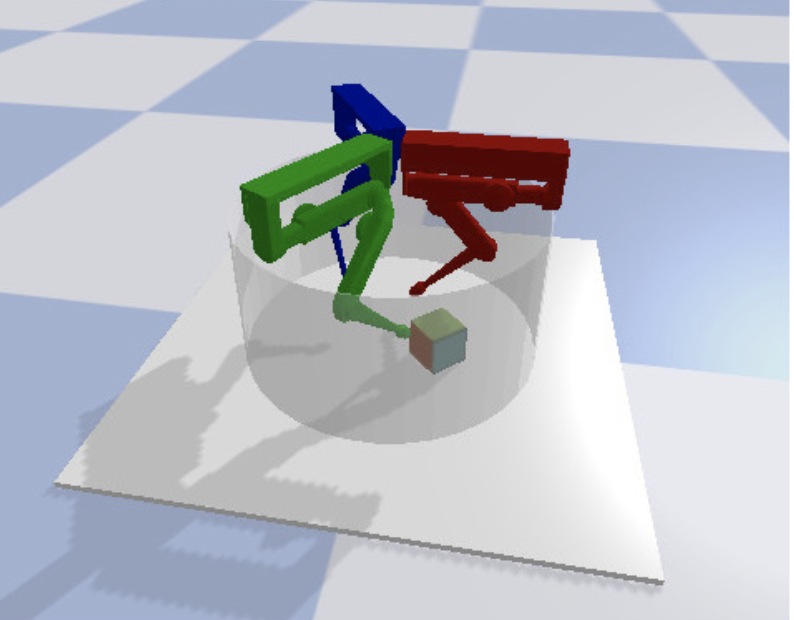}
    \caption{\centering TriFinger: Reach Task}
    \label{fig:trf-eval:reach-task} 
  \end{subfigure} \quad
  \begin{subfigure}[t]{0.15\textwidth}
    \includegraphics[width=\textwidth]{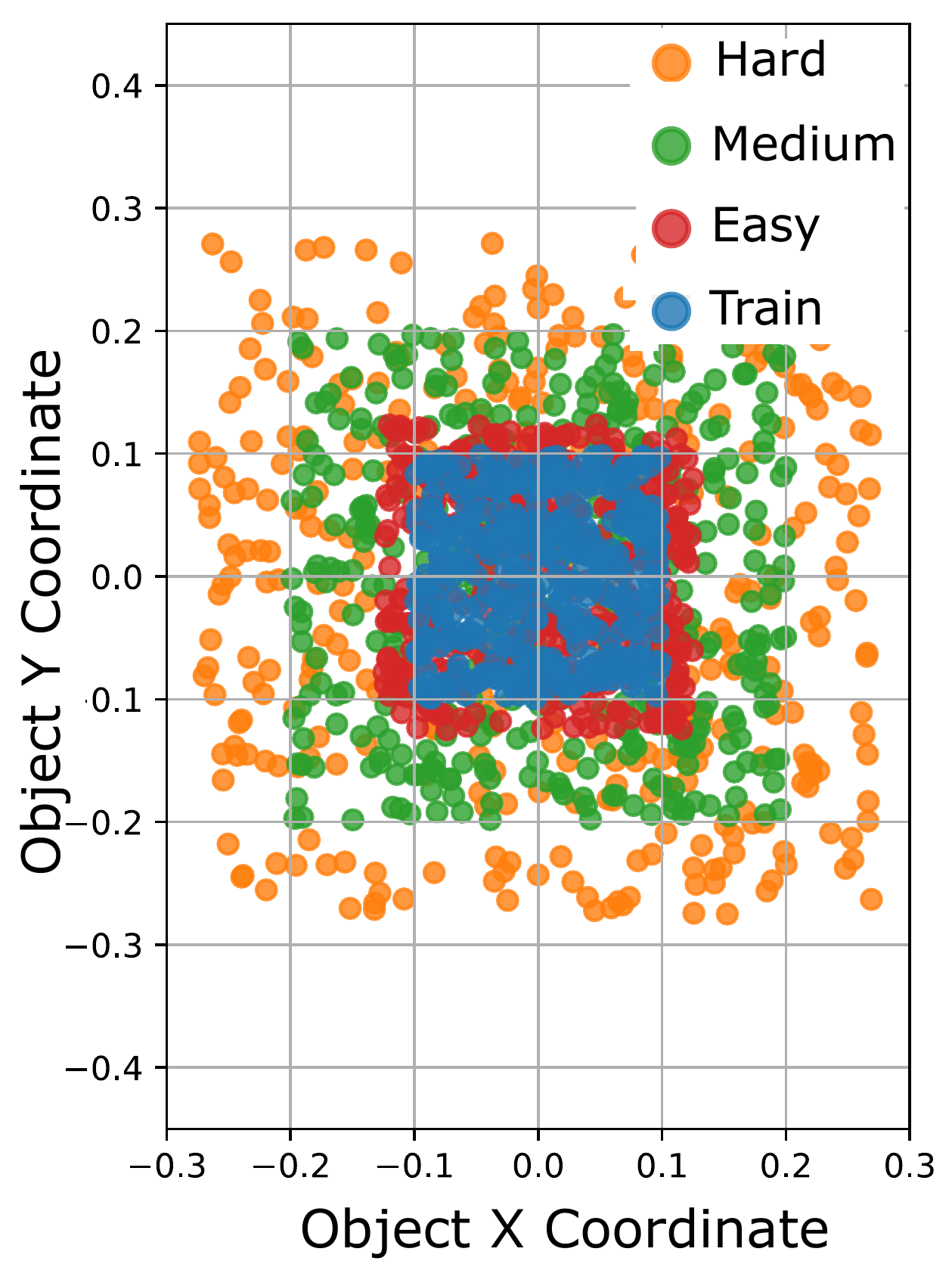}
    \caption{\small \centering Test Distribution}
    \label{fig:hab-eval:gen-type-1} 
  \end{subfigure}\quad
  \begin{subfigure}[t]{0.24\textwidth}
    \includegraphics[width=\textwidth]{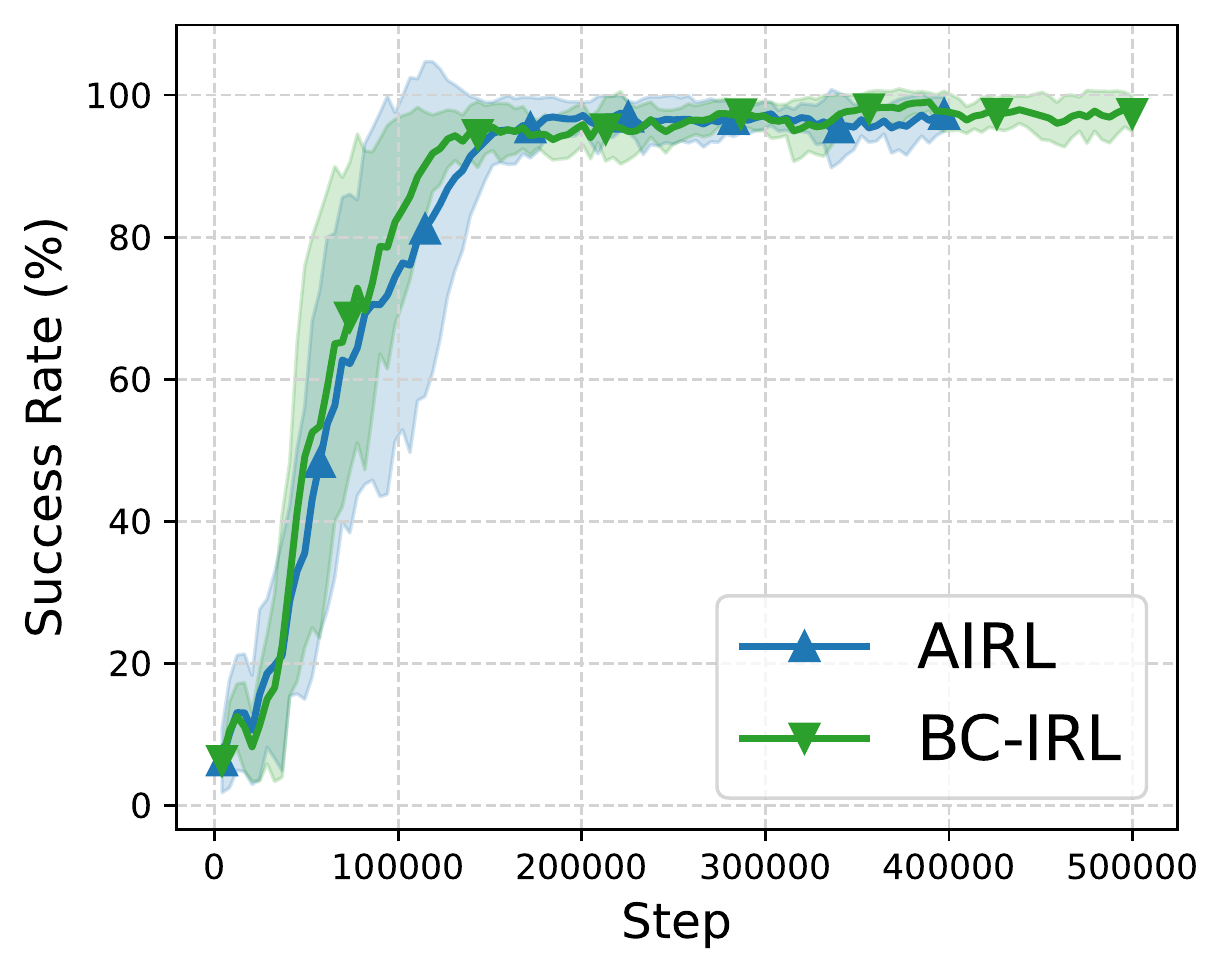}
    \caption{IRL Training}
    \label{fig:rl_curves:train} 
  \end{subfigure}
  \caption{\small
    (a+b) Visualization of the Fetch and TriFinger reach tasks. 
    c) 2D cross-section of the end-effector goal sampling regions in the reaching task. 
    The reward function is trained on goals from the blue points; the learned reward must train policies to accomplish goals from Easy, Medium, and Hard test distributions of orange, green, and red points. 
    d) Training curves during reward learning on Habitat Task, all methods succeed in training.
  }
  \label{fig:hab-overview}
\end{figure}

In our experiments, we aim to answer the following questions: 
(1) Can \genmethod learn reward functions that can train policies from scratch? 
(2) Does \genmethod learn rewards that can generalize to unseen states and goals better than IRL baselines in complex environments?
(3) Can learned rewards transfer better than policies learned directly with imitation learning? 
We show the first in \Cref{sec:exps:train_reward} and the next two in \Cref{sec:exps:eval_reward}.
We evaluate on two continuous control tasks: 1) Fetch reaching task \cite{szot2021habitat} (Fig~\ref{fig:hab-eval:pick-task}), and the TriFinger reaching task \cite{ahmed2021causalworld} (Fig~\ref{fig:trf-eval:reach-task}).

\subsection{Reward Training phase: Learning Rewards to Match the Expert}
\label{sec:exps:train_reward} 

\paragraph{Experimental Setup and Evaluation Metrics}

In the Fetch reaching task, setup in the Habitat 2.0 simulator \cite{szot2021habitat}, the robot must move its end-effector to a 3D goal location $g$ which changes between episodes.
The action space of the agent is the desired velocities for each of the 7 joints on the robot arm.
The robot succeeds if the end-effector is within 0.1m of the target position by the $20$ time step maximum episode length.
During reward learning, the goal $g$ is sampled from a $0.2$ meter length unit cube in front of the robot, $g \sim \mathcal{U}([0]^3, [0.2]^3)$.
We provide 100 demonstrations.

\begin{wraptable}{L}{0.6\textwidth}
\vspace{-10pt}
\begin{minipage}{0.6\textwidth}
  \resizebox{1.0\textwidth}{!}{
    \input{sections/tables/exp_train_smaller.tex}
  }
 \caption{\small
     Success rates for Fetch Reach and distance to goal for Trifinger Reach tasks in training policies to achieve the goal in the same start state and goal distributions as the expert demonstrations.
     Averages and standard deviations are from 3 seeds on Fetch Reach, and 5 seeds on Trifinger Reach with 100 episodes per seed.
  }
  \label{table:rewards_train}
\end{minipage}
\vspace{-10pt}
\end{wraptable}

For the Trifinger reaching task, each finger must move its fingertip to a 3D goal position.
The fingers must travel a different distance and avoid getting blocked by another finger. 
Each finger has 3 joints, creating a 9D action and state space. 
The fingers are joint position controlled. We use a time horizon of $T=5$ time steps.
We provide a single demonstration.
We report the final  distance to the demonstrated goal, $(g-g^\text{demo})^2$ in meters.

\paragraph{Evaluation and Baselines}
We evaluate \ppomethod by how well the reward it can train new policies from scratch in the same start state and goal distribution as the demonstrations.
Given the pointmass results \Cref{sec:reward_analysis}, we compare \ppomethod to AIRL,  the best performing baseline for reward learning. 
More details on baseline choice, policy and reward representation, and hyperparameters are described in the Appendix (\ref{app:reach-details}).

\paragraph{Results and Analysis} 
As \Cref{table:rewards_train} confirms, our method and baselines are able to imitate the demonstrations when policies are evaluated in the same task setting as the expert.
All methods are able to achieve a near 100\% success rate and low distance to goal.
Methods also learn with similar sample efficiency as shown in the learning curves in \Cref{fig:rl_curves:train}.
These high-success rates indicate \ppomethod and AIRL learn rewards that capture the expert behavior and train policies to mimic the expert.
When training policies in the same state/goal distribution as the expert, rewards from \ppomethod follow any constraints followed by the experts, just like the IRL baselines.

\subsection{Test Phase: Evaluating Reward and Policy Generalization}
\label{sec:exps:eval_reward} 
In this section, we evaluate how learned rewards and policies can generalize to new task settings with increased starting state and goal sampling noise.
We evaluate the generalization ability of rewards by evaluating how well they can train new policies to reach the goal in new start and goal distributions not seen in the demonstrations.
This evaluation captures the reality that it is infeasible to collect demonstrations for every possible start/goal configuration.
We thus aim to learn rewards from demonstrations that can generalize beyond the start/goal configurations present in those demonstrations.
We quantify reward generalization ability by whether the reward can train a policy to perform the task in the new start/goal configurations.

For the Fetch Reach task, we evaluate on three wider test goal sampling distributions $g \sim \mathcal{U}([0]^3, [g_{\text{max}}]^3)$: \useeasy, \usemedium, and \usehard, all visualized in \Cref{fig:hab-eval:gen-type-1}.
Similarly, we evaluate on new state regions, which increase the starting and goal initial state distributions but exclude the regions from training, exposing the reward to only unseen initial states and goals. 
In Trifinger, we sample start configurations from around the start joint position in the demonstrations, with increasingly wider distributions ($s_0 \sim \mathcal{N}(s_0^\text{demo}, \delta)$, with $\delta =0.01, 0.03, 0.05)$. 

\begin{table*}[t!]
  \centering
  \resizebox{1\textwidth}{!}{
    \input{sections/tables/reach_eval.tex}
  }
  \vspace{-5pt}
  \caption{\small
  Success rates for the reaching task comparing the generalization capabilities of IRL and imitation learning methods. 
  ``(\Scratch)" transfers the learned reward from IRL methods and trains a newly initialized policy in the test setting.
  ``(\Zero)" transfers the policy without training in the new setting.
  The Easy, Medium, and Hard indicate the difficulty of generalization where the end-effector goal is sampled from $g \sim \mathcal{U}([0]^3, [g_{\text{max}}]^3)$.
  }
  \vspace{-15pt}
  \label{table:reach_eval} 
\end{table*}

We evaluate reward function performance by how well the reward function can train new policies from scratch.
However, now the reward must generalize to inferring rewards in the new start state and goal distributions.
We additionally compare to two imitation learning baselines: Generative Adversarial Imitation Learning (GAIL) \cite{ho2016generative} and Behavior Cloning (BC).
We compare different methods of transferring the learned reward and policy to the test setting:

\textbf{1) Reward}: Transfer only the reward from the above training phase and train a newly initialized policy in the test setting.

\looseness=-1
\textbf{2) Policy}: Transfer only the policy from the above training phase and immediately evaluate the policy without further training in the test setting. This compares transferring learned rewards and transferring learned policies. We use this transfer strategy to compare against direct imitation learning methods. 

\textbf{3) Reward+Policy}: Transfer the reward and policy and then fine-tune the policy using the learned reward in the test setting. Results for this setting are in \Cref{sec:adapt}.

\paragraph{Results and Analysis} 
The results in \Cref{table:reach_eval} show \ppomethod learns rewards that generalize better than IRL baselines to new settings.
In the hardest generalization setting, \ppomethod achieves over twice the success rate of AIRL.
AIRL struggles to transfer its learned reward to harder generalization settings, with performance decreasing as the goal sampling distribution becomes larger and has less overlap with the training goal distribution.
In the ``Hard" start region generalization setting, the performance of AIRL degrades to 34\% success rate.
On the other hand, \ppomethod learns a generalizable reward and performs well even in the ``Hard" generalization strategy, achieving 76\% success.
This trend is true both for generalization to new start state distributions and for new start state regions.
The results for Trifinger Reach in \Cref{table:trf_quant_results} support these findings with rewards learned via \ppomethod generalizing better to training policies from scratch in all three test distributions. 
All training curves for training policies from scratch with learned rewards are in \Cref{sec:rl_training}.

\begin{wraptable}{R}{0.42\textwidth}
\begin{minipage}{0.42\textwidth}
\vspace{-15pt}
\resizebox{1\textwidth}{!}{
  \begin{tabular}{lcc}
    \toprule
    \footnotesize
& \textbf{\ppomethod} & \textbf{AIRL} \\
\midrule
    \textbf{Test $\delta=0.01$} &  \textbf{0.0065 {\scriptsize $\pm$ 0.002 }}  &  0.012 {\scriptsize $\pm$ 0.0017 }  \\
    \textbf{Test $\delta=0.03$} &  \textbf{0.0061 {\scriptsize $\pm$ 0.002 }}  &  0.012 {\scriptsize $\pm$ 0.0008 }  \\
    \textbf{Test $\delta=0.05$} &  \textbf{0.0061 {\scriptsize $\pm$ 0.001 }}  &  0.0117 {\scriptsize $\pm$ 0.0015 }  \\
    \bottomrule
  \end{tabular}
}
 \caption{\small
   Distance to the goal for Trifinger reach, evaluating how rewards generalize to training policies in new start/ goal distributions.
  }
  \label{table:trf_quant_results}
  \vspace{-15pt}
\end{minipage}
\end{wraptable}

\looseness=-1
Furthermore, the results in \Cref{table:reach_eval} also demonstrate that transferring rewards ``(Reward)" is more effective for generalization than transferring policies ``(Policy)". 
Transferring the reward to train new policies typically outperforms transferring only the policy for all IRL approaches.
Additionally, training from scratch with rewards learned via IRL outperforms non-reward learning imitation learning methods that only permit transferring the policy zero-shot. 
The policies learned by GAIL and BC generalize worse than training new policies from scratch with the reward learned by \ppomethod, with BC and GAIL achieving 35\% and 37\% success rates in the ``Hard" generalization setting while our method achieves 76\% success.
The superior performance of \ppomethod over BC highlights the important differences between the two methods with our method learning a reward and training the policy with PPO on the learned reward.

In \Cref{sec:adapt}, we also show the \adapt transfer setting and demonstrate \ppomethod also outperforms baselines in this setting.
In \Cref{sec:supp:reach} we also analyze performance with the number of demos, different inner and outer loop learning rates, and number of inner loop updates.

%% file: sections/tables/exp_train_smaller.tex
\begin{tabular}{ccccccc}
\toprule
 & \textbf{\ppomethod} & \textbf{AIRL} \\
\midrule
  \textbf{Fetch Reach (Success) $ \uparrow$} & 1.00 {\small $\pm$ 0.00 }  &  0.96 {\small $\pm$ 0.00 } \\
  \textbf{Trifinger Reach (Goal Dist) $ \downarrow$} &  0.002 {\scriptsize $\pm$ 0.0015 }  &  0.007 {\scriptsize $\pm$ 0.0017 } \\
\bottomrule
\end{tabular}

%% file: sections/tables/reach_eval.tex
\begin{tabular}{ccccccc}
 & \textbf{\ppomethod} & \textbf{AIRL} & \textbf{\ppomethod} & \textbf{AIRL} & \textbf{BC} & \textbf{GAIL} \\
 &  (\Scratch) & (\Scratch) & (\Zero) & (\Zero) & (\Zero) & (\Zero)\\
\midrule
\textbf{Start Distrib: \useeasy} & \textbf{ 1.00 {\small $\pm$ 0.00 } } &  0.96 {\small $\pm$ 0.04 }  &  0.94 {\small $\pm$ 0.03 }  &  0.89 {\small $\pm$ 0.01 }  &  0.86 {\small $\pm$ 0.00 }  &  0.88 {\small $\pm$ 0.02 }  \\
\textbf{Start Distrib: \usemedium} & \textbf{ 1.00 {\small $\pm$ 0.00 } } &  0.94 {\small $\pm$ 0.03 }  &  0.70 {\small $\pm$ 0.10 }  &  0.57 {\small $\pm$ 0.03 }  &  0.56 {\small $\pm$ 0.01 }  &  0.58 {\small $\pm$ 0.02 }  \\
\textbf{Start Distrib: \usehard} & \textbf{ 0.76 {\small $\pm$ 0.16 } } &  0.34 {\small $\pm$ 0.01 }  &  0.48 {\small $\pm$ 0.06 }  &  0.38 {\small $\pm$ 0.04 }  &  0.35 {\small $\pm$ 0.01 }  &  0.37 {\small $\pm$ 0.04 }  \\
\hline
\textbf{State Region: \useeasy} & \textbf{ 1.00 {\small $\pm$ 0.00 } } &  0.98 {\small $\pm$ 0.02 }  &  0.87 {\small $\pm$ 0.09 }  &  0.74 {\small $\pm$ 0.02 }  &  0.73 {\small $\pm$ 0.00 }  &  0.76 {\small $\pm$ 0.04 }  \\
\textbf{State Region: \usemedium} & \textbf{ 1.00 {\small $\pm$ 0.00 } } &  0.88 {\small $\pm$ 0.10 }  &  0.70 {\small $\pm$ 0.06 }  &  0.58 {\small $\pm$ 0.03 }  &  0.52 {\small $\pm$ 0.01 }  &  0.54 {\small $\pm$ 0.01 }  \\
\textbf{State Region: \usehard} & \textbf{ 0.78 {\small $\pm$ 0.13 } } &  0.34 {\small $\pm$ 0.04 }  &  0.49 {\small $\pm$ 0.08 }  &  0.42 {\small $\pm$ 0.02 }  &  0.39 {\small $\pm$ 0.02 }  &  0.42 {\small $\pm$ 0.03 }  \\
\bottomrule
\end{tabular}

%% file: sections/discussion.tex
\vspace{-10pt}
\section{Discussion and Future Work}
\vspace{-5pt}
\label{sec:discussion} 

We propose a new IRL framework for learning generalizable rewards with bi-level gradient-based optimization. 
By meta-learning rewards, our framework can optimize alternative outer-level objectives instead of the maximum entropy objective commonly used in prior work.
We propose \ppomethod an instantiation of our new framework, which uses PPO for policy optimization in the inner loop and an action matching objective in the outer loop.
We demonstrate that \ppomethod learns rewards that generalize better than baselines.
Potential negative social impacts of this work are that learning reward functions from data could result in less interpretable rewards, leading to more opaque behaviors from agents that optimize the learned reward.

Future work will explore alternative instantiations of the \genmethod framework, such as utilizing sample efficient off-policy methods like SAC or model-based methods in the inner loop. 
Model-based methods are especially appealing because a single dynamics model could be shared between tasks and learning reward functions for new tasks could be achieved purely using the model. 
Finally, other outer loop objectives rather than action matching are also possible.

%% file: sections/acknowledgements.tex
\section{Acknowledgments}
The Georgia Tech effort was supported in part by NSF, ONR YIP, and ARO PECASE. The views and conclusions contained herein are those of the authors and should not be interpreted as necessarily representing the official policies or endorsements, either expressed or implied, of the U.S. Government, or any sponsor.

%% file: supp/pm_experiments.tex
\section{Further Point Mass Navigation Results}

\subsection{Qualitative Results for all Methods in Point Mass Navigation}
\label{sec:pm_all_qual} 

\begin{figure*}[t]
  \centering
  \begin{subfigure}[t]{0.24\textwidth}
    \includegraphics[width=\textwidth]{figures/exp/toy_task/mirl_reward.pdf}
    \caption{\genmethod Reward}
  \end{subfigure}
  \begin{subfigure}[t]{0.24\textwidth}
    \includegraphics[width=\textwidth]{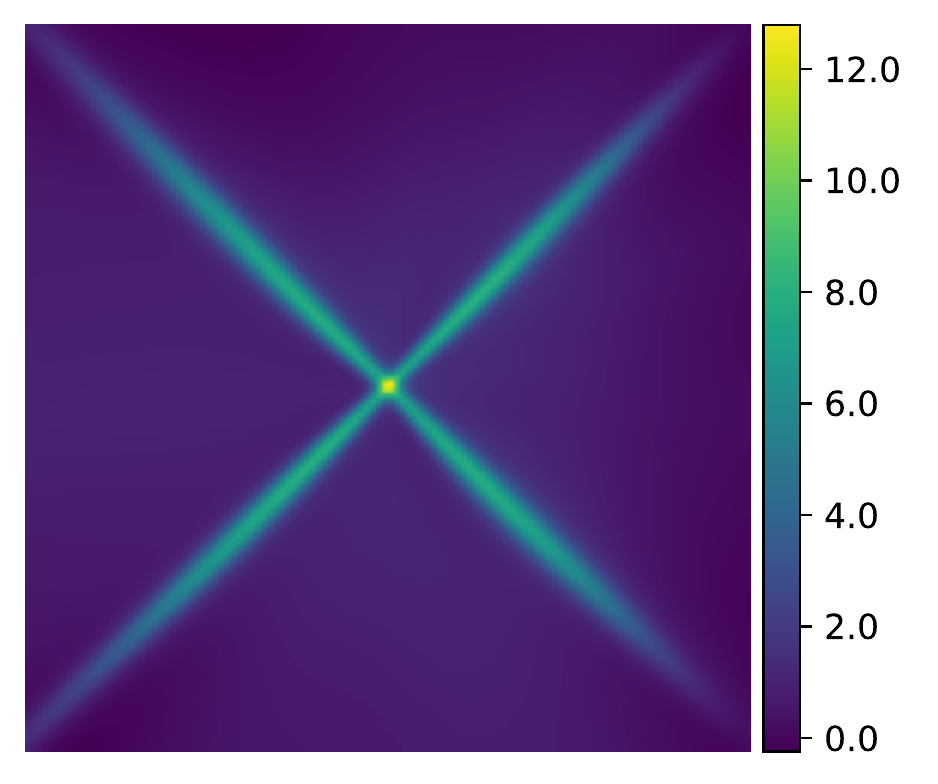}
    \caption{MaxEnt Reward}
  \end{subfigure}
  \begin{subfigure}[t]{0.24\textwidth}
    \includegraphics[width=\textwidth]{figures/exp/toy_task/airl_reward.pdf}
    \caption{AIRL Reward}
  \end{subfigure}
  \begin{subfigure}[t]{0.24\textwidth}
    \includegraphics[width=\textwidth]{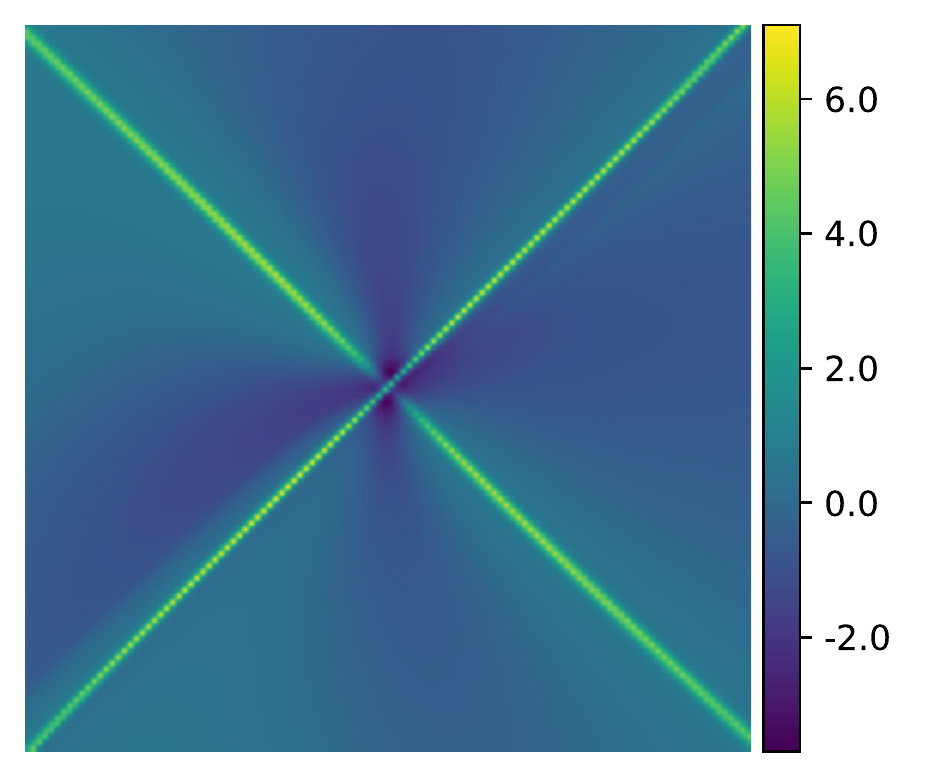}
    \caption{GCL Train}
  \end{subfigure}
  \begin{subfigure}[t]{0.24\textwidth}
    \includegraphics[width=\textwidth]{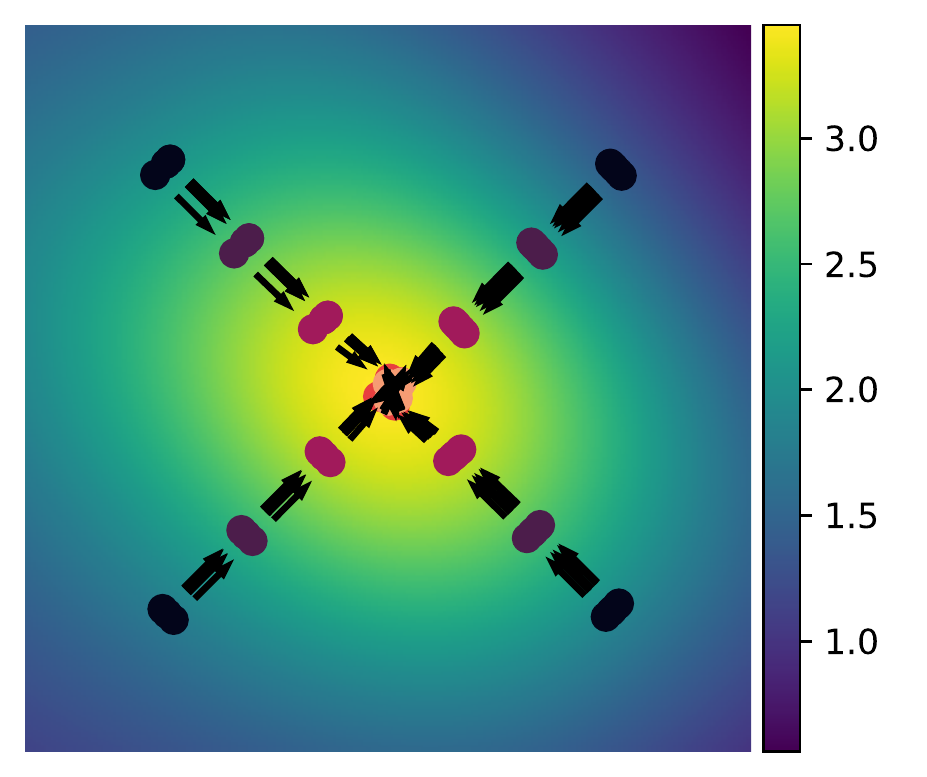}
    \caption{\genmethod Train}
  \end{subfigure}
  \begin{subfigure}[t]{0.24\textwidth}
    \includegraphics[width=\textwidth]{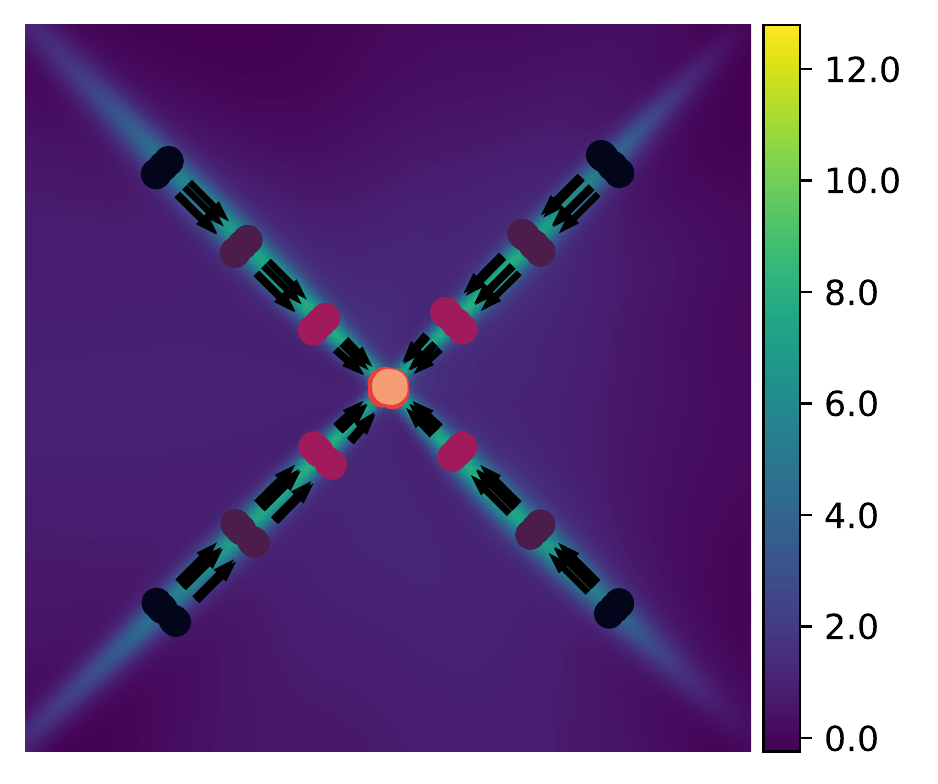}
    \caption{MaxEnt Train}
  \end{subfigure}
  \begin{subfigure}[t]{0.24\textwidth}
    \includegraphics[width=\textwidth]{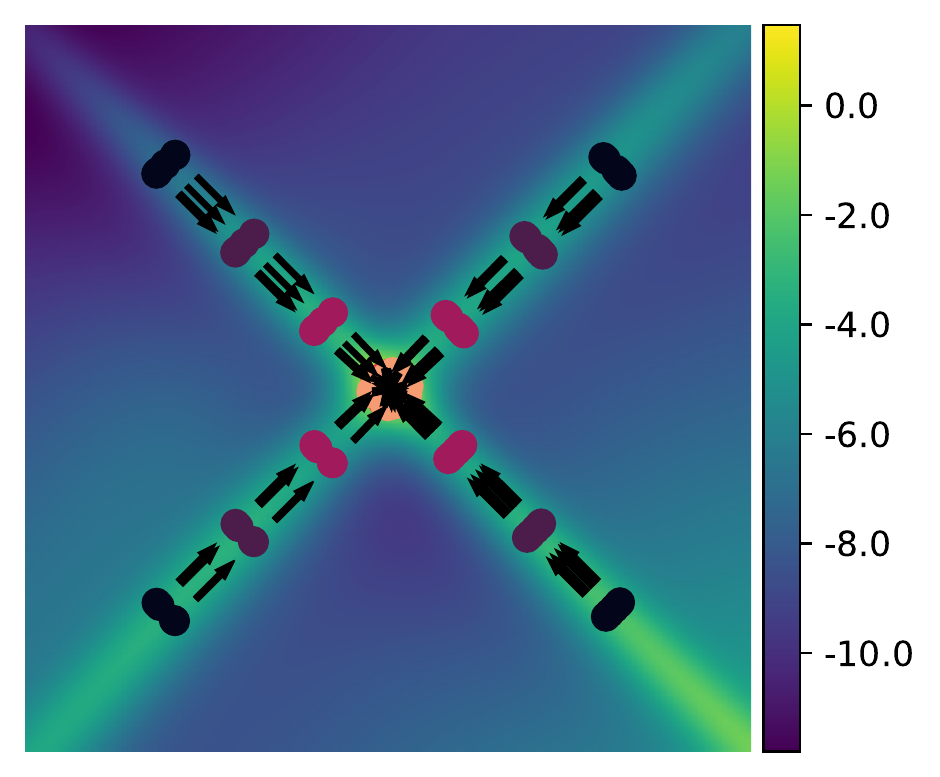}
    \caption{AIRL Train}
  \end{subfigure}
  \begin{subfigure}[t]{0.24\textwidth}
    \includegraphics[width=\textwidth]{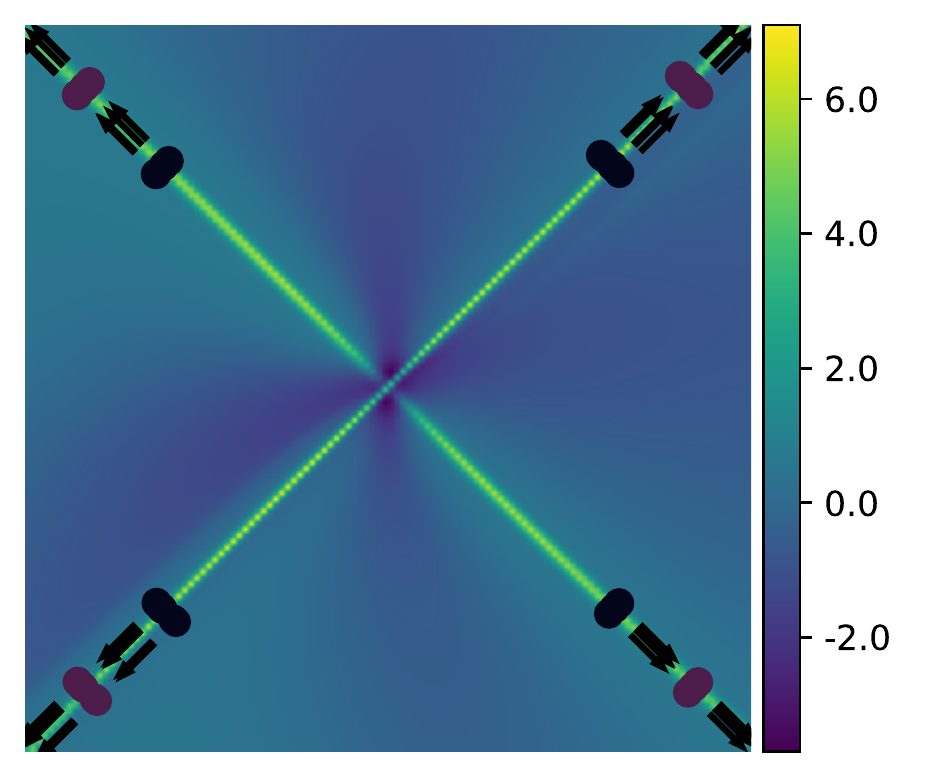}
    \caption{GCL Train}
  \end{subfigure}
  \begin{subfigure}[t]{0.24\textwidth}
    \includegraphics[width=\textwidth]{figures/exp/toy_task/mirl_eval_rollouts.pdf}
    \caption{\genmethod Test}
  \end{subfigure}
  \begin{subfigure}[t]{0.24\textwidth}
    \includegraphics[width=\textwidth]{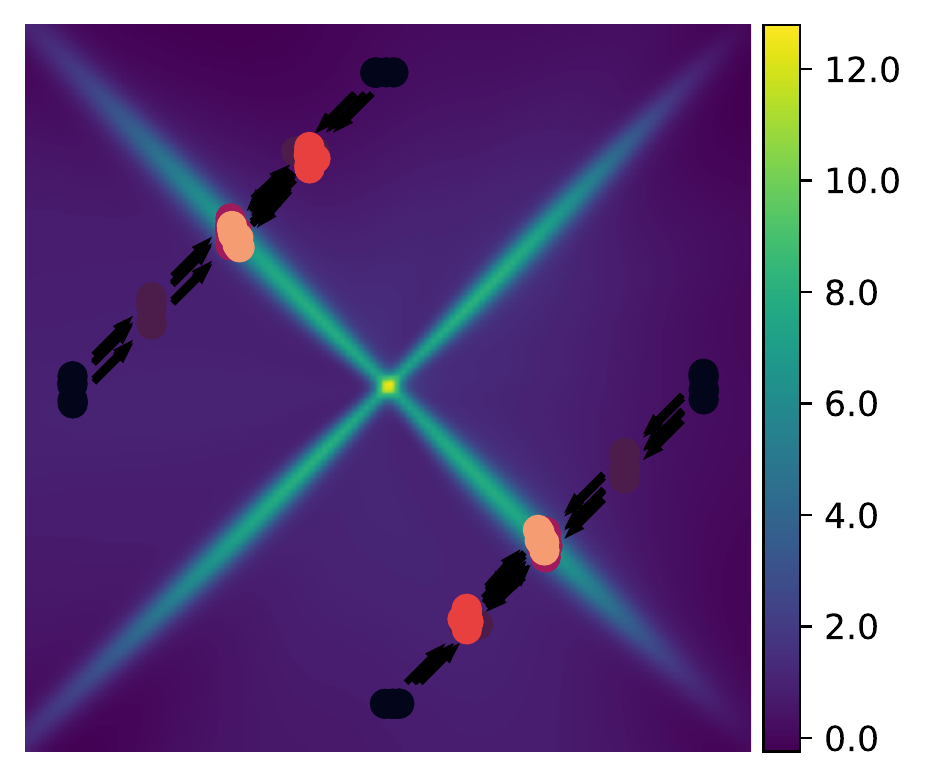}
    \caption{MaxEnt Test}
  \end{subfigure}
  \begin{subfigure}[t]{0.24\textwidth}
    \includegraphics[width=\textwidth]{figures/exp/toy_task/airl_eval_rollouts.pdf}
    \caption{AIRL Test}
  \end{subfigure}
  \begin{subfigure}[t]{0.24\textwidth}
    \includegraphics[width=\textwidth]{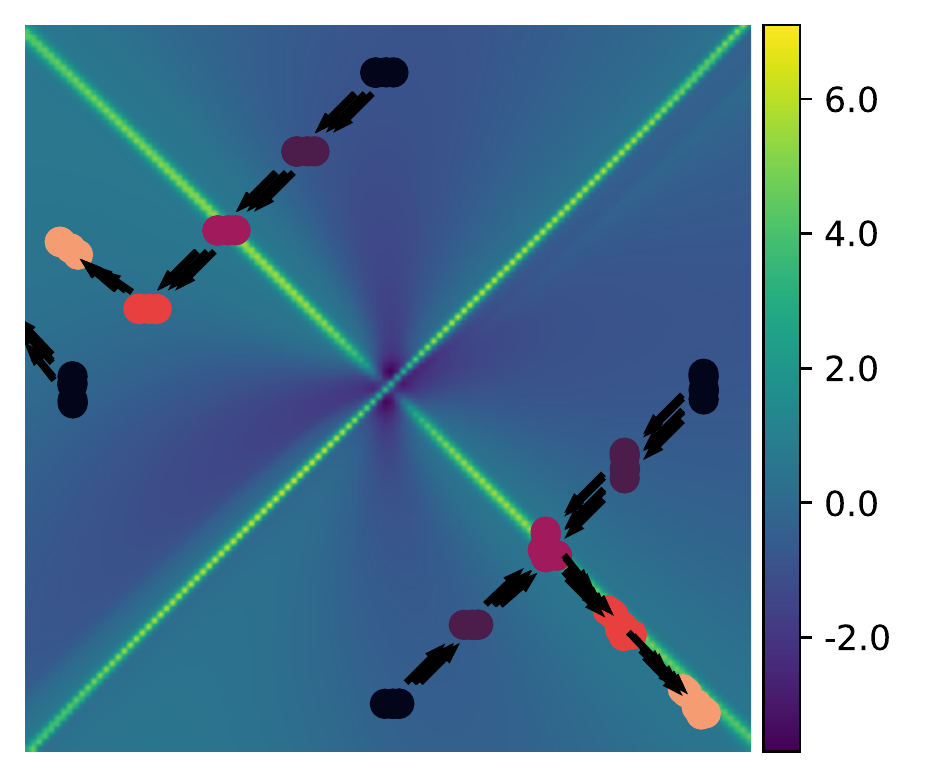}
    \caption{GCL Test}
  \end{subfigure}
  \caption{\small
    Qualitative results for all methods on the point mass navigation task without the obstacle.
  }
  \label{fig:all_qual_pm} 
\end{figure*}

Visualizations of the reward functions from all methods for the regular pointmass task are displayed in \Cref{fig:all_qual_pm}.

\subsection{Obstacle Point Mass Navigation}
\label{sec:pmo_nav}

\input{sections/figures/pmo}

The obstacle point mass navigation task incorporates asymmetric dynamics with an off-centered obstacle. 
This environment is the same as the point mass navigation task from \Cref{sec:reward_analysis}, except there is an obstacle blocking the path to the center and the agent only spawns in the top-right hand corner.
This task has a trajectory length of $T=50$ time steps with 100 demonstrations.
\Cref{fig:qual_results:pmo_demo} visualizes the expert demonstrations where darker points are earlier time steps.

\begin{table}
  \centering
 \resizebox{0.75\textwidth}{!}{
  \input{sections/tables/pmo}
 }
  \caption{\small
    Distance to the goal for the point mass navigation task where numbers are mean and standard error for 3 seeds and 100 evaluation episodes per seed.  
    \pmtrain is policy trained during reward learning. 
    MaxEnt does not learn a policy during reward learning thus its performance is ``NA".
    \pmevaltrain uses the learned reward to train a policy from scratch on the same distribution used to train the reward. 
    \pmevaltest  measures the ability of the learned reward to generalize to a new starting state distribution. 
  }
  \label{table:pmo_quant_results}
  \vspace{-15pt}
\end{table}

\begin{figure*}[t]
  \centering
  \begin{subfigure}[t]{0.24\textwidth}
    \includegraphics[width=\textwidth]{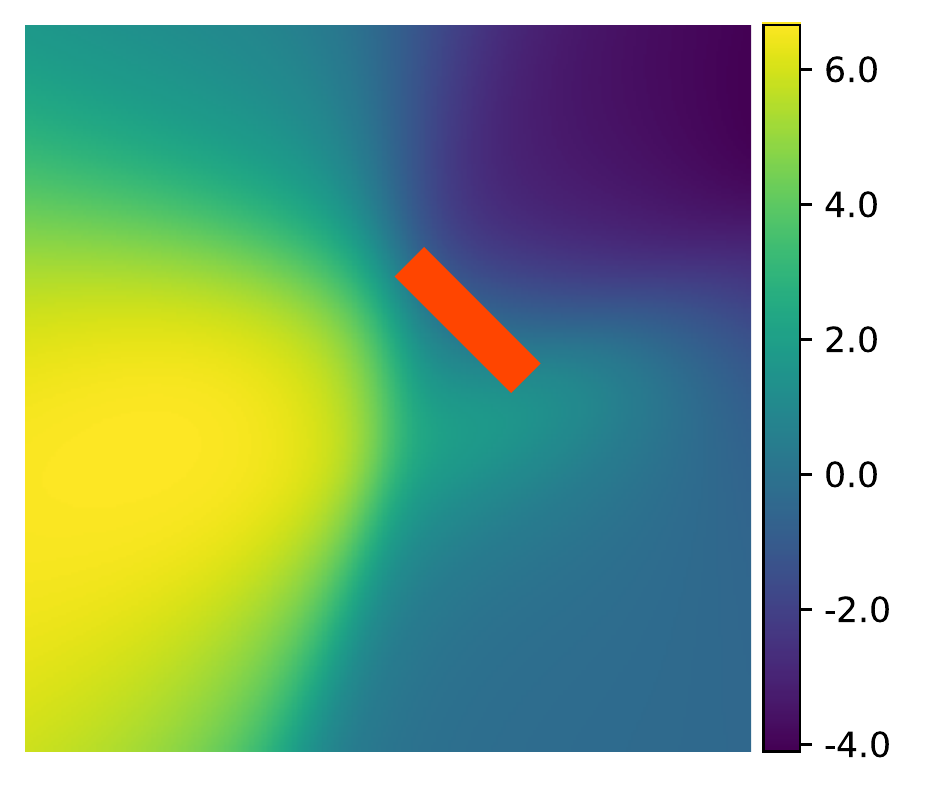}
    \caption{\genmethod Reward}
  \end{subfigure}
  \begin{subfigure}[t]{0.24\textwidth}
    \includegraphics[width=\textwidth]{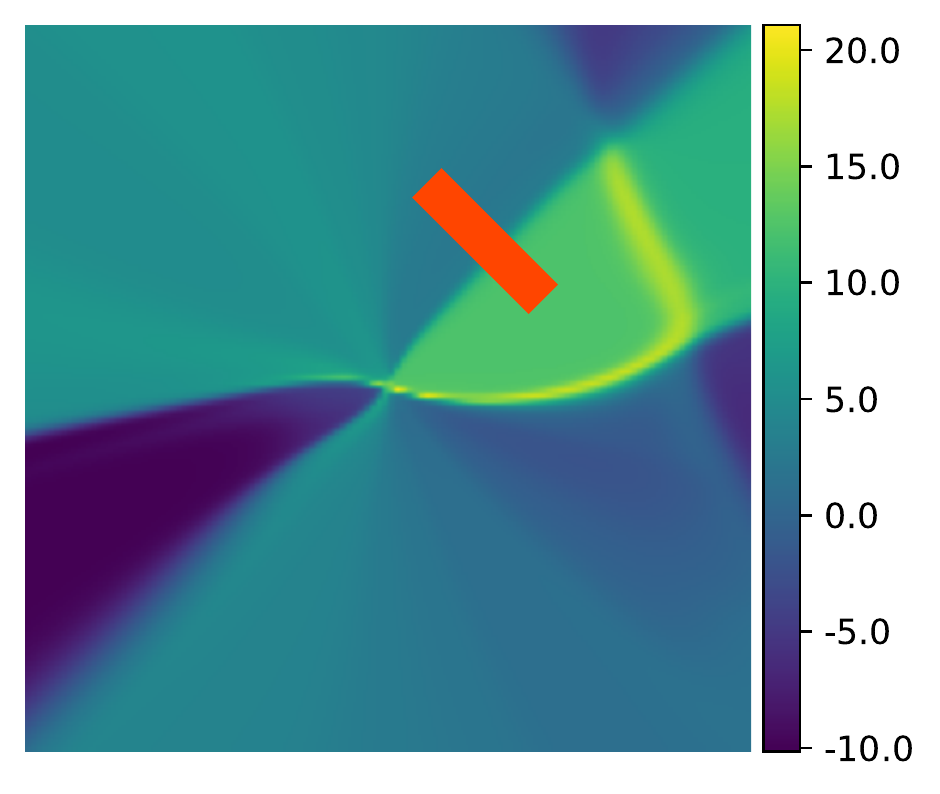}
    \caption{MaxEnt Reward}
  \end{subfigure}
  \begin{subfigure}[t]{0.24\textwidth}
    \includegraphics[width=\textwidth]{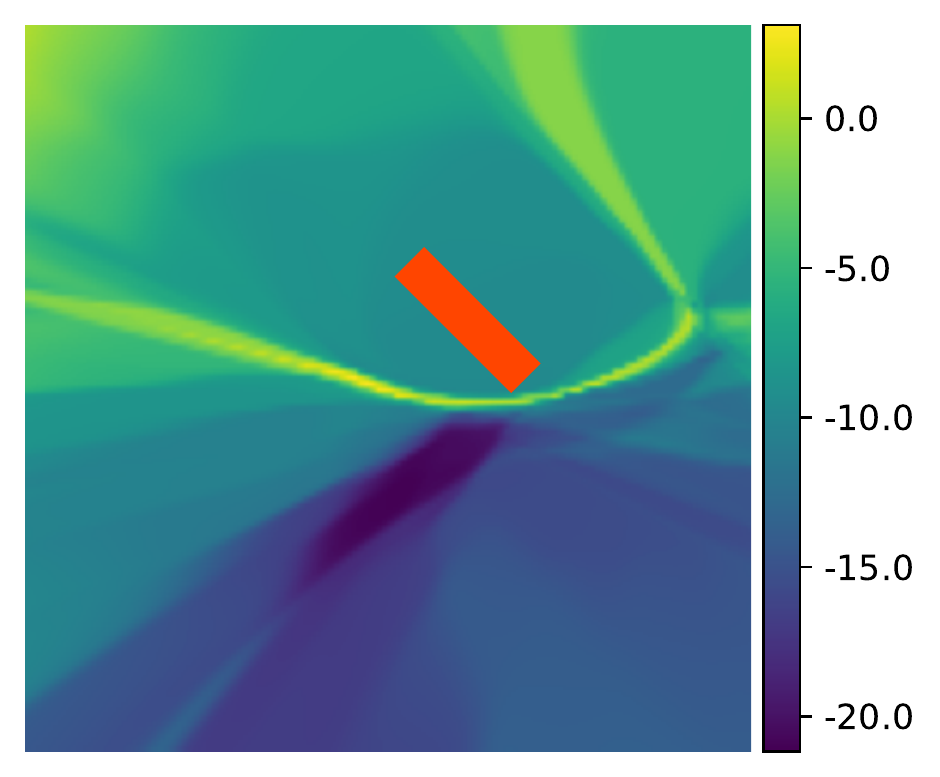}
    \caption{AIRL Reward}
  \end{subfigure}
  \begin{subfigure}[t]{0.24\textwidth}
    \includegraphics[width=\textwidth]{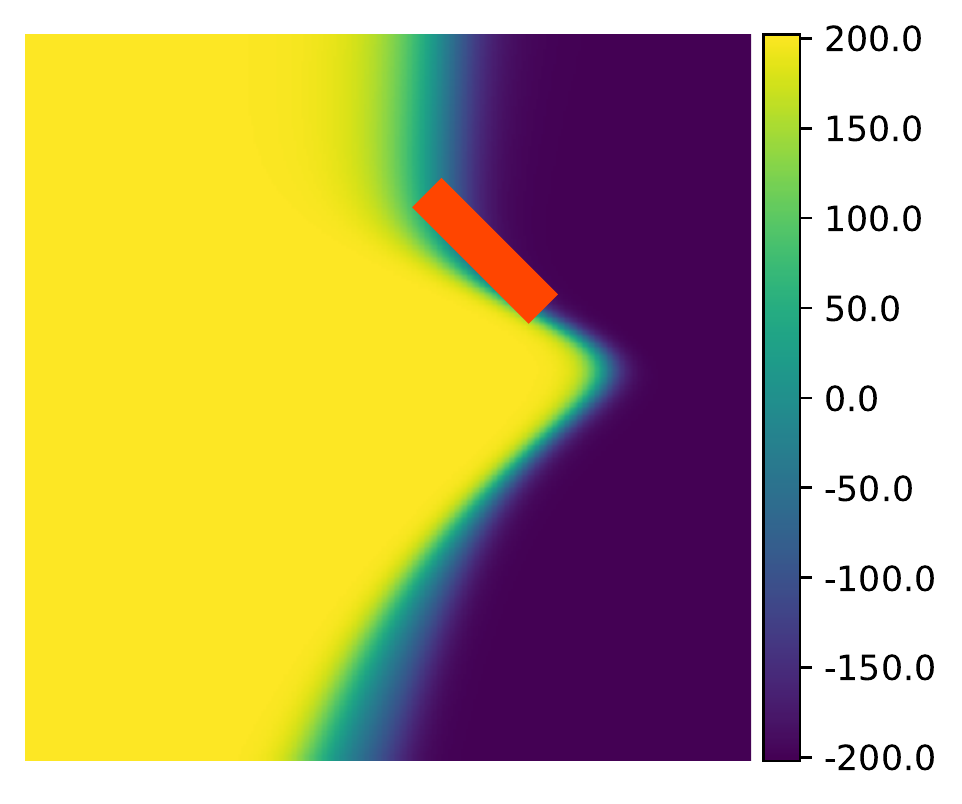}
    \caption{GCL Train}
  \end{subfigure}
  \begin{subfigure}[t]{0.24\textwidth}
    \includegraphics[width=\textwidth]{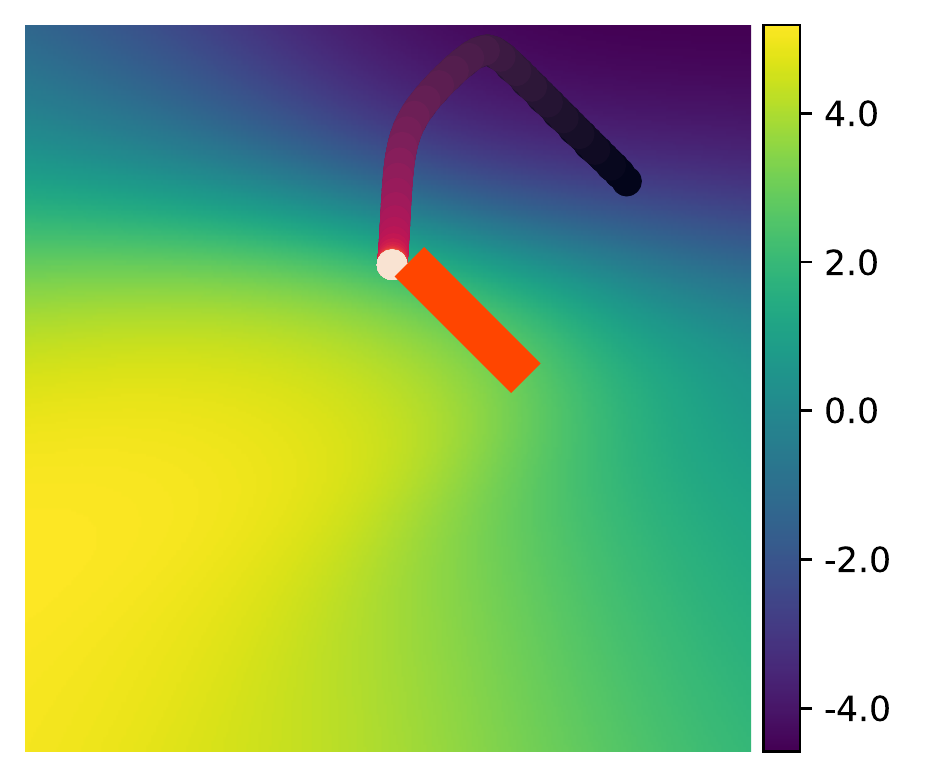}
    \caption{\genmethod Train}
  \end{subfigure}
  \begin{subfigure}[t]{0.24\textwidth}
    \includegraphics[width=\textwidth]{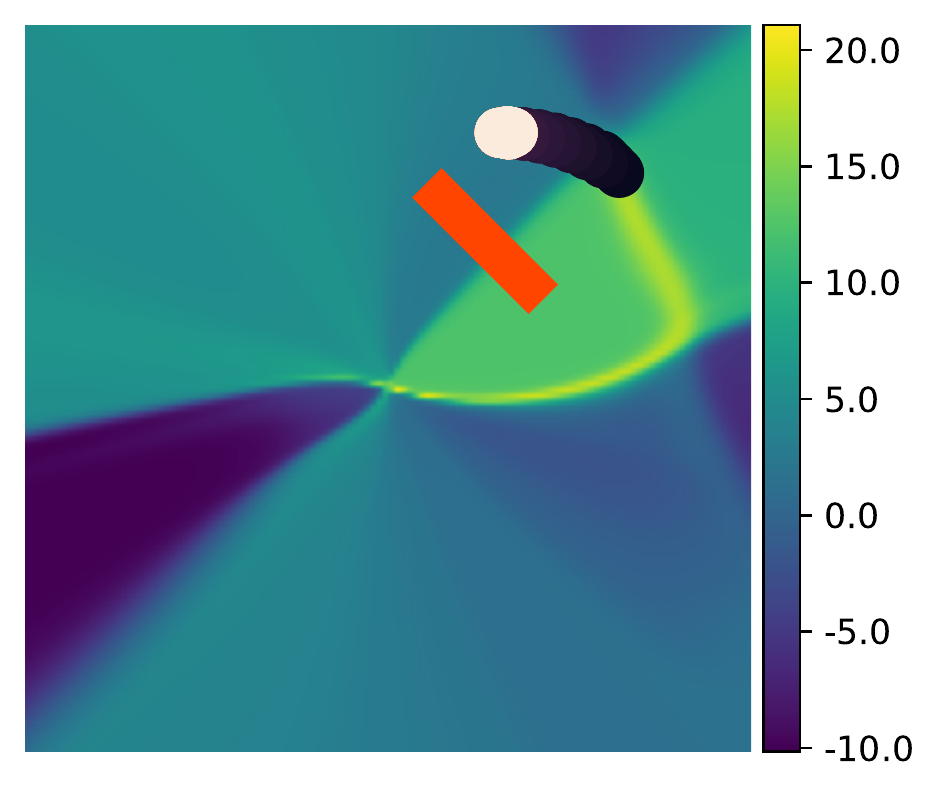}
    \caption{MaxEnt Train}
  \end{subfigure}
  \begin{subfigure}[t]{0.24\textwidth}
    \includegraphics[width=\textwidth]{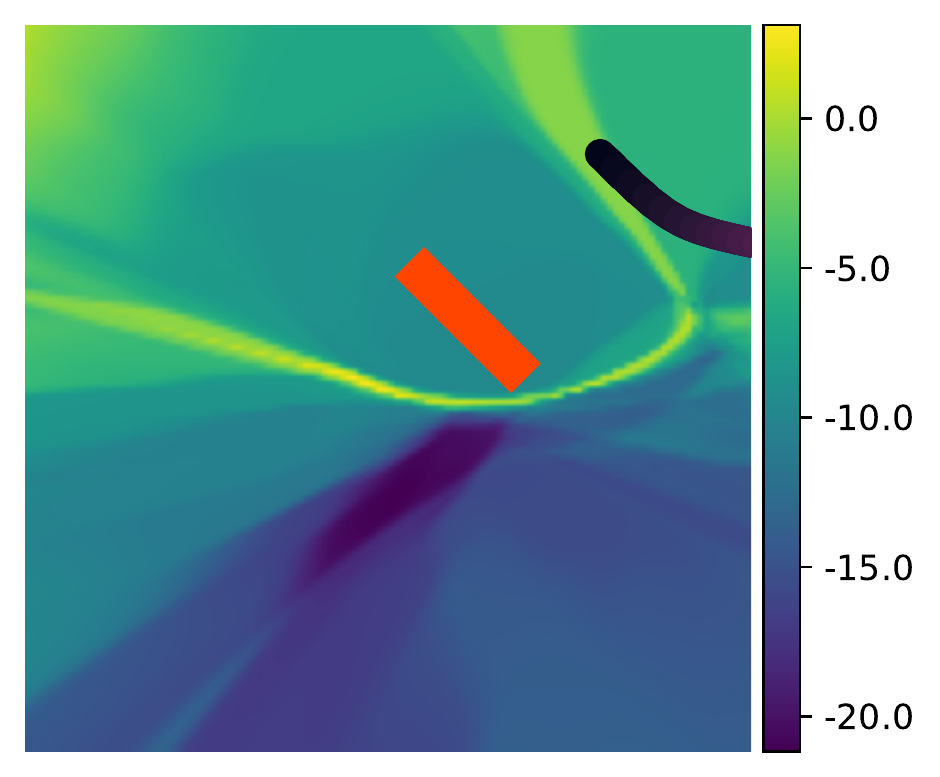}
    \caption{AIRL Train}
  \end{subfigure}
  \begin{subfigure}[t]{0.24\textwidth}
    \includegraphics[width=\textwidth]{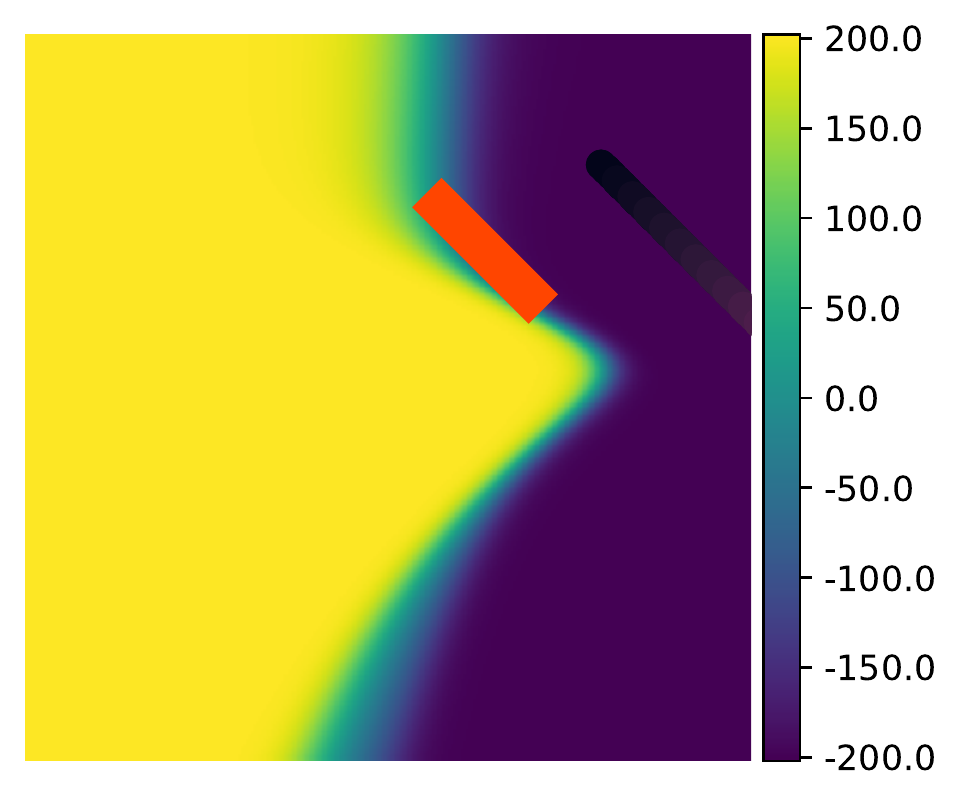}
    \caption{GCL Train}
  \end{subfigure}
  \begin{subfigure}[t]{0.24\textwidth}
    \includegraphics[width=\textwidth]{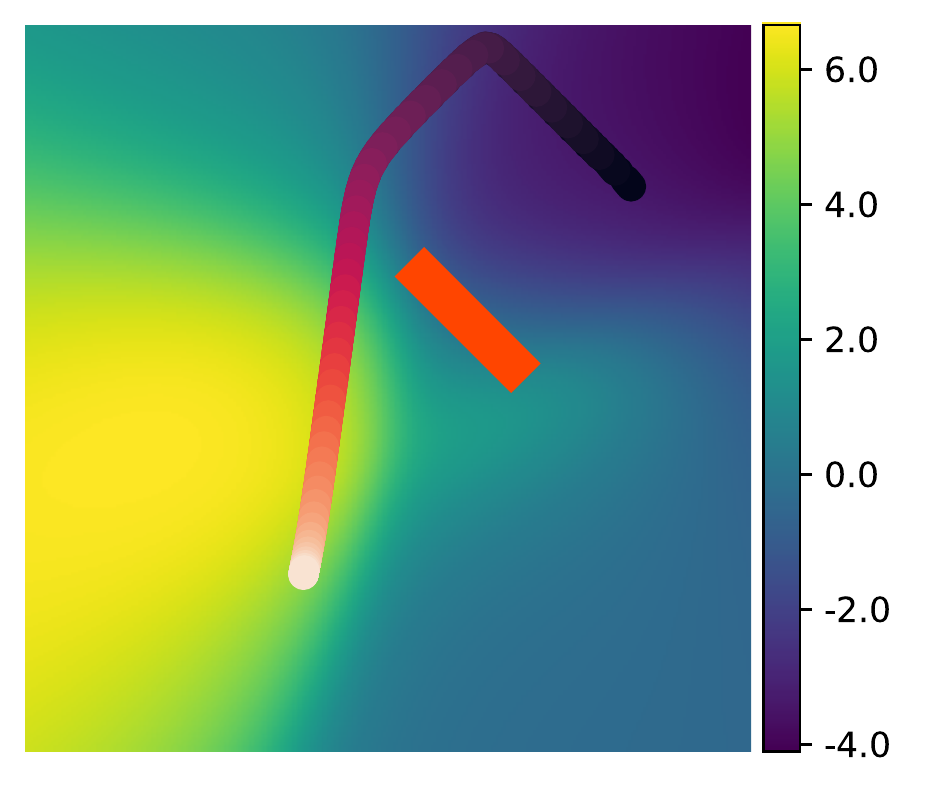}
    \caption{\genmethod Test}
  \end{subfigure}
  \begin{subfigure}[t]{0.24\textwidth}
    \includegraphics[width=\textwidth]{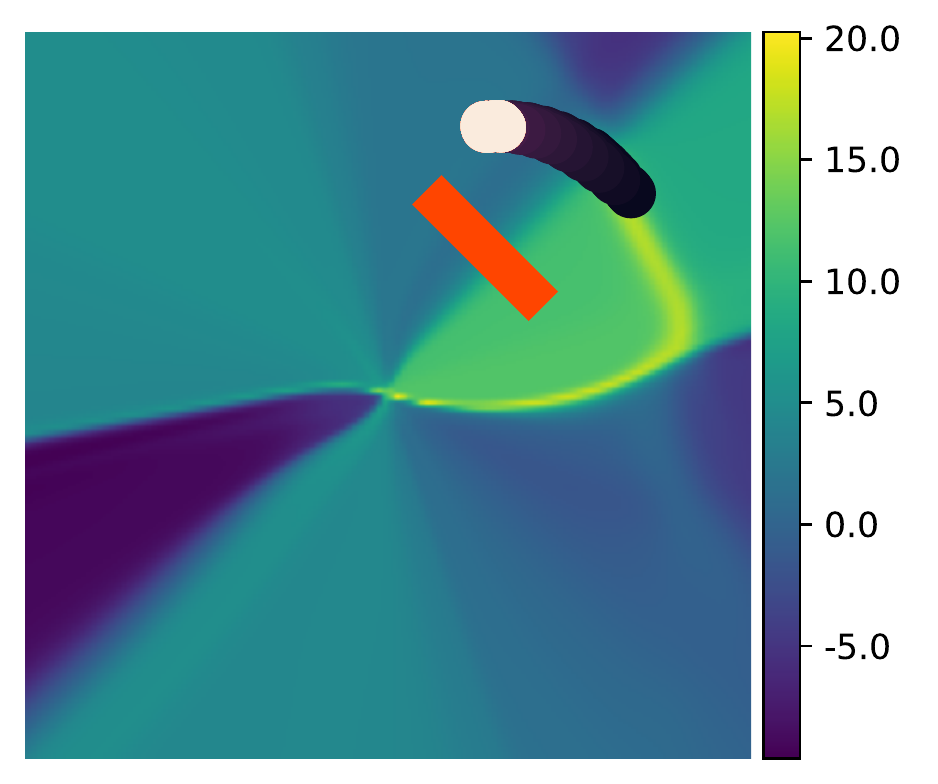}
    \caption{MaxEnt Test}
  \end{subfigure}
  \begin{subfigure}[t]{0.24\textwidth}
    \includegraphics[width=\textwidth]{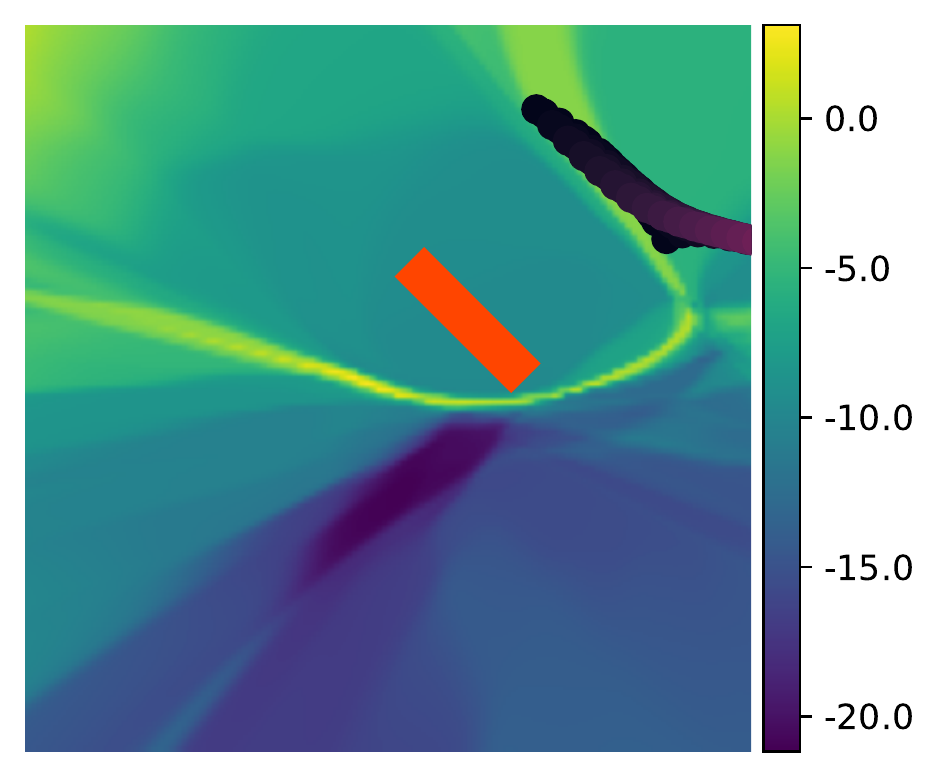}
    \caption{AIRL Test}
  \end{subfigure}
  \begin{subfigure}[t]{0.24\textwidth}
    \includegraphics[width=\textwidth]{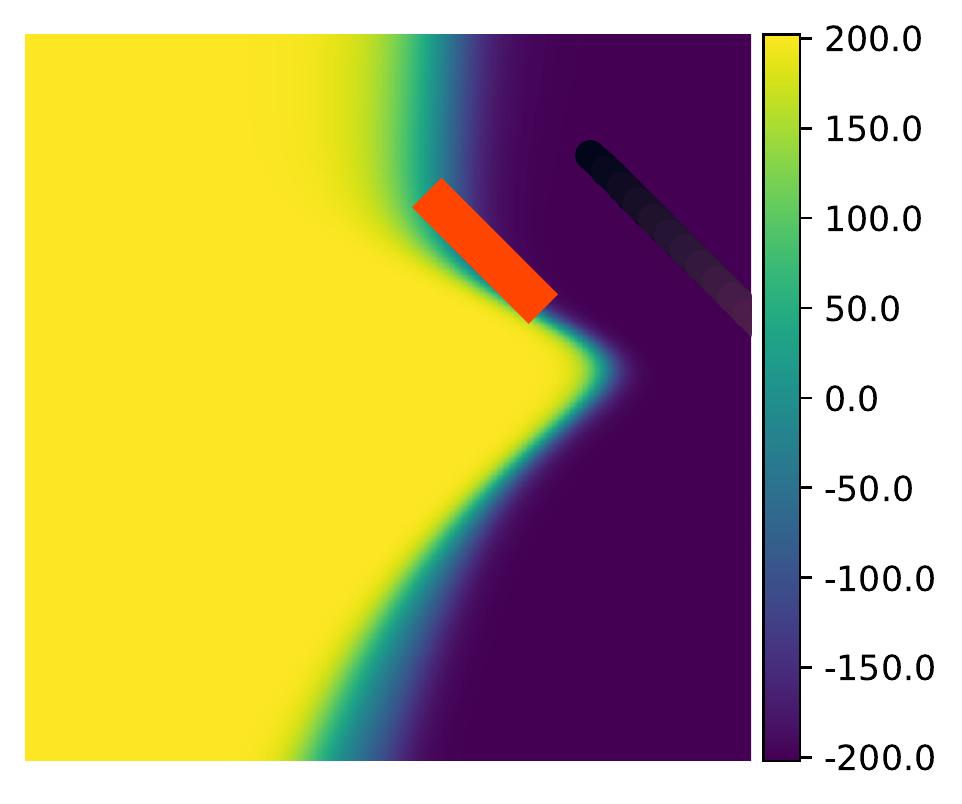}
    \caption{GCL Test}
  \end{subfigure}
  \caption{\small
    Qualitative results for all methods on the point mass navigation task with the obstacle.
  }
  \label{fig:all_qual_pmo} 
\end{figure*}

The results in \Cref{table:pmo_quant_results} are consistent with the non-obstacle point mass task where \genmethod generalizes better than a variety of MaxEnt IRL baselines.
In the train setting, \genmethod learns rewards that match the expert behavior with avoiding the obstacle and even achieves better performance than baselines in this task with 0.08 distance to the goal versus 0.41 to the goal for the best performing baseline in the train setting, f-IRL.
\genmethod generalizes better than baselines achieving 0.79 distance to goal compared to the best performing baseline MaxEnt, which also has access to oracle information. 
The reward learned by \genmethod visualized in \Cref{fig:qual_results:pmo_mirl} shows \genmethod learns a complex reward to account for the obstacle.
\Cref{fig:all_qual_pmo} visualizes the rewards for all methods.

\subsection{Comparison to Manually Defined Rewards}
\label{sec:manual_rewards} 

We compare the rewards learned by \genmethod to two hand-coded rewards. We visualize how well the learned rewards can train policies from scratch in the evaluation distribution in the point navigation with obstacle task. The reward learned by \genmethod therefore must generalize. On the other hand, the hand-coded rewards do not require any learning. We include a sparse reward for achieving the goal, which does not require domain knowledge when implementing the reward. We also implement a dense reward, defined as the change in Euclidean distance to the goal where $r_t = d_{t-1} - d_{t}$ where $d_{t}$ is the distance of the agent to the goal at time $t$.

\Cref{fig:reward_cmp} shows policy training curves for the learned and hand-defined rewards. The sparse reward performs poorly and the policy fails to get closer to the goal. On the other hand, the rewards learned by \genmethod guide the policy closer to the goal. The dense reward, which incoporates more domain knowledge about the task, performs better than the learned reward.

\subsection{Analyzing Number of Inner Loop Updates}
\label{sec:mirl_n_inner_iters}

\begin{figure}
  \centering
  \begin{subfigure}{0.38\textwidth}
    \includegraphics[width=\textwidth]{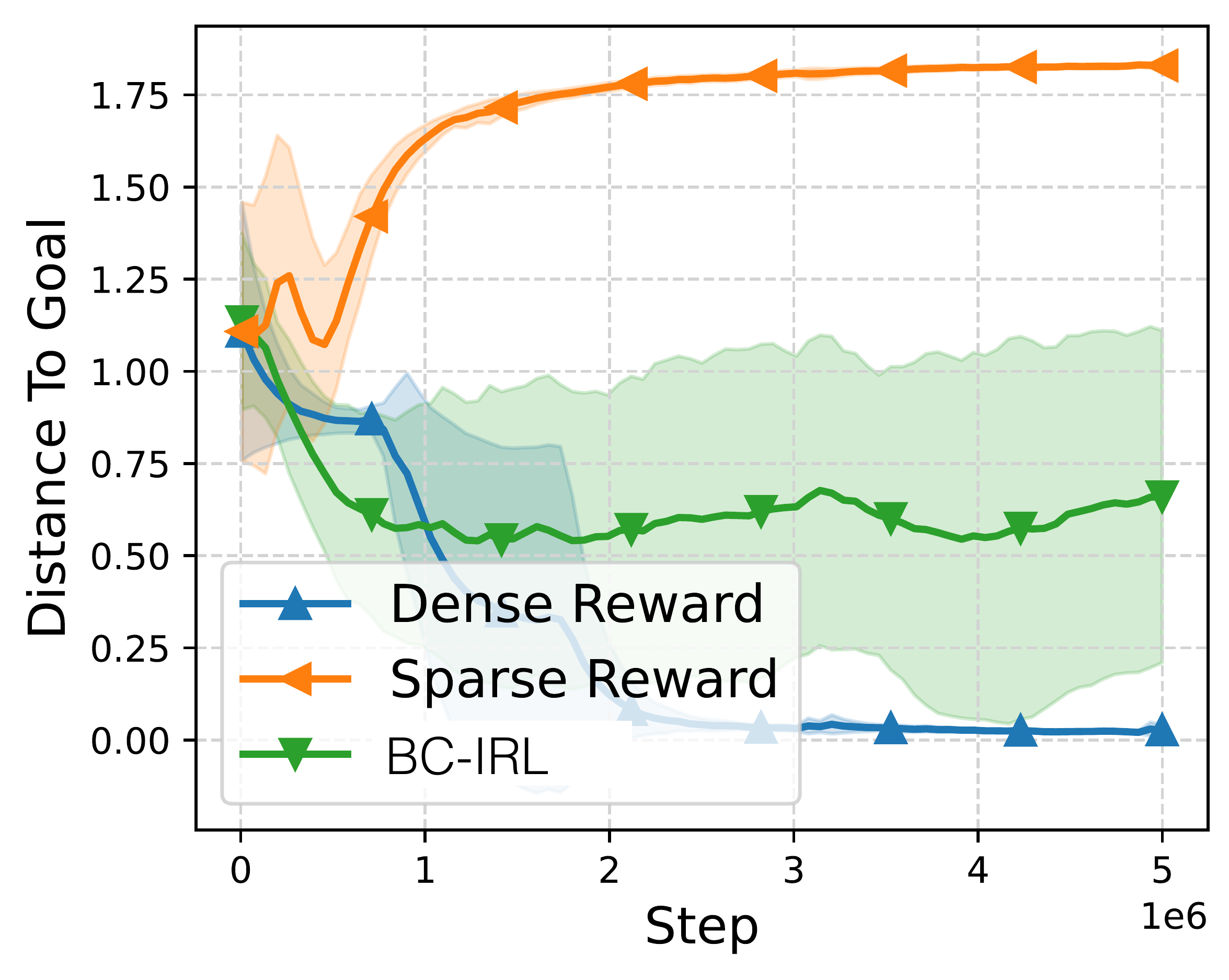}
    \caption{Hand-coded rewards vs. \genmethod.}
    \label{fig:reward_cmp}
  \end{subfigure}
  \begin{subfigure}{0.38\textwidth}
    \includegraphics[width=\textwidth]{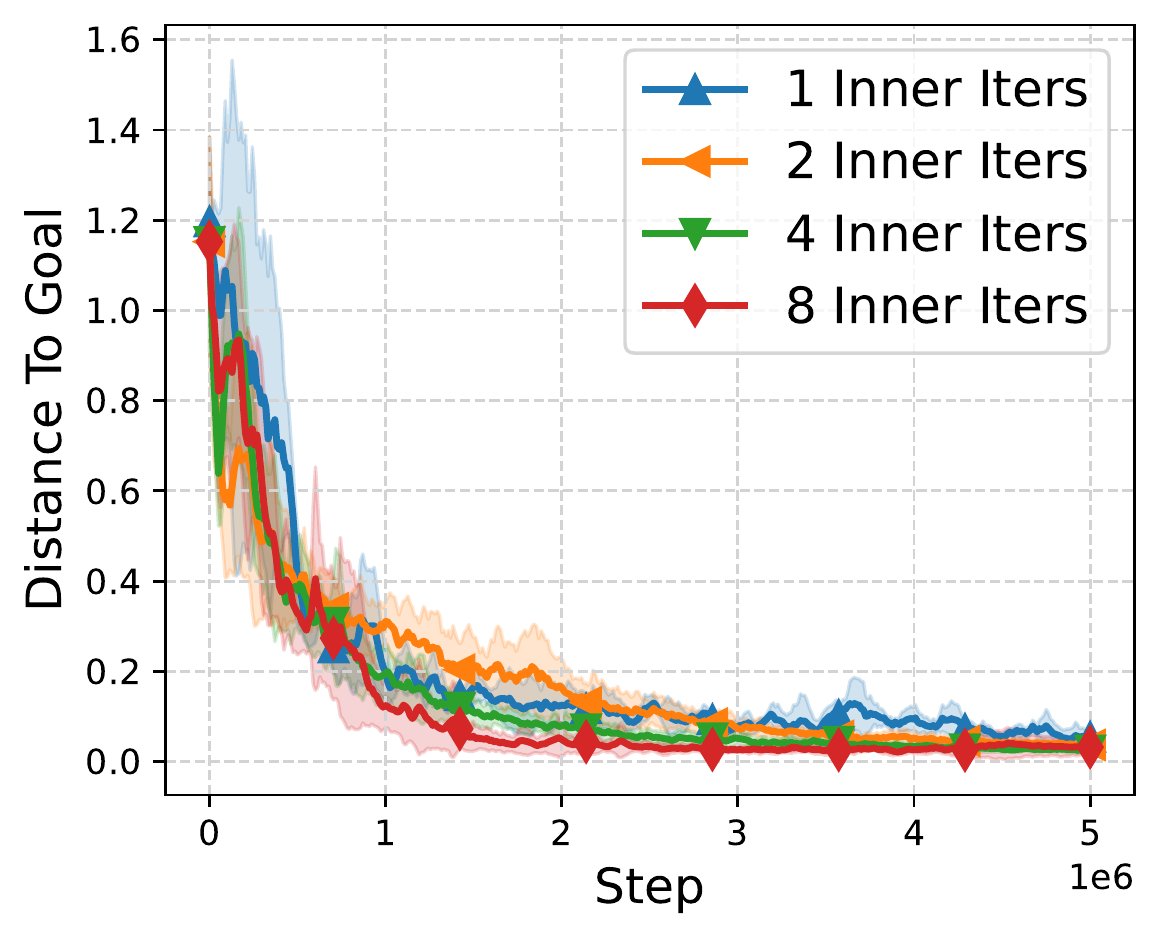}
    \caption{\# inner loop updates in \genmethod.}
    \label{fig:n_inner}
  \end{subfigure}
  \caption{
    Left: Comparing the reward learned from \genmethod to two manually hand-coded rewards. Right: Comparing different number of inner loop steps in \genmethod.
  }
\end{figure}

As described in \Cref{sec:method:ppo_mirl}, a hyperparameter in \ppomethod is the number of inner loop policy optimization steps $K$, for each reward function update. In our experiments, we selected $K=1$. In \Cref{fig:n_inner} we examine the training performance of \ppomethod in the point navigation task with no obstacle for various choices of $K$. We find that a wide variety of $K$ values perform similarly. We, therefore, selected $K=1$ since it runs the fastest, with no need to track multiple policy updates in the meta optimization.

\subsection{\genmethod with Model-Based Policy Optimization}
\label{14355}
We compare \ppomethod to a version of \genmethod that uses model-based RL in the inner loop inspired by \cite{das2020model}.
A direct comparison to \cite{das2020model} is not possible because their method assumes access to a pre-trained dynamics model, while in our work, we do not assume access to a ground truth or pre-trained dynamics model. 
However, we compare to a version of \cite{das2020model} in the point mass navigation task with a ground truth dynamics model. 
Specifically, we use gradient-based MPC in the inner loop optimization as in \cite{das2020model}, but the BC IRL outer loop objective. 
With the BC outer loop objective, it also learns generalizable rewards in the point mass navigation task achieving $0.06 \pm 0.03$ distance to goal in ``Eval (Train)" and $0.07 \pm 0.03$ in ``Eval (Test)". 
However, in the point mass navigation task with the obstacle, this method fails to learn a reward and struggles to minimize the outer loop objective. 
We hypothesize that in longer horizon tasks, the MPC inner loop optimization in [9] easily gets stuck in local minimas and struggles to differentiate through the entire MPC optimization.

%% file: sections/figures/pmo.tex
\begin{figure*}[t]
  \centering
   \begin{subfigure}[t]{0.19\textwidth}
   \centering
    \includegraphics[width=0.84\textwidth]{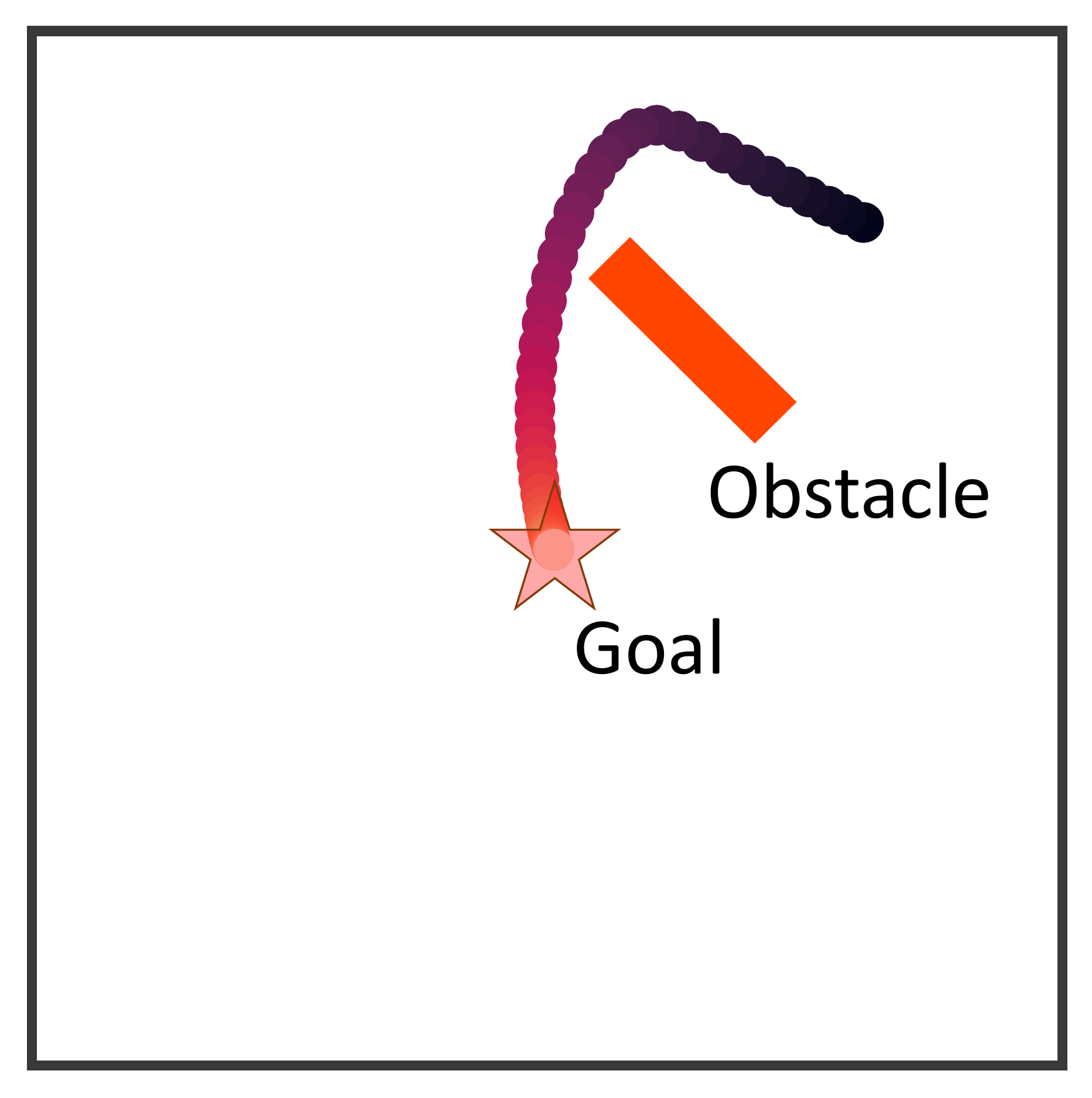}
    \caption{
      \small Task +Demos
    }
    \label{fig:qual_results:pmo_demo} 
  \end{subfigure}
  \begin{subfigure}[t]{0.19\textwidth}
    \includegraphics[width=\textwidth]{figures/exp/pmo/mirl_reward.pdf}
    \caption{
      \small
      Ours Reward
    }
    \label{fig:qual_results:pmo_mirl} 
  \end{subfigure}
  \begin{subfigure}[t]{0.19\textwidth}
    \includegraphics[width=\textwidth]{figures/exp/pmo/airl_reward.pdf}
    \caption{
      \small
      AIRL Reward 
    }
    \label{fig:qual_results:pmo_airl} 
  \end{subfigure}
  \begin{subfigure}[t]{0.19\textwidth}
    \includegraphics[width=\textwidth]{figures/exp/pmo/mirl_eval_rollouts.pdf}
    \caption{
      \small
      Ours Test
    }
    \label{fig:qual_results:eval_pmo_mirl}
  \end{subfigure}
  \begin{subfigure}[t]{0.19\textwidth}
    \includegraphics[width=\textwidth]{figures/exp/pmo/airl_eval_rollouts.pdf}
    \caption{
      \small
      AIRL Test
    }
    \label{fig:qual_results:eval_pmo_airl}
  \end{subfigure}
  \caption{\small
    Results on the point mass navigation tasks (top row: no-obstacle, bottom row: obstacle task). We show the learned reward functions of our method (b, g) vs. AIRL (c, h) and the policies learned from scratch using those reward functions (d, e, i, j). 
  }
  \vspace{-10pt}
  \label{fig:pmo_qual_results} 
\end{figure*}

%% file: sections/tables/pmo.tex
\begin{tabular}{cccccc}
\toprule
\footnotesize
 & \textbf{\genmethod} & \textbf{AIRL} & \textbf{GCL} & \textbf{MaxEnt} & \textbf{f-IRL} \\
\midrule
  \textbf{Train} &  0.08 {\scriptsize $\pm$ 0.00 }  &  0.62 {\scriptsize $\pm$ 0.50 }  &  1.30 {\scriptsize $\pm$ 0.71 }  & NA & 0.41 {\scriptsize $ \pm$ 0.35}\\
  \textbf{Eval (Train)} &  0.62 {\scriptsize $\pm$ 0.60 }  &  1.42 {\scriptsize $\pm$ 0.17 }  &  2.02 {\scriptsize $\pm$ 0.18 }  &  1.07 {\scriptsize $\pm$ 0.39 } & 0.61 {\scriptsize $ \pm $ 0.33} \\
  \textbf{Eval (Test)} & \textbf{ 0.79 {\scriptsize $\pm$ 0.65 } } &  1.42 {\scriptsize $\pm$ 0.18 }  &  2.01 {\scriptsize $\pm$ 0.18 }  &  0.83 {\scriptsize $\pm$ 0.01 } & 1.53 {\scriptsize $ \pm$ 0.24} \\
\bottomrule
\end{tabular}

%% file: supp/experiments.tex
\section{Reach Task: Further Experiment Results}
\label{sec:supp:reach} 

\subsection{RL-Training Curves}
\label{sec:rl_training} 
\begin{figure*}[h]
  \begin{subfigure}[t]{0.24\textwidth}
    \includegraphics[width=\textwidth]{figures/curves/reach_train.pdf}
    \caption{IRL Training}
    \label{fig:supp_rl_curves:train} 
  \end{subfigure}
  \begin{subfigure}[t]{0.24\textwidth}
    \includegraphics[width=\textwidth]{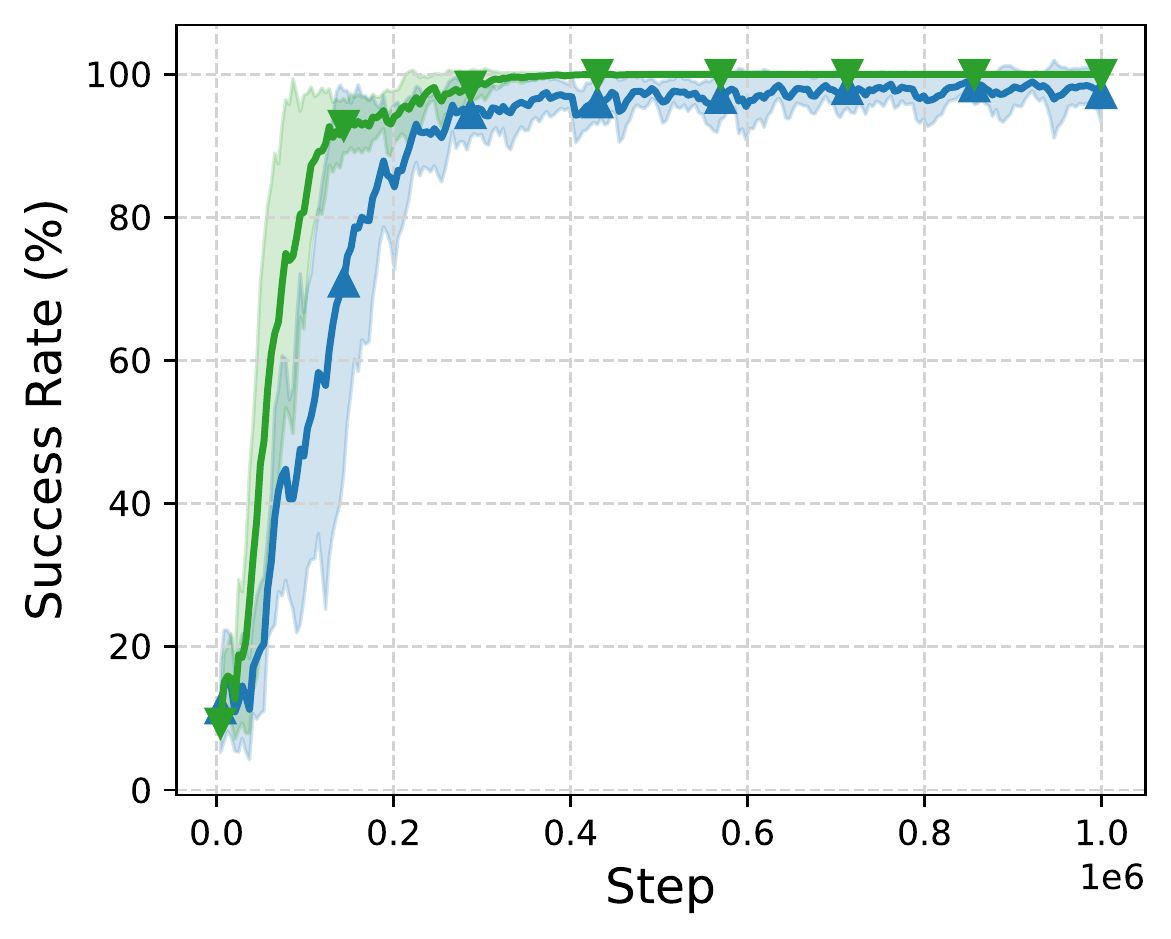}
    \caption{Start Distrib: \seasy}
    \label{fig:rl_curves:x25} 
  \end{subfigure}
  \begin{subfigure}[t]{0.24\textwidth}
    \includegraphics[width=\textwidth]{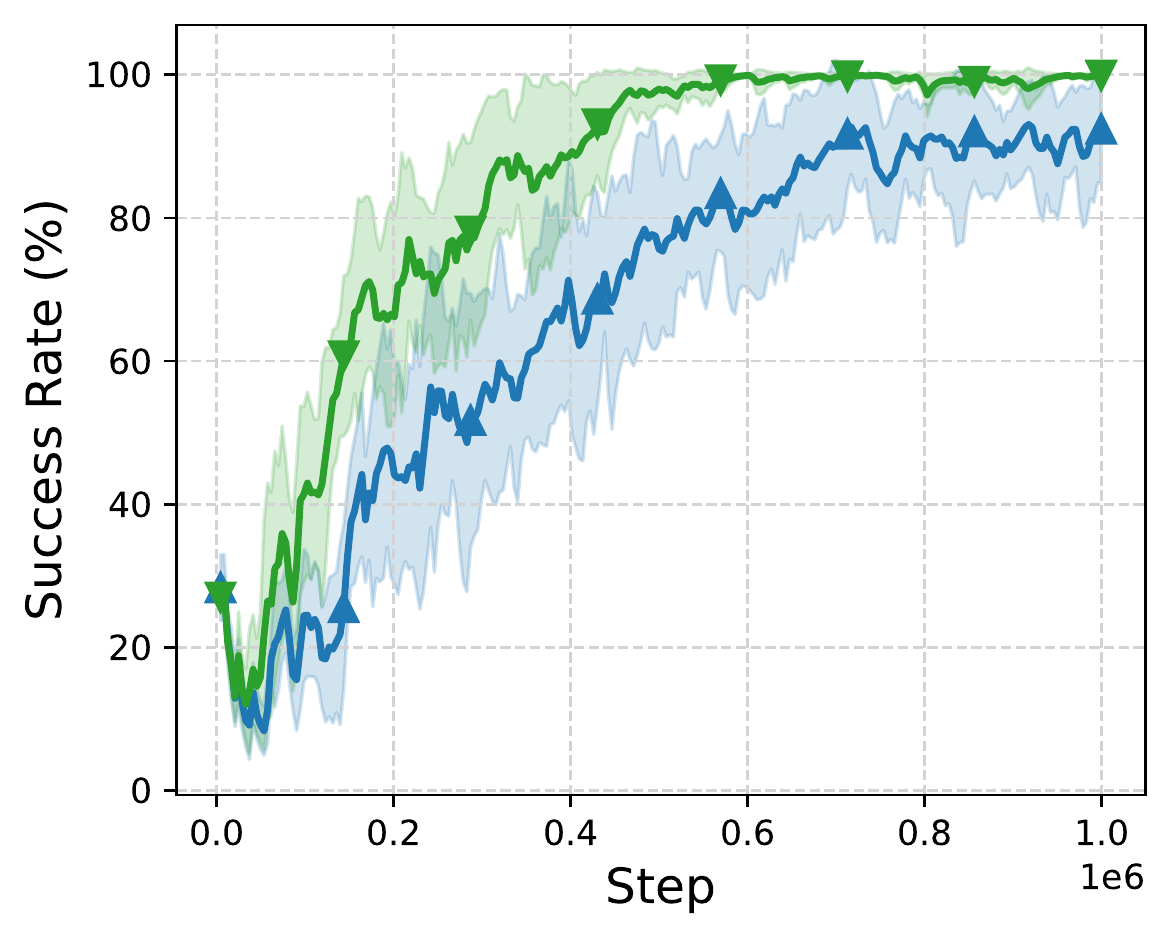}
    \caption{Start Distrib: \smed}
    \label{fig:rl_curves:x75} 
  \end{subfigure}
  \begin{subfigure}[t]{0.24\textwidth}
    \includegraphics[width=\textwidth]{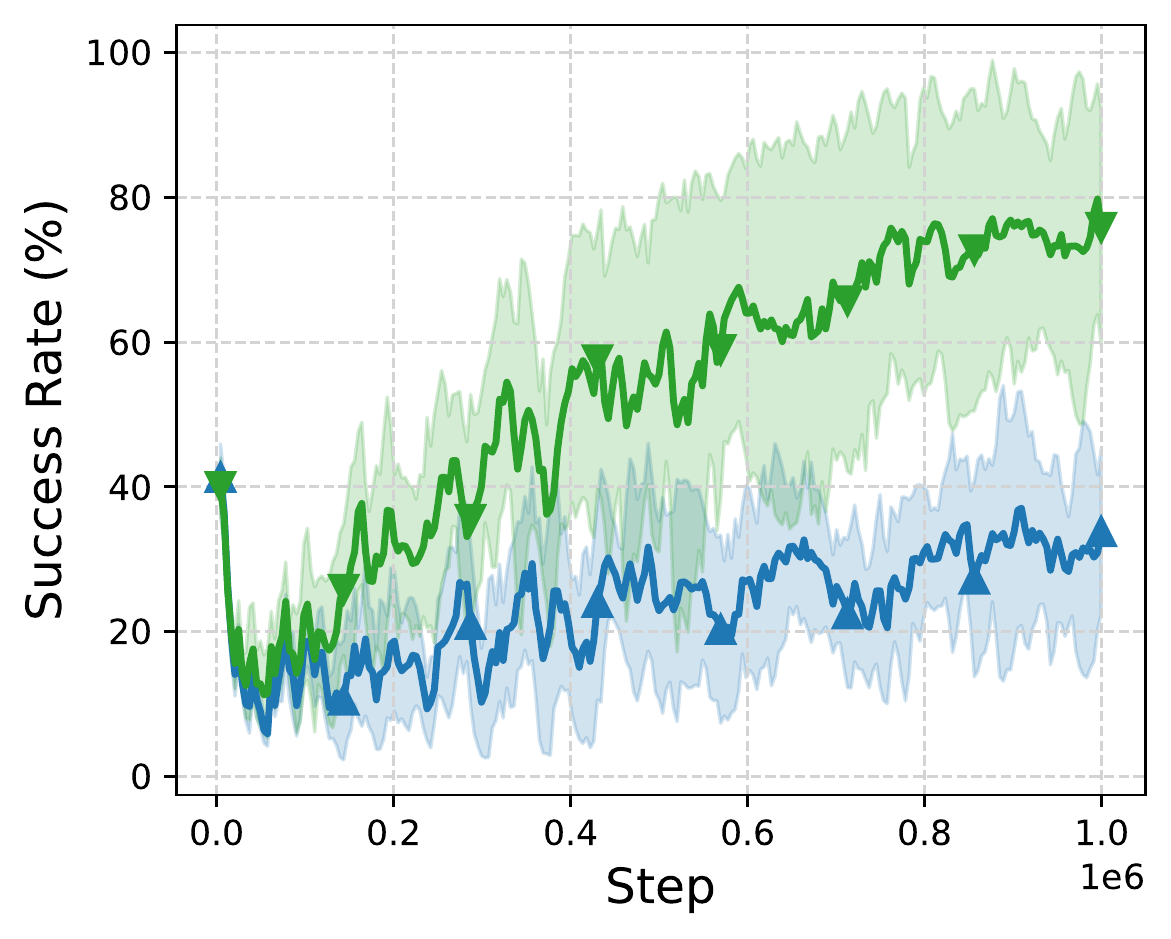}
    \caption{Start Distrib: \shard}
    \label{fig:rl_curves:x100} 
  \end{subfigure}
  \caption{
    Learning curves for the training setting and \scratch transfer strategies from \Cref{table:reach_eval}.
    All results are for 3 seeds and error regions show the standard deviation in success rate between seeds.
  }
  \label{fig:rl_curves} 
\end{figure*}

In \Cref{fig:rl_curves} we visualize the training curves for the RL training used in \Cref{table:reach_eval}.
\Cref{fig:supp_rl_curves:train} shows policy learning progress during the IRL training phase. 
In each setting, the performance is measured by using the current reward to train a policy and computing the success rate of the policy.
\Cref{fig:rl_curves:x25} to \Cref{fig:rl_curves:x100} show the policy learning curves at test time, in the generalization settings, where the reward is frozen and must generalize to learn new policies on new goals (``\Scratch" transfer strategy).
These plots show that all methods learn similarly during IRL training (\Cref{fig:supp_rl_curves:train}). 
When transferring the learned rewards to test settings we see that \ppomethod performs better in training successful policies as the generalization difficulty increases with the most difficult generalization in \Cref{fig:rl_curves:x100}.

\subsection{Transfer Reward+Policy Setting}
\label{sec:adapt} 

\begin{table*}[h]
  \centering
  \resizebox{0.95\textwidth}{!}{
    \input{sections/tables/reach_table_adapt.tex}
  }
  \caption{
    Results for the \adapt transfer strategy where the trained policy and reward are transferred to the test setting and the policy is fine-tuned.
  }
  \label{table:reach_adapt} 
\end{table*}

Here, we evaluate the \textbf{\adapt} transfer strategy to new environment settings where both the reward and policy are transferred. 
In the new setting, \adapt uses the transferred reward to fine-tune the pre-trained transferred policy with RL.
We show results in \Cref{table:reach_adapt} for the \adapt transfer strategy alongside the \scratch transfer strategy from \Cref{table:reach_eval}.
We find that \adapt performs slightly better than \scratch in the Hard setting of generalization to new starting state distributions but otherwise performs similarly. 
Even in the \adapt setting, AIRL struggles to learn a good policy in the Medium and Hard settings, achieving $38\%$ and $81\%$ success rate respectively.

\subsection{Analyzing the Number Demonstrations}\label{sec:num_demos} 

\begin{table*}[h!]
  \centering
  \resizebox{0.95\textwidth}{!}{
    \input{sections/tables/demo_ablate.tex}
  }
  \caption{
    Comparing the number of demonstrations for \ppomethod and AIRL across the train, medium, and hard settings. 
  }
  \label{table:ablate_num_demos}
\end{table*}

We analyze the effect of the number of demonstrations used for reward learning in \Cref{table:ablate_num_demos}.
We find that using fewer demonstrations does not affect the training performance of \ppomethod and AIRL.
We also find our method does just as well with 5 demos as 100 in the +75\% noise setting, with any number of demonstrations achieving near-perfect success rates. 
On the other hand, the performance of AIRL degrades from 93\% success rate with 100 demonstrations to 84\% in the +75\% noise setting.
In the +100\% noise setting, fewer demonstrations hurt performance for both methods, with our method dropping from 76\% success to 69\% success and AIRL from 38\% success to 42\% success.

\subsection{\genmethod Hyperpararameter Analysis}
\label{sec:mirl_ablate}
\ppomethod requires a learning rate for the policy optimization and a learning rate for the reward optimization. 
We compare the performance of our algorithm for various choices of policy and reward learning rates in \Cref{table:lr_ablate}.
We find that across many different learning rate settings our method achieves high rates of success, but high policy learning rates have a detrimental effect.
High reward learning rates have a slight negative impact but are not as severe.

\begin{table}
  \centering
  \input{sections/tables/lr_ablate.tex}
  \caption{
    Comparing choice of learning rate for the inner and outer loops for \ppomethod on the train setting.
    Numbers display success rate.
  }
  \label{table:lr_ablate}
\end{table}

%% file: sections/tables/reach_table_adapt.tex
\begin{tabular}{ccccc}
 & \textbf{\ppomethod (\Adapt)} & \textbf{\ppomethod (\Scratch)} & \textbf{AIRL (\Adapt)} & \textbf{AIRL (\Scratch)} \\
\midrule
\textbf{\usetrain} &  1.00 {\scriptsize $\pm$ 0.00 }  &  1.00 {\scriptsize $\pm$ 0.00 }  &  0.96 {\scriptsize $\pm$ 0.00 }  &  0.96 {\scriptsize $\pm$ 0.00 }  \\
\hline
\textbf{Start Distrib: \useeasy} & \textbf{ 1.00 {\scriptsize $\pm$ 0.00 } } &  \textbf{1.00 {\scriptsize $\pm$ 0.00 }}  &  0.92 {\scriptsize $\pm$ 0.00 }  &  0.96 {\scriptsize $\pm$ 0.04 }  \\
\textbf{Start Distrib: \usemedium} & \textbf{ 1.00 {\scriptsize $\pm$ 0.00 } } &  \textbf{1.00 {\scriptsize $\pm$ 0.00 }}  &  0.81 {\scriptsize $\pm$ 0.07 }  &  0.93 {\scriptsize $\pm$ 0.05 }  \\
\textbf{Start Distrib: \usehard} & \textbf{ 0.80 {\scriptsize $\pm$ 0.14 } } &  0.76 {\scriptsize $\pm$ 0.16 }  &  0.38 {\scriptsize $\pm$ 0.03 }  &  0.38 {\scriptsize $\pm$ 0.06 }  \\
\hline
\textbf{State Region: \useeasy} & \textbf{ 1.00 {\scriptsize $\pm$ 0.00 } } &  \textbf{1.00 {\scriptsize $\pm$ 0.00 }}  &  0.05 {\scriptsize $\pm$ 0.02 }  &  \textbf{1.00 {\scriptsize $\pm$ 0.01 }}  \\
\textbf{State Region: \usemedium} & \textbf{ 1.00 {\scriptsize $\pm$ 0.00 } } &  \textbf{1.00 {\scriptsize $\pm$ 0.00 }}  &  0.17 {\scriptsize $\pm$ 0.09 }  &  0.91 {\scriptsize $\pm$ 0.02 }  \\
\textbf{State Region: \usehard} &  0.76 {\scriptsize $\pm$ 0.15 }  & \textbf{ 0.78 {\scriptsize $\pm$ 0.13 } } &  0.21 {\scriptsize $\pm$ 0.11 }  &  0.32 {\scriptsize $\pm$ 0.08 }  \\
\bottomrule
\end{tabular}

%% file: sections/tables/demo_ablate.tex
\begin{tabular}{ccccccc}
\toprule
 & \textbf{\ppomethod (Train)} & \textbf{\ppomethod \smed} & \textbf{\ppomethod \shard} & \textbf{AIRL (Train)} & \textbf{AIRL \smed} & \textbf{AIRL \shard} \\
\midrule
\textbf{100 Demos} &  1.00 {\scriptsize $\pm$ 0.0 }  &  1.00 {\scriptsize $\pm$ 0.00 }  &  0.76 {\scriptsize $\pm$ 0.16 }  &  0.96 {\scriptsize $\pm$ 0.00 }  &  0.93 {\scriptsize $\pm$ 0.05 }  &  0.38 {\scriptsize $\pm$ 0.06 }  \\
\textbf{50 Demos} &  0.96 {\scriptsize $\pm$ 0.05 }  &  0.91 {\scriptsize $\pm$ 0.16 }  &  0.70 {\scriptsize $\pm$ 0.17 }  &  0.96 {\scriptsize $\pm$ 0.01 }  &  0.88 {\scriptsize $\pm$ 0.09 }  &  0.40 {\scriptsize $\pm$ 0.05 }  \\
\textbf{25 Demos} &  0.99 {\scriptsize $\pm$ 0.01 }  &  0.99 {\scriptsize $\pm$ 0.01 }  &  0.56 {\scriptsize $\pm$ 0.12 }  &  0.95 {\scriptsize $\pm$ 0.01 }  &  0.84 {\scriptsize $\pm$ 0.03 }  &  0.48 {\scriptsize $\pm$ 0.09 }  \\
\textbf{5 Demos} &  0.99 {\scriptsize $\pm$ 0.01 }  &  1.00 {\scriptsize $\pm$ 0.01 }  &  0.69 {\scriptsize $\pm$ 0.19 }  &  0.97 {\scriptsize $\pm$ 0.01 }  &  0.84 {\scriptsize $\pm$ 0.03 }  &  0.42 {\scriptsize $\pm$ 0.02 }  \\
\bottomrule
\end{tabular}

%% file: sections/tables/lr_ablate.tex
\begin{tabular}{cc|cccc}
& \multicolumn{5}{c}{Reward LR} \\
 &  & \textbf{1e-4} & \textbf{1e-3} & \textbf{1e-2} & \textbf{1e-1} \\
\cmidrule{2-6}
\multirow{4}{1em}{\rotatebox{90}{Policy LR}} & \textbf{1e-4} &  0.96 {\scriptsize $\pm$ 0.01 }  &  0.96 {\scriptsize $\pm$ 0.01 }  &  0.89 {\scriptsize $\pm$ 0.11 }  &  0.71 {\scriptsize $\pm$ 0.45 }  \\
 & \textbf{1e-3} &  0.97 {\scriptsize $\pm$ 0.03 }  &  1.00 {\scriptsize $\pm$ 0.00 }  &  1.00 {\scriptsize $\pm$ 0.00 }  &  0.66 {\scriptsize $\pm$ 0.56 }  \\
 & \textbf{1e-2} &  0.14 {\scriptsize $\pm$ 0.23 }  &  0.28 {\scriptsize $\pm$ 0.48 }  &  0.11 {\scriptsize $\pm$ 0.17 }  &  0.59 {\scriptsize $\pm$ 0.53 }  \\
 & \textbf{1e-1} &  0.64 {\scriptsize $\pm$ 0.55 }  &  0.33 {\scriptsize $\pm$ 0.55 }  &  0.02 {\scriptsize $\pm$ 0.04 }  &  0.32 {\scriptsize $\pm$ 0.54 }  \\
\cmidrule{2-6}

\end{tabular}

%% file: supp/pm_details.tex
\section{Further 2D Point Navigation Details}
\label{sec:pm_hyperparams} 

\begin{figure}[!h]
  \centering
  \includegraphics[width=0.40\textwidth]{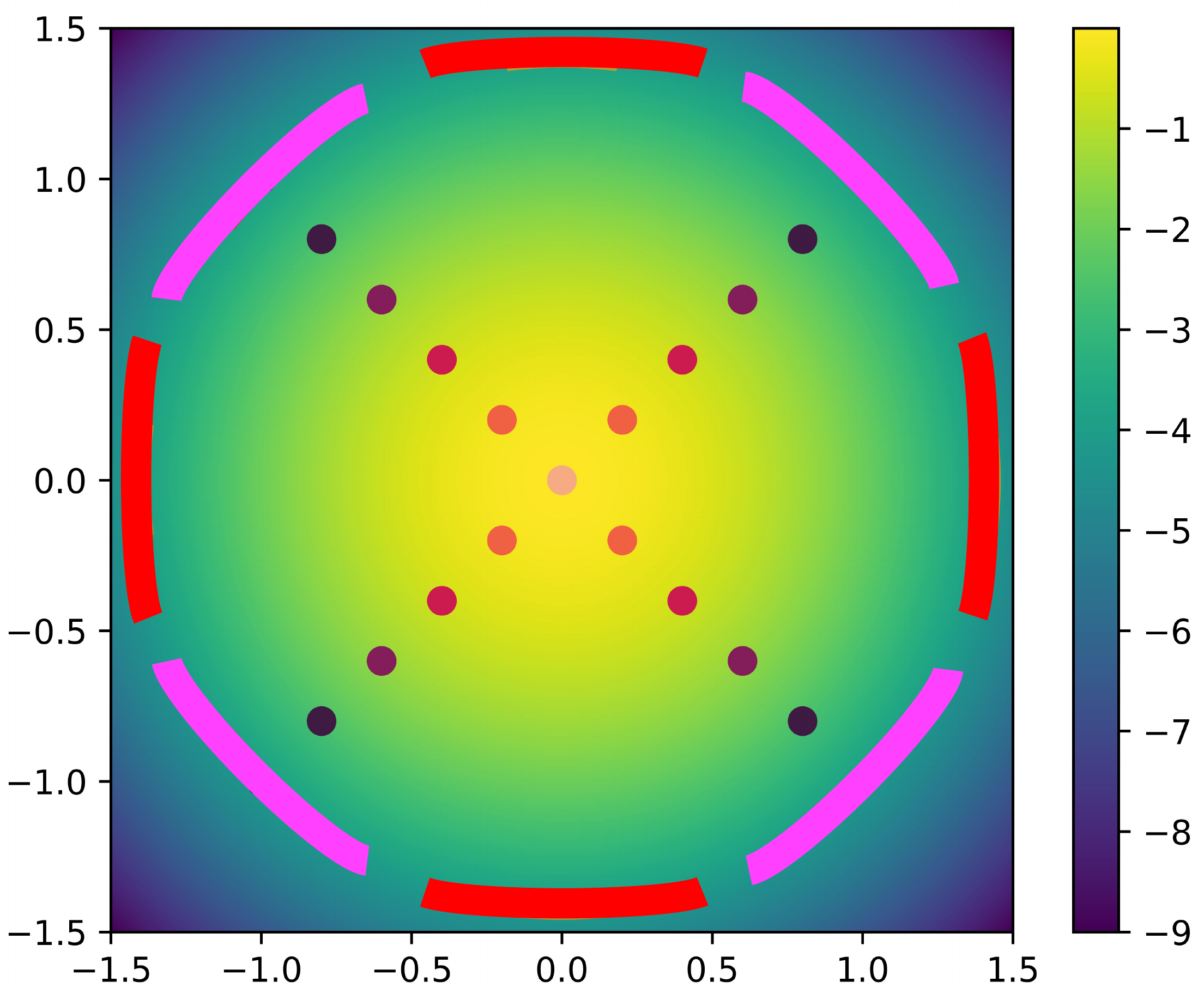}
  \caption{
    The starting state distribution for the 2D point navigation task with the demonstrations and negative distance to the goal overlaid.
    The training start state distribution where the reward is learned is in \textcolor[rgb]{1.0, 0.03, 0.11}{red}.
    The test start state distribution where the reward is transferred is in \textcolor[rgb]{1.0,0.23,0.98}{magenta}.
  }
  \label{fig:pm_nav_start_state}
\end{figure}

The start state distributions for the 2D point navigation task are illustrated in \Cref{fig:pm_nav_start_state}.
The reward is learned using the start distribution in red on 4 equally spaced points from the center.
Four demonstrations are also provided in this train start state distribution from each of the four corners.
The reward is then transferred and a new policy is trained with the start state distribution in the magenta color. 
This start state distribution has no overlap with the train distribution and is also equally spaced.
The reward must therefore generalize to providing rewards in this new state distribution.

The hyperparameters for the methods from the 2D point navigation task in \Cref{sec:reward_analysis} are detailed in \Cref{tab:pm_hyperparam} for the no obstacle version and \Cref{tab:pmo_hyperparam} for the obstacle version of the task.
The reward function / discriminator for all methods was a neural network with 1 hidden layer and 128 hidden dimension size with $ tanh$-activations between the layers.
Adam \cite{kingma2014adam} was used for policy and reward optimization.
All RL training used 5M steps of experience for the training and testing setting for the navigation no obstacle task.
f-IRL uses the same optimization and neural network hyperparameters for the discriminator and reward function.
Like in \cite{ni2020f}, we clamp the output of the reward function within the range $ [-10, 10]$ and found this was beneficial for learning.
In the navigation with obstacle task, training used 15M steps of experience and testing used 5M steps of experience. 
All experiments were run on a Intel(R) Core(TM) i9-9900X CPU @ 3.50GHz.

\begin{table}
  \centering
  \begin{tabular}{cccccc}
    \toprule
    Hyperparameter             & \ppomethod & AIRL & GCL & MaxEnt & f-IRL\\ 
    \midrule
    Reward Learning Rate        & 1e-4 & 1e-3 & 3e-4 & 1e-3 & 3e-4\\
    Reward Batch Size    & 20   & 20   & 20   & 20 & 20\\
    Policy Learning Rate        & 1e-4 & 1e-4 & 3e-4 & 3e-4 & 3e-4 \\
    Policy Learning Rate Decay  & True & True & True & False & False\\
    Policy \# Mini-batches      & 4    & 4    & 4    & 4 & 4 \\
    Policy \# Epochs per Update & 4    & 4    & 4    & 4 & 4 \\
    Policy Entropy Coefficient  & 1e-4 & 1e-4 & 1e-4 & 1e-4 & 1e-4 \\
    Discount Factor $\gamma$    & 0.99 & 0.99 & 0.99 & 0.99 & 0.99 \\
    Policy Batch Size                & 1280  & 1280  & 1280  & 1280 & 1280 \\
    \bottomrule
  \end{tabular}
  \caption{
    2D navigation without obstacle method hyperparameters.
    These hyperparameters were used both in the training and reward transfer settings.
  }
  \label{tab:pm_hyperparam}
\end{table}

\begin{table}
  \centering
  \begin{tabular}{ccccc}
    \toprule
    Hyperparameter             & \ppomethod & AIRL & GCL & MaxEnt \\
    \midrule
    Reward Learning Rate        & 1e-4 & 1e-3 & 3e-4 & 1e-3 \\
    Reward Batch Size    & 256   & 256   & 256   & 256 \\
    Policy Learning Rate        & 3e-4 & 3e-4 & 3e-4 & 3e-4 \\
    Policy Learning Rate Decay  & True & True & True & False \\
    Policy \# Mini-batches      & 4    & 4    & 4    & 4 \\
    Policy \# Epochs per Update & 4    & 4    & 4    & 4 \\
    Policy Entropy Coefficient  & 1e-4 & 1e-4 & 1e-4 & 1e-4 \\
    Discount Factor $\gamma$    & 0.99 & 0.99 & 0.99 & 0.99 \\
    Policy Batch Size                & 6400  & 6400  & 6400  & 6400 \\
    \bottomrule
  \end{tabular}
  \caption{
    2D navigation with obstacle method hyperparameters.
    These hyperparameters were used both in the training and reward transfer settings.
  }
  \label{tab:pmo_hyperparam}
\end{table}

%% file: supp/reach_details.tex
\section{Further Reach Task Details}
\label{app:reach-details}

\subsection{Choice of Baselines}
The ``Exact MaxEntIRL" approach is excluded because it cannot be computed exactly for high-dimensional state spaces.
GCL is excluded because of its poor performance on the toy task relative to other methods.
We also compare to the following imitation learning methods which learn only policies and no transferable reward:
\begin{itemize}
\item \textbf{Behavioral Cloning (BC)} \cite{bain1995framework}: Train a policy using supervised learning to match the actions in the expert dataset. 
\item \textbf{Generative Adversarial Imitation Learning (GAIL)} \cite{ho2016generative}: Trains a discriminator to distinguish expert from agent transitions and then use the discriminator confusion score as the reward. This reward is coupled with the current policy \cite{finn2016connection} (referred to as a ``pseudo-reward") and therefore cannot train policies from scratch.
\end{itemize}

\subsection{Policy+Network representation}
All methods use a neural network to represent the policy and reward with 1 hidden layer, 128 hidden units, and $tanh$-activation functions between the layers. 
We use PPO as the policy optimization method for all methods. 
All methods in all tasks use demonstrations obtained from a policy trained with PPO using a manually engineered reward. 

\subsection{Hyperparameters}
\label{sec:hyperparams} 
The hyperparameters for all methods from the Reaching task are described in \Cref{tab:hyperparam}.
The Adam optimizer \cite{kingma2014adam} was used for policy and reward optimization.
All RL training used 1M steps of experience for the training and testing settings.
The \scratch and \adapt transfer strategies trained policies with the same set of hyperparameters.

\begin{table}
  \centering
  \begin{tabular}{ccccc}
    \toprule
    Hyperparameter & \ppomethod & AIRL & GAIL & BC \\
    \midrule
    Reward Learning Rate        & 3e-4  & 1e-4 & 1e-4 & NA \\
    Reward Batch Size           & 128   & 128  & 128  & NA \\
    Policy Learning Rate        & 3e-4  & 1e-4 & 1e-4 & 1e-4 \\
    Policy Learning Rate Decay  & False & True & True & False \\
    Policy \# Mini-batches      & 4     & 4    & 4    & NA \\
    Policy \# Epochs per Update & 4     & 4    & 4    & NA \\
    Policy Entropy Coefficient  & 0.0   & 0.0  & 0.0  & NA \\
    Discount Factor $\gamma$    & 0.99  & 0.99 & 0.99 & 0.99 \\
    Policy Batch Size                & 4096  & 4096 & 4096 & NA \\
    \bottomrule
  \end{tabular}
  \caption{
    Method hyperparameters for the Fetch reaching task. 
    These hyperparameters were used both in the training and reward transfer settings.
  }
  \label{tab:hyperparam}
\end{table}

%% file: supp/trifinger-details.tex
\section{Trifinger experiment details}
\label{sec:trifinger_details}

\subsection{Policy+Network representation}
All methods use a neural network to represent the policy and reward with 1 hidden layer, 128 hidden units, and $tanh$-activation functions between the layers. 
We use PPO as the policy optimization method for all methods. 
All methods in all tasks use demonstrations obtained from a policy trained with PPO using a manually engineered reward. 

\subsection{Hyperparameters}

\begin{table}
  \centering
  \begin{tabular}{ccc}
    \toprule
    Hyperparameter & \ppomethod & AIRL \\
    \midrule
    Reward Learning Rate        & 1e-3  & 1e-3 \\
    Reward Batch Size           & 6   & 6  \\
    Policy Learning Rate        & 1e-4  & 1e-3  \\
    Policy Learning Rate Decay  & False & False \\
    Policy \# Mini-batches      & 4     & 4   \\
    Policy \# Epochs per Update & 2     & 2   \\
    Policy Entropy Coefficient  & 0.005   & 0.005 \\
    Discount Factor $\gamma$    & 0.99  & 0.99 \\
    Policy Batch Size           & 40 & 40 \\
    \bottomrule
  \end{tabular}
  \caption{
    Method hyperparameters for the Trifinger reaching task. 
    These hyperparameters were used both in the training and reward transfer settings.
  }
  \label{tab:hyperparam-trf}
\end{table}

The hyperparameters for all methods for the Trifinger reaching task are described in \Cref{tab:hyperparam-trf}.
The Adam optimizer \cite{kingma2014adam} was used for policy and reward optimization.
All RL training used 500k steps of experience for the reward training phase and 100k steps of experience for policy optimization in test settings.

%% file: main.bbl
\begin{thebibliography}{46}
\providecommand{\natexlab}[1]{#1}
\providecommand{\url}[1]{\texttt{#1}}
\expandafter\ifx\csname urlstyle\endcsname\relax
  \providecommand{\doi}[1]{doi: #1}\else
  \providecommand{\doi}{doi: \begingroup \urlstyle{rm}\Url}\fi

\bibitem[Abbeel \& Ng(2004)Abbeel and Ng]{abbeel2004apprenticeship}
Pieter Abbeel and Andrew~Y Ng.
\newblock Apprenticeship learning via inverse reinforcement learning.
\newblock In \emph{Proceedings of the twenty-first international conference on
  Machine learning}, pp.\ ~1, 2004.

\bibitem[Abbeel et~al.(2010)Abbeel, Coates, and Ng]{abbeel2010autonomous}
Pieter Abbeel, Adam Coates, and Andrew~Y Ng.
\newblock Autonomous helicopter aerobatics through apprenticeship learning.
\newblock \emph{The International Journal of Robotics Research}, 29\penalty0
  (13):\penalty0 1608--1639, 2010.

\bibitem[Ahmed et~al.(2021)Ahmed, Tr{\"a}uble, Goyal, Neitz, W{\"u}thrich,
  Bengio, Sch{\"o}lkopf, and Bauer]{ahmed2021causalworld}
Ossama Ahmed, Frederik Tr{\"a}uble, Anirudh Goyal, Alexander Neitz, Manuel
  W{\"u}thrich, Yoshua Bengio, Bernhard Sch{\"o}lkopf, and Stefan Bauer.
\newblock Causalworld: A robotic manipulation benchmark for causal structure
  and transfer learning.
\newblock In \emph{International Conference on Learning Representations}, 2021.

\bibitem[Bain \& Sammut(1995)Bain and Sammut]{bain1995framework}
Michael Bain and Claude Sammut.
\newblock A framework for behavioural cloning.
\newblock In \emph{Machine Intelligence 15}, pp.\  103--129, 1995.

\bibitem[Barde et~al.(2020)Barde, Roy, Jeon, Pineau, Pal, and
  Nowrouzezahrai]{barde2020adversarial}
Paul Barde, Julien Roy, Wonseok Jeon, Joelle Pineau, Christopher Pal, and Derek
  Nowrouzezahrai.
\newblock Adversarial soft advantage fitting: Imitation learning without policy
  optimization.
\newblock \emph{arXiv preprint arXiv:2006.13258}, 2020.

\bibitem[Bechtle et~al.(2021)Bechtle, Molchanov, Chebotar, Grefenstette,
  Righetti, Sukhatme, and Meier]{bechtle2019meta}
Sarah Bechtle, Artem Molchanov, Yevgen Chebotar, Edward Grefenstette, Ludovic
  Righetti, Gaurav Sukhatme, and Franziska Meier.
\newblock Meta-learning via learned loss.
\newblock \emph{ICPR}, 2021.

\bibitem[Boularias et~al.(2011)Boularias, Kober, and
  Peters]{boularias2011relativeirl}
Abdeslam Boularias, Jens Kober, and Jan Peters.
\newblock Relative entropy inverse reinforcement learning.
\newblock In \emph{Proceedings of the Fourteenth International Conference on
  Artificial Intelligence and Statistics}, pp.\  182--189, 2011.

\bibitem[Dadashi et~al.(2020)Dadashi, Hussenot, Geist, and
  Pietquin]{dadashi2020primal}
Robert Dadashi, L{\'e}onard Hussenot, Matthieu Geist, and Olivier Pietquin.
\newblock Primal wasserstein imitation learning.
\newblock \emph{arXiv preprint arXiv:2006.04678}, 2020.

\bibitem[Das et~al.(2020)Das, Bechtle, Davchev, Jayaraman, Rai, and
  Meier]{das2020model}
Neha Das, Sarah Bechtle, Todor Davchev, Dinesh Jayaraman, Akshara Rai, and
  Franziska Meier.
\newblock Model-based inverse reinforcement learning from visual
  demonstrations.
\newblock \emph{arXiv preprint arXiv:2010.09034}, 2020.

\bibitem[Englert et~al.(2017)Englert, Vien, and
  Toussaint]{englert2017inverseRL}
Peter Englert, Ngo~Anh Vien, and Marc Toussaint.
\newblock Inverse kkt: Learning cost functions of manipulation tasks from
  demonstrations.
\newblock \emph{The International Journal of Robotics Research}, 36\penalty0
  (13-14):\penalty0 1474--1488, 2017.

\bibitem[Finn et~al.(2016{\natexlab{a}})Finn, Christiano, Abbeel, and
  Levine]{finn2016connection}
Chelsea Finn, Paul Christiano, Pieter Abbeel, and Sergey Levine.
\newblock A connection between generative adversarial networks, inverse
  reinforcement learning, and energy-based models.
\newblock \emph{arXiv preprint arXiv:1611.03852}, 2016{\natexlab{a}}.

\bibitem[Finn et~al.(2016{\natexlab{b}})Finn, Levine, and
  Abbeel]{finn2016guided}
Chelsea Finn, Sergey Levine, and Pieter Abbeel.
\newblock Guided cost learning: Deep inverse optimal control via policy
  optimization.
\newblock In \emph{International conference on machine learning}, pp.\  49--58.
  PMLR, 2016{\natexlab{b}}.

\bibitem[Finn et~al.(2016{\natexlab{c}})Finn, Levine, and
  Abbeel]{finn2016guidedirl}
Chelsea Finn, Sergey Levine, and Pieter Abbeel.
\newblock Guided cost learning: Deep inverse optimal control via policy
  optimization.
\newblock In \emph{International conference on machine learning}, pp.\  49--58,
  2016{\natexlab{c}}.

\bibitem[Finn et~al.(2017)Finn, Abbeel, and Levine]{finn2017model}
Chelsea Finn, Pieter Abbeel, and Sergey Levine.
\newblock Model-agnostic meta-learning for fast adaptation of deep networks.
\newblock In \emph{International Conference on Machine Learning}, pp.\
  1126--1135. PMLR, 2017.

\bibitem[Fu et~al.(2017)Fu, Luo, and Levine]{fu2017learning}
Justin Fu, Katie Luo, and Sergey Levine.
\newblock Learning robust rewards with adversarial inverse reinforcement
  learning.
\newblock \emph{arXiv preprint arXiv:1710.11248}, 2017.

\bibitem[Gleave \& Habryka(2018)Gleave and Habryka]{gleave2018multi}
Adam Gleave and Oliver Habryka.
\newblock Multi-task maximum entropy inverse reinforcement learning.
\newblock \emph{arXiv preprint arXiv:1805.08882}, 2018.

\bibitem[Grefenstette et~al.(2019)Grefenstette, Amos, Yarats, Htut, Molchanov,
  Meier, Kiela, Cho, and Chintala]{higher}
Edward Grefenstette, Brandon Amos, Denis Yarats, Phu~Mon Htut, Artem Molchanov,
  Franziska Meier, Douwe Kiela, Kyunghyun Cho, and Soumith Chintala.
\newblock Generalized inner loop meta-learning.
\newblock \emph{arXiv preprint arXiv:1910.01727}, 2019.

\bibitem[Haarnoja et~al.(2018)Haarnoja, Zhou, Abbeel, and
  Levine]{haarnoja2018soft}
Tuomas Haarnoja, Aurick Zhou, Pieter Abbeel, and Sergey Levine.
\newblock Soft actor-critic: Off-policy maximum entropy deep reinforcement
  learning with a stochastic actor.
\newblock In \emph{International conference on machine learning}, pp.\
  1861--1870. PMLR, 2018.

\bibitem[Ho \& Ermon(2016)Ho and Ermon]{ho2016generative}
Jonathan Ho and Stefano Ermon.
\newblock Generative adversarial imitation learning.
\newblock \emph{arXiv preprint arXiv:1606.03476}, 2016.

\bibitem[Iscen et~al.(2018)Iscen, Caluwaerts, Tan, Zhang, Coumans, Sindhwani,
  and Vanhoucke]{iscen2018policies}
Atil Iscen, Ken Caluwaerts, Jie Tan, Tingnan Zhang, Erwin Coumans, Vikas
  Sindhwani, and Vincent Vanhoucke.
\newblock Policies modulating trajectory generators.
\newblock In \emph{Conference on Robot Learning}, pp.\  916--926. PMLR, 2018.

\bibitem[Jaegle et~al.(2021)Jaegle, Sulsky, Ahuja, Bruce, Fergus, and
  Wayne]{jaegle2021imitation}
Andrew Jaegle, Yury Sulsky, Arun Ahuja, Jake Bruce, Rob Fergus, and Greg Wayne.
\newblock Imitation by predicting observations.
\newblock In \emph{International Conference on Machine Learning}, pp.\
  4665--4676. PMLR, 2021.

\bibitem[{Kalakrishnan} et~al.(2013){Kalakrishnan}, {Pastor}, {Righetti}, and
  {Schaal}]{kalakrishnan_2013_irl}
M.~{Kalakrishnan}, P.~{Pastor}, L.~{Righetti}, and S.~{Schaal}.
\newblock Learning objective functions for manipulation.
\newblock In \emph{2013 IEEE International Conference on Robotics and
  Automation}, pp.\  1331--1336, 2013.

\bibitem[Kalashnikov et~al.(2018)Kalashnikov, Irpan, Pastor, Ibarz, Herzog,
  Jang, Quillen, Holly, Kalakrishnan, Vanhoucke, and Levine]{qt-opt}
Dmitry Kalashnikov, Alex Irpan, Peter Pastor, Julian Ibarz, Alexander Herzog,
  Eric Jang, Deirdre Quillen, Ethan Holly, Mrinal Kalakrishnan, Vincent
  Vanhoucke, and Sergey Levine.
\newblock Scalable deep reinforcement learning for vision-based robotic
  manipulation.
\newblock In \emph{2nd Annual Conference on Robot Learning, CoRL 2018,
  Z{\"{u}}rich, Switzerland, 29-31 October 2018, Proceedings}, volume~87 of
  \emph{Proceedings of Machine Learning Research}, pp.\  651--673. {PMLR},
  2018.
\newblock URL \url{http://proceedings.mlr.press/v87/kalashnikov18a.html}.

\bibitem[Kingma \& Ba(2014)Kingma and Ba]{kingma2014adam}
Diederik~P Kingma and Jimmy Ba.
\newblock Adam: A method for stochastic optimization.
\newblock \emph{arXiv preprint arXiv:1412.6980}, 2014.

\bibitem[{Kumar} et~al.(2021){Kumar}, {Fu}, {Pathak}, and
  {Malik}]{kumar2021rma}
Ashish {Kumar}, Zipeng {Fu}, Deepak {Pathak}, and Jitendra {Malik}.
\newblock Rma: Rapid motor adaptation for legged robots.
\newblock \emph{RSS}, 2021.

\bibitem[Lee et~al.(2021)Lee, Szot, Sun, and Lim]{lee2021generalizable}
Youngwoon Lee, Andrew Szot, Shao-Hua Sun, and Joseph~J Lim.
\newblock Generalizable imitation learning from observation via inferring goal
  proximity.
\newblock \emph{Advances in Neural Information Processing Systems}, 34, 2021.

\bibitem[Levine \& Koltun(2012)Levine and Koltun]{levine2012continuous}
Sergey Levine and Vladlen Koltun.
\newblock Continuous inverse optimal control with locally optimal examples.
\newblock In \emph{Proceedings of the 29th International Coference on
  International Conference on Machine Learning}, pp.\  475--482, 2012.

\bibitem[Lillicrap et~al.(2015)Lillicrap, Hunt, Pritzel, Heess, Erez, Tassa,
  Silver, and Wierstra]{lillicrap2015continuous}
Timothy~P Lillicrap, Jonathan~J Hunt, Alexander Pritzel, Nicolas Heess, Tom
  Erez, Yuval Tassa, David Silver, and Daan Wierstra.
\newblock Continuous control with deep reinforcement learning.
\newblock \emph{arXiv preprint arXiv:1509.02971}, 2015.

\bibitem[Ng et~al.(2000)Ng, Russell, et~al.]{ng2000algorithms}
Andrew~Y Ng, Stuart~J Russell, et~al.
\newblock Algorithms for inverse reinforcement learning.
\newblock In \emph{Icml}, volume~1, pp.\ ~2, 2000.

\bibitem[Ni et~al.(2020)Ni, Sikchi, Wang, Gupta, Lee, and Eysenbach]{ni2020f}
Tianwei Ni, Harshit Sikchi, Yufei Wang, Tejus Gupta, Lisa Lee, and Benjamin
  Eysenbach.
\newblock f-irl: Inverse reinforcement learning via state marginal matching.
\newblock \emph{arXiv preprint arXiv:2011.04709}, 2020.

\bibitem[Ni et~al.(2021)Ni, Sikchi, Wang, Gupta, Lee, and Eysenbach]{ni2021f}
Tianwei Ni, Harshit Sikchi, Yufei Wang, Tejus Gupta, Lisa Lee, and Ben
  Eysenbach.
\newblock f-irl: Inverse reinforcement learning via state marginal matching.
\newblock In \emph{Conference on Robot Learning}, pp.\  529--551. PMLR, 2021.

\bibitem[Orsini et~al.(2021)Orsini, Raichuk, Hussenot, Vincent, Dadashi,
  Girgin, Geist, Bachem, Pietquin, and Andrychowicz]{orsini2021matters}
Manu Orsini, Anton Raichuk, L{\'e}onard Hussenot, Damien Vincent, Robert
  Dadashi, Sertan Girgin, Matthieu Geist, Olivier Bachem, Olivier Pietquin, and
  Marcin Andrychowicz.
\newblock What matters for adversarial imitation learning?
\newblock \emph{arXiv preprint arXiv:2106.00672}, 2021.

\bibitem[Osa et~al.(2018)Osa, Pajarinen, Neumann, Bagnell, Abbeel, and
  Peters]{osa2018algorithmic}
Takayuki Osa, Joni Pajarinen, Gerhard Neumann, J.~Andrew Bagnell, Pieter
  Abbeel, and Jan Peters.
\newblock An algorithmic perspective on imitation learning.
\newblock \emph{Foundations and Trends® in Robotics}, 2018.

\bibitem[Ross et~al.(2011)Ross, Gordon, and Bagnell]{ross2011reduction}
St{\'e}phane Ross, Geoffrey Gordon, and Drew Bagnell.
\newblock A reduction of imitation learning and structured prediction to
  no-regret online learning.
\newblock In \emph{Proceedings of the fourteenth international conference on
  artificial intelligence and statistics}, pp.\  627--635. JMLR Workshop and
  Conference Proceedings, 2011.

\bibitem[Schulman et~al.(2017)Schulman, Wolski, Dhariwal, Radford, and
  Klimov]{schulman2017proximal}
John Schulman, Filip Wolski, Prafulla Dhariwal, Alec Radford, and Oleg Klimov.
\newblock Proximal policy optimization algorithms.
\newblock \emph{arXiv preprint arXiv:1707.06347}, 2017.

\bibitem[Seyed~Ghasemipour et~al.(2019)Seyed~Ghasemipour, Gu, and
  Zemel]{seyed2019smile}
Seyed~Kamyar Seyed~Ghasemipour, Shixiang~Shane Gu, and Richard Zemel.
\newblock Smile: Scalable meta inverse reinforcement learning through
  context-conditional policies.
\newblock \emph{Advances in Neural Information Processing Systems}, 32, 2019.

\bibitem[Szot et~al.(2021)Szot, Clegg, Undersander, Wijmans, Zhao, Turner,
  Maestre, Mukadam, Chaplot, Maksymets, et~al.]{szot2021habitat}
Andrew Szot, Alexander Clegg, Eric Undersander, Erik Wijmans, Yili Zhao, John
  Turner, Noah Maestre, Mustafa Mukadam, Devendra~Singh Chaplot, Oleksandr
  Maksymets, et~al.
\newblock Habitat 2.0: Training home assistants to rearrange their habitat.
\newblock \emph{Advances in Neural Information Processing Systems}, 34, 2021.

\bibitem[Wang et~al.(2021)Wang, Li, and Chan]{wang2021meta}
Pin Wang, Hanhan Li, and Ching-Yao Chan.
\newblock Meta-adversarial inverse reinforcement learning for decision-making
  tasks.
\newblock \emph{arXiv preprint arXiv:2103.12694}, 2021.

\bibitem[Wijmans et~al.(2019)Wijmans, Kadian, Morcos, Lee, Essa, Parikh, Savva,
  and Batra]{wijmans2019dd}
Erik Wijmans, Abhishek Kadian, Ari Morcos, Stefan Lee, Irfan Essa, Devi Parikh,
  Manolis Savva, and Dhruv Batra.
\newblock Dd-ppo: Learning near-perfect pointgoal navigators from 2.5 billion
  frames.
\newblock \emph{arXiv preprint arXiv:1911.00357}, 2019.

\bibitem[Wulfmeier et~al.(2017)Wulfmeier, Rao, Wang, Ondruska, and
  Posner]{wulfmeier2017large}
Markus Wulfmeier, Dushyant Rao, Dominic~Zeng Wang, Peter Ondruska, and Ingmar
  Posner.
\newblock Large-scale cost function learning for path planning using deep
  inverse reinforcement learning.
\newblock \emph{The International Journal of Robotics Research}, 36\penalty0
  (10):\penalty0 1073--1087, 2017.

\bibitem[Xu \& Denil(2019)Xu and Denil]{xu2019positive}
Danfei Xu and Misha Denil.
\newblock Positive-unlabeled reward learning.
\newblock \emph{arXiv preprint arXiv:1911.00459}, 2019.

\bibitem[Xu et~al.(2019)Xu, Ratner, Dragan, Levine, and Finn]{xu2019learning}
Kelvin Xu, Ellis Ratner, Anca Dragan, Sergey Levine, and Chelsea Finn.
\newblock Learning a prior over intent via meta-inverse reinforcement learning.
\newblock In \emph{International Conference on Machine Learning}, pp.\
  6952--6962. PMLR, 2019.

\bibitem[Yu et~al.(2019)Yu, Yu, Finn, and Ermon]{yu2019meta}
Lantao Yu, Tianhe Yu, Chelsea Finn, and Stefano Ermon.
\newblock Meta-inverse reinforcement learning with probabilistic context
  variables.
\newblock \emph{Advances in Neural Information Processing Systems},
  32:\penalty0 11772--11783, 2019.

\bibitem[Ziebart et~al.(2008)Ziebart, Maas, Bagnell, and
  Dey]{ziebart2008maximum}
Brian~D Ziebart, Andrew~L Maas, J~Andrew Bagnell, and Anind~K Dey.
\newblock Maximum entropy inverse reinforcement learning.
\newblock In \emph{Aaai}, volume~8, pp.\  1433--1438. Chicago, IL, USA, 2008.

\bibitem[Zolna et~al.(2019)Zolna, Reed, Novikov, Colmenarej, Budden, Cabi,
  Denil, de~Freitas, and Wang]{zolna2019task}
Konrad Zolna, Scott Reed, Alexander Novikov, Sergio~Gomez Colmenarej, David
  Budden, Serkan Cabi, Misha Denil, Nando de~Freitas, and Ziyu Wang.
\newblock Task-relevant adversarial imitation learning.
\newblock \emph{arXiv preprint arXiv:1910.01077}, 2019.

\bibitem[Zolna et~al.(2020)Zolna, Saharia, Boussioux, Hui, Chevalier-Boisvert,
  Bahdanau, and Bengio]{zolna2020combating}
Konrad Zolna, Chitwan Saharia, L{\'e}onard Boussioux, David Yu-Tung Hui, Maxime
  Chevalier-Boisvert, Dzmitry Bahdanau, and Yoshua Bengio.
\newblock Combating false negatives in adversarial imitation learning.
\newblock \emph{arXiv preprint arXiv:2002.00412}, 2020.

\end{thebibliography}
